\pdfoutput=1

\documentclass[11pt]{article}

\usepackage[final]{acl}

\usepackage{times}
\usepackage{latexsym}

\usepackage[T1]{fontenc}

\usepackage[utf8]{inputenc}

\usepackage{microtype}

\usepackage{inconsolata}

\usepackage{xspace}
\usepackage{rotating}
\usepackage{adjustbox}
\usepackage{booktabs}
\usepackage{multirow}
\usepackage{graphicx}
\usepackage{booktabs}
\usepackage{makecell}
\usepackage{fontawesome}
\usepackage{xcolor}
\usepackage{amsmath}
\usepackage{bbm}
\usepackage[framemethod=tikz]{mdframed}
\usepackage{enumitem}
\usepackage{float}
\usepackage{tikz}
\usepackage{colortbl}
\usepackage{wrapfig}
\usepackage{subcaption}
\usepackage{adjustbox}
\usepackage{comment}
\usepackage{pifont}
\usepackage{bm}

\newlist{compactitem}{enumerate}{3} %
\setlist[compactitem,1]{label=\textbullet, nosep, leftmargin=*}

\newlist{cspecialitem}{enumerate}{3} %
\setlist[cspecialitem,1]{label=$c_\arabic*$:, nosep, leftmargin=*}%

\usepackage{scalerel,xparse}

\definecolor{realized}{RGB}{0,147,191} %
\definecolor{inbetween}{RGB}{87,230,163} %
\definecolor{unrealized}{RGB}{197,252,233} %
\definecolor{darkpurple}{RGB}{146,0,219} %
\definecolor{lightpurple}{RGB}{252,243,255} %
\definecolor{darkgreen}{rgb}{0.0, 0.39, 0.0}

\newcommand\sect[1]{\S\ref{#1}}

\newcommand{\realized}{\tikz{\fill[realized, rounded corners=2pt] (0,0) rectangle (0.3,0.3);}}
\newcommand{\unrealized}{\tikz{\fill[unrealized, rounded corners=2pt] (0,0) rectangle (0.3,0.3);}}

\usepackage{amsmath}

\newmdenv[linecolor=darkpurple,linewidth=2pt,roundcorner=10pt,backgroundcolor=lightpurple]{customframe}

\NewDocumentCommand\staremoji{}{{\Large$\star$}}

\usepackage{amssymb}
\NewDocumentCommand\thumbemoji{}{{\large$\checkmark$}}

\newcommand{\quotes}[1]{``#1''}

\newcommand{\cmark}{{\color{darkgreen}\ding{51}}}%
\newcommand{\xmark}{{\color{red}\ding{55}}}%

\definecolor{darkblue}{rgb}{0, 0, 0.5}
\hypersetup{
    colorlinks = true, %
    linkcolor = darkblue, %
    citecolor = darkblue, %
    filecolor = black, %
    urlcolor = darkblue, %
    pdfborder = {0 0 0} %
}

\title{On Evaluating Explanation Utility for Human-AI Decision Making in NLP}

\author{%
\begin{tabular}{c}
Fateme Hashemi Chaleshtori \quad\quad Atreya Ghosal \quad\quad Alexander Gill\\
Purbid Bambroo  \quad\quad Ana Marasovi\'c
\end{tabular} \\
Kahlert School of Computing\\
University of Utah\\
\texttt{fateme.hashemi@utah.edu}
}

\begin{document}

\maketitle

\setlength{\abovedisplayskip}{3pt}
\setlength{\belowdisplayskip}{3pt}

\begin{abstract}
\emph{Is explainability a false promise?} %
This debate has emerged from the insufficient evidence that explanations help people in situations they are introduced for. %
More human-centered, \emph{application-grounded} evaluations of explanations are needed to settle this. %
Yet, with no established guidelines for such studies in NLP, researchers accustomed to standardized proxy evaluations must discover appropriate measurements, tasks, datasets, and sensible models for human-AI teams in their studies.  %

To aid with this, we first review existing metrics suitable for application-grounded evaluation. %
We then establish criteria to select appropriate datasets, and using them, we find that only 4 out of over 50 datasets available for explainability research in NLP meet them. %
We then demonstrate the importance of reassessing the state of the art to form and study human-AI teams: teaming people with models for certain tasks might only now start to make sense, and for others, it remains unsound. %
Finally, we present the exemplar studies of human-AI decision-making for one of the identified tasks\,---\,verifying the correctness of a legal claim given a contract. %
Our results show that providing AI predictions, with or without explanations, does not cause decision makers to speed up their work without compromising performance. %
We argue for revisiting the setup of human-AI teams and improving automatic deferral of instances to AI, where explanations could play a useful role\footnote{Our code and human study templates: \url{https://github.com/utahnlp/nlp-explanation-utility-guideline/}}.

\end{abstract}

\section{Introduction}

Decision makers can make use of imperfect models if they can detect when they are correct. %
Explanations of individual predictions are proposed to this end as they are expected to reveal useful signals about the model's reasoning process \citep{10.1145/3442188.3445923}. %
Before undertaking realistic evaluations involving people, NLP researchers aspired to first implement working methods. %
Thus, prior NLP explainability work has mostly focused on overcoming technical challenges and used proxy evaluations. %
Consequently, human-centered evaluations of explanations grounded in real NLP applications are scarce. %
There is a prevailing perspective that this now needs to change since explainability methods passed proof-of-concept tests (see \href{https://xai-hcee.github.io/}{Human-centered Evaluations of Explanations Tutorial}).  %
However, given that this is a nascent NLP research space and the notable variation among prior studies  (see Table \ref{tab:overview-prior-work}), choosing explanation evaluation measurements, tasks, models to explain, baseline conditions, and many other study design choices is not straightforward. %
This paper aims to alleviate this difficulty by providing guidelines. 

An existing resource for the development and evaluation of explanations in NLP already includes over 50 datasets. %
Can these be used for application-grounded evaluations of explanations? 
To answer this, in \sect{sec:task_appropriateness}, we establish criteria to assess each dataset's suitability for computing explanation usefulness with the measurements overviewed in \sect{sec:background}. %
We discover that 17/53 datasets are apt for studying appropriate reliance and complementary human-AI team performance but only involve low risk. 
4/53 additionally involve higher risk and do not have quality concerns. %
We recommend using these four, as high-stakes scenarios benefit more from effective explanations.

We introduce the final criterion for dataset selection based on how the likelihood of hazards, and therefore risk, changes if the performance of a state-of-the-art model peaks or is so low the model cannot be collaborated with. %
We show that this criterion requires continuous assessment of model performance in a rapidly evolving field like NLP. 

Finally, we present explanation usefulness studies for a task identified by our meta-analysis: verifying a legal claim given a contract. %
These serve as exemplars to NLP researchers planning similar studies. %
We use both the common human-AI setup, where people make all final decisions with AI assistance, and an overlooked setup, where people decide only for those referred to them by a deferral model. %
We isolate the effect of explanations, use strong baseline conditions, deploy multiple cognitive forcing functions, integrate real-world situations, and implement attention checks tailored to the application. %
We show that input highlights and influential training examples do not improve human decision making assisted with the model's predictions and confidence. %
However, we find that major advances are needed for functional human-AI teams in NLP, \emph{regardless of explanations}. %
We discover that successful deferral is a promising direction to this end.

\section{Review of Application-Grounded Explanation Evaluation}
\label{sec:background}

In Table \ref{tab:overview-prior-work} (Appendix), we overview prior human-centered application-grounded evaluations of explanations of NLP models \cite{10.1145/3287560.3287590, DBLP:conf/iui/FengB19, gonzalez-etal-2021-explanations, 10.1145/3411764.3445717, DBLP:conf/iui/SchemmerKBBS23, DBLP:conf/nips/MozannarLWSDS23, joshi-etal-2023-machine, DBLP:journals/corr/abs-2310-12558}. %
There is a notable variation in the choice of models explained, explanations used, evaluation measurements, baseline conditions, datasets, and outcomes among them. %
To conclusively establish, or disprove, the value of explanations for human-AI decision-making in NLP, more research is needed together with a more rigorous evaluation protocol designed to collectively guide us towards settling this matter. %
To assist with running such studies, we start with an overview of suitable explanation usefulness measurements as they are not yet widely adopted in NLP research.

\paragraph{Taxonomy of explanation evaluation.} \citet{doshivelez2017rigorous} categorize evaluations of explanations as: (1) \emph{proxy} (no humans, proxy tasks; e.g., the proportion of all features selected as important), (2) \emph{human-grounded} (with humans, but simplified tasks; e.g., simulatability), or (3) \emph{application-grounded} (with humans, realistic tasks; e.g., human-AI decision making). %
Human-AI decision-making, which is the focus of this paper, is one of the six usage contexts within explainable AI \cite{Liao_Zhang_Luss_Doshi-Velez_Dhurandhar_2022}. %
Forward and counterfactual simulatability, that are common in NLP \citep{xie-etal-2022-calibrating, DBLP:conf/aaai/AroraPSCLN22}, are human- but not application-grounded; \citet{10.1145/3377325.3377498} show that explanations affect simulatability and human-AI decision-making differently. %

\paragraph{Reliance definitions.} It is often asserted that explanations can deter people from rejecting correct predictions, i.e., \textbf{underreliance}. %
This expectation stems from assuming that the model is correct for the right reasons, and explanations are anticipated to unveil this. %
Explanations could also aid people in rejecting incorrect predictions, thereby countering \textbf{overreliance}. %
This becomes possible when explanations present information that appears illogical, self-contradictory, or inconsistent with what the person already knows. %
The ultimate goal is \textbf{appropriate reliance}\,---\,have people accept correct predictions and dismiss erroneous ones. %
A gain in the average rate at which people do so upon seeing explanations quantifies their usefulness. 

\citet{fok2023search} define desired reliance behavior based on expected performance, which implies it is fine to accept a ``super-human'' model's predictions (or reject a ``sub-human'' model's predictions) even if some are wrong (correct). 

\paragraph{Measuring reliance.}
Researchers rarely ask people to accept/reject predictions to measure reliance, except \citet{gonzalez-etal-2021-explanations}. %
Instead, people are often shown model predictions and asked to make the final decision.\footnote{The accept/reject setup might seem a step removed from actual decision making, thus less grounded in application.}  %
Overreliance is measured by how often the final decision agrees with the model's when it is wrong \citep{DBLP:journals/pacmhci/VasconcelosJGGBK23}. %
A possible confounder is that people might make the same wrong decisions as the model, not because they are blindly following it, but because they genuinely find the same wrong answer to be correct. %

\citet{DBLP:conf/iui/SchemmerKBBS23} thus propose that participants first make a guess unassisted, then reevaluate upon viewing the model's prediction. %
They propose reporting the fraction of times a person (1) flips their initial, wrong judgment after seeing a correct model prediction, and (2) sticks with their initial, correct judgment after seeing a wrong model prediction.\footnote{The switch percentage \citep{10.1145/3351095.3372852} is related.} %
\citet{10.1145/3397481.3450650} aim for a similar procedure but allow scrolling to the model prediction while making the first guess, potentially influencing the standalone guess. %
\citet{joshi-etal-2023-machine}'s approach is similar to (1), but a person needs to flip their initial, wrong answer to the correct one upon seeing AI's explanation but not its prediction.\footnote{\citet{joshi-etal-2023-machine} also study whether model explanations support the human ability to reason about new situations where the same logic applies, like human explanations do \citep{Blanchard2018StabilityBA, Vasilyeva2022ExplanationsAC}. %
However, their approach cannot be applied once the model is deployed. %
}

This approach to measuring reliance where people make all final decisions with AI assistance is also how \textbf{complementary human-AI team performance} is typically measured. %
Human-AI teams should surpass the accuracy of both the AI alone and the human alone \cite{10.1145/3411764.3445717}, and explanations could provide a boost. %
For this to even be possible, the performance of a state-of-the-art model or time-constrained people alone should not already peak. %
It should not be too low either, because then collaborating with such a model would not be advisable. %
Instead of reporting the gain (if any) of teaming up and explanations, \citet{DBLP:conf/iui/FengB19} perform a regression analysis. %

Providing a model prediction with its confidence is a simple, yet stronger \textbf{baseline condition} for explanation usefulness compared to predictions only \cite{10.1145/3411764.3445717}. %
However, not all prior studies test the explanations relative to displaying model confidence; see Table \ref{tab:overview-prior-work} (Appendix). %
When measuring reliance or complementary performance, it is common to ask annotators to self-report their \textbf{confidence} in their decisions and \textbf{trust} in the AI model on a case-by-case basis or as a post-task survey. %

\paragraph{Deferral.} Integrating a deferral model \cite{Dvijotham2023EnhancingTR}, $\mathcal{M}_D$, that decides whether an instance can be correctly processed by a prediction model, $\mathcal{M}_P$, presents an alternative to having people make all final decisions with AI in the loop. %
Explanation usefulness has not been studied in this human-AI team setup. %
Feeding explanations to $\mathcal{M}_D$ could enhance its correctness, if explanations indeed indicate when $\mathcal{M}_P$ is correct or wrong, as commonly assumed. %
They could also assist the human reviewer that gets a small fraction of instances deemed hard for $\mathcal{M}_P$. 
Knowing that a highly accurate model likely made a mistake on a given example, along with its reasoning for it, can nudge the reviewer to consider why the model erred and preempt them from making the same mistake. %
Surprisingly, we find that these explanations point to the correct evidence (\sect{sec:study2}).

\section{Analysis of Task Appropriateness}
\label{sec:task_appropriateness}

In this section, we present criteria that can be used to determine the suitability of tasks for application-grounded human evaluations of explanations  (\sect{sec:data_criteria}) and analyze 53 existing datasets introduced for developing and evaluating explanations in NLP (\sect{sec:exnlp_categorization}). We refer to a task as its realization in the data. 

\subsection{Task Criteria}
\label{sec:data_criteria}

We determine that the following criteria must be fulfilled to ensure that evaluations are rooted in genuine human-AI interactions:%
\begin{customframe}
\begin{cspecialitem}
\item The task has a meaningful connection to a real-world application, involving people who seek model outputs and act on them. 
\item The dataset inputs must be realistic. 
\item Task instances require a notable effort from people, or people are bad at them. 
\end{cspecialitem}
\end{customframe}

An application must exist if the goal is to evaluate the usefulness of explanations within the context of human-AI interaction for that application. %
For example, \textsc{CommonsenseQA} \citep{talmor-etal-2019-commonsenseqa}, one of the most popular datasets in NLP explainability research, has no associated application as people do not need answers to questions such as ``At the end of your meal what will a waiter do? serve food, eat, set table, serve meal, or present bill''. %
A dataset might be related to an application, but the task could be narrowed down to take inputs that would never realistically occur. %
For instance, \textsc{PubHealth} \citep{kotonya-toni-2020-explainable-automated} has actionable outputs but lacks realistic task inputs. %
The task is to verify a claim based on a professional fact-checking report on the same claim that won't be available for an unverified claim post-deployment. 
Finally, if a task instance is easy and quick to handle, collaboration with AI is unnecessary: the person already knows what to do, making the reliance irrelevant. %
For example, while there might be a use for sentiment classification of laptop reviews \citep{pontiki-etal-2014-semeval}, their brief average length of only 15 words allows people to 
correctly and confidently gauge sentiment without assistance. %
Hence, concerns about under- or overreliance do not arise in this context because people never end up %
really relying on anything.

These three criteria are sufficient if the sole focus is on reliance/complementary performance. %
However, the definition of human trust in AI \citep{10.1145/3442188.3445923} implies that trust inherently involves risk, as one cannot accept vulnerability when none exists. %
Thus, studying human trust in AI demands an extra criterion: 
\begin{customframe}
\begin{cspecialitem}[start=4]
\item  There is some undesirable event that can possibly (but not certainly) occur when collaborating with models for the task. 
\end{cspecialitem}
\end{customframe}

Although risk is not pivotal to defining sound studies of reliance and human-AI teams, \textbf{we urge giving precedence to tasks involving higher risk because under- and over\-re\-li\-ance have more pronounced consequences for them.} 
It is more valuable to develop explanations that boost appropriate reliance for them, and this is how the need for explanations is often motivated. 
The necessity of the data selection criterion $c_4$ might be questioned if risk can only be simulated. We expect simulations of risk to provide more meaningful insights about high-risk scenarios than using a task that inherently lacks any moderate risk.

\subsection{Categorization of ExNLP Tasks}
\label{sec:exnlp_categorization}

We analyze datasets that are reported on the website that collects datasets for explainable NLP \citep{wiegreffe_marasovic_2021} according to how they satisfy the criteria in \sect{sec:data_criteria}.\footnote{\url{https://exnlpdatasets.github.io/}; We excluded datasets in non-English languages and multimodal datasets.}
In Appendix \ref{sec:exnlp_task_cards}, we report details of our decisions for each task and provide an overview in Table \ref{tab:exnlp_analysis}. %
We use \begin{tikzpicture} \fill[realized, rounded corners=1pt] (0,0) rectangle (0.24,0.24);\end{tikzpicture} if a benchmark criterion is satisfied, and \begin{tikzpicture} \fill[unrealized, rounded corners=1pt] (0,0) rectangle (0.24,0.24);\end{tikzpicture} otherwise. 
A suitable dataset for application-grounded evaluations of explanations should have an application ($c_1$) and realistic inputs ($c_2$) as well as either require notable effort, or be a difficult task for people ($c_3$), and ideally more than low levels of risk ($c_4$). 
We mark tasks that satisfy $c_{\{1,2,3\}}$, i.e., those suitable for studying reliance with \thumbemoji\xspace and those that satisfy all criteria and that should be prioritized with \staremoji. 

\paragraph{Are ExNLP tasks connected to real-world applications beyond debugging?} We first determine that we can imagine people using the outputs of a model trained on dataset instances. %
E.g., sentiment predictions of reviews can be used to decide whether to make a purchase. %
We then assess that task instances resemble what models can realistically access to make their predictions in the future (unlike the fact-checking example in \sect{sec:data_criteria}). %
If both of these two conditions are met, we deem that a task is connected to real-world application, and not otherwise. 
We find that 30/53 (56.6\%) datasets have an associated application and realistic inputs, i.e., fulfill the central requirement for \emph{application}-grounded evaluations, but 23/53 (43.4\%) do not. %

\begin{table}[H]
\centering
\small
\resizebox{0.84\columnwidth}{!}{
\begin{tabular}{llcccc}
\toprule
& & $c_1\land c_2$ & $c_3$ & $c_4$ & \\ 
\midrule
\hyperlink{work1}{W1},\hyperlink{work6}{6}  & \hyperlink{dataset1}{D1} & \realized & \realized & \unrealized & \thumbemoji \\ 
\hyperlink{work2}{W2} & \hyperlink{dataset2}{D2}& \realized & \realized & \unrealized & \thumbemoji \\ 
\multirow{2}{*}{\hyperlink{work3}{W3}} & \hyperlink{dataset3}{D3} & \unrealized & \realized & - & \\ 
& \hyperlink{dataset4}{D4} & \realized & \unrealized & \unrealized & \\ 
\hyperlink{work4}{W4} & \hyperlink{dataset5}{D5} & \realized & \realized & \unrealized & \thumbemoji\\ 
\multirow{2}{*}{\hyperlink{work5}{W5}} & \hyperlink{dataset6}{D6} & \realized & \realized & \unrealized & \thumbemoji \\
& \hyperlink{dataset7}{D7} & \unrealized & \unrealized & - & \\ 
\hyperlink{work7}{W7} & \hyperlink{dataset8}{D8} & \unrealized & \realized & \unrealized & \\
\hyperlink{work8}{W8} & \hyperlink{dataset9}{D9} & \realized & \unrealized & \unrealized & \\ 
& \hyperlink{dataset10}{D10} & \realized & \unrealized & \unrealized & \\ 
& \hyperlink{dataset11}{D11} & \unrealized & \realized & - & \\ 
& \hyperlink{dataset12}{D12} & \realized & \unrealized & \unrealized & \\ 
& \hyperlink{dataset13}{D13} & \unrealized & \unrealized & - & \\ 
& \hyperlink{dataset14}{D14} & \unrealized & \unrealized & - & \\ 
& \hyperlink{dataset15}{D15} & \unrealized & \realized & - & \\ 
& \hyperlink{dataset16}{D16} & \unrealized & \realized & - & \\ 
& \hyperlink{dataset17}{D17} & \realized & \realized & \realized & \staremoji$^*$ \\ 
& \hyperlink{dataset18}{D18} & \realized & \realized & \realized & \staremoji$^*$ \\ 
& \hyperlink{dataset19}{D19} & \unrealized & \unrealized & - & \\ 
& \hyperlink{dataset20}{D20} & \unrealized & \unrealized & - & \\ 
& \hyperlink{dataset21}{D21} & \unrealized & \unrealized & - & \\ 
& \hyperlink{dataset22}{D22} & \realized & \realized & \unrealized & \thumbemoji \\ 
& \hyperlink{dataset23}{D23} & \realized & - & - & \\ 
& \hyperlink{dataset24a}{D24a} & \realized & \unrealized & \unrealized & \\ 
& \hyperlink{dataset24b}{D24b} & \realized & \unrealized & \unrealized & \\ 
& \hyperlink{dataset25}{D25} & \unrealized & \unrealized & - & \\ 
& \hyperlink{dataset26}{D26} & \unrealized & \unrealized & - & \\ 
& \hyperlink{dataset27}{D27} & \unrealized & \unrealized & - & \\ 
& \hyperlink{dataset28}{D28} & \realized & \realized & \unrealized & \\ 
& \hyperlink{dataset29}{D29} & \realized & \realized & \unrealized & \thumbemoji \\ 
& \hyperlink{dataset30}{D30} & \unrealized & \unrealized & - & \\ 
& \hyperlink{dataset31}{D31} & \realized & \realized & \unrealized & \thumbemoji\\
& \hyperlink{dataset32}{D32} & \realized & \realized & \unrealized & \thumbemoji\\ 
& \hyperlink{dataset33}{D33} & \unrealized & \unrealized & - & \\ 
& \hyperlink{dataset34}{D34} & \realized & \realized & \unrealized & \thumbemoji\\ 
& \hyperlink{dataset35}{D35} & \realized & \unrealized & \unrealized & \\ 
& \hyperlink{dataset36}{D36} & \realized & \unrealized & \unrealized & \\ 
& \hyperlink{dataset37}{D37} & \realized & - & - & \\ 
& \hyperlink{dataset38}{D38} & \realized & \realized & \realized& \staremoji \\ 
& \hyperlink{dataset39}{D39} & \unrealized & \unrealized & - & \\ 
& \hyperlink{dataset40}{D40} & \realized & \unrealized & \unrealized & \\ 
& \hyperlink{dataset41}{D41} & \realized & - & - & \\ 
& \hyperlink{dataset42}{D42} & \realized & \realized & \realized & \staremoji \\ 
& \hyperlink{dataset43}{D43} & \unrealized & \unrealized & - & \\ 
& \hyperlink{dataset44}{D44} & \realized & \realized & \realized & \staremoji\\ 
& \hyperlink{dataset45}{D45} & \unrealized & - & - & \\ 
& \hyperlink{dataset46}{D46} & \unrealized & \realized & - & \\ 
& \hyperlink{dataset47}{D47} & \realized & - & \unrealized & \\ 
& \hyperlink{dataset48}{D48} & \realized & - & - & \\ 
& \hyperlink{dataset49}{D49} & \realized & \realized & \realized & \staremoji\\ 
& \hyperlink{dataset50}{D50} & \realized & \realized & \realized & \staremoji\\
& \hyperlink{dataset51}{D51} & \unrealized & \realized & - & \\ 
& \hyperlink{dataset52}{D52} & \unrealized & - & - & \\ 
\bottomrule
\end{tabular}
}
\caption{Categorization of ExNLP datasets. See \sect{sec:exnlp_categorization} for a description of symbols.}
\label{tab:exnlp_analysis}
\end{table}

\paragraph{Do ExNLP tasks require notable human effort? Are people skilled at solving these tasks?} We estimate effort using the avg.\ length of task inputs,  
 anticipating that longer inputs demand more effort. %
The maximum average length that we decide does not need notable effort is 272 words, taking around a minute to read \citep{Rayner2016SoMT}. %
We estimate human ability using reported human performance when available. %
We find that 25/53 (47.2\%) tasks either require notable effort or people do not excel at it, the data is not available for 2/53 (3.8\%), we are not able to estimate the human ability for 5/53 (9.4\%), and for 21/53 (39.6\%) inputs are too short while people do the task well. %
Of 25 requiring notable effort or people are not good at them, 17/53 (32.1\%) also have associated applications and realistic inputs. %
That is, 32.1\% of ExNLP datasets are suitable for studying appropriate reliance and complementary team performance.

\paragraph{Are ExNLP tasks associated with high-risk situations?} 
Motivated by \citet{10.1145/3411764.3445088}, we answer this question from the perspective of 2 stakeholders: (i) people acting on the model output (e.g., doctors) and (ii) decision subjects (e.g., patients). %
We first determine possible hazards. %
We decide what a hazard's level of risk is\,---\,low, moderate, or high\,---\,based on its severity and likelihood. %
We estimate the likelihood based on the performance of the state-of-the-art model, expecting that the higher the performance is, the lower the likelihood. %
We subjectively determine their worst-case severity. %
We find that among the 17 remaining datasets, only 6 cause hazards that are not benign. %
Upon manual inspection of examples of this data, we discovered problems with \protect\hyperlink{dataset17}{D17} and \protect\hyperlink{dataset18}{D18} (see Appendix \ref{sec:liar_raw_analysis}). %
We exclude them and recommend prioritizing 4 datasets for application-grounded evaluation of explanations in NLP: EvidenceInference v2 \emph{with document retrieval} (\protect\hyperlink{dataset38}{D38}),  SciFact-Open (\protect\hyperlink{dataset44}{D44}), ContractNLI (\protect\hyperlink{dataset49}{D49}), and Indian Legal Documents Corpus (ILDC; \protect\hyperlink{dataset50}{D50}).

\subsection{Task Checks with Model Performance}
\label{sec:model_appropriateness}

The final check for a dataset's suitability is based on a chosen model performance: %
\begin{customframe}
\begin{cspecialitem}[start=5]
\item  The model performance should be high enough to warrant collaboration, but not so high that it can operate effectively on its own without human oversight.
\end{cspecialitem}
\end{customframe}
If a model rarely makes mistakes, the likelihood and thus risk of hazards are typically low. %
These are tasks we recommend deprioritizing. %
\citet{fok2023search} argue that using the predictions is viable in this case. %
However, a worse model should not be chosen when a better-performing one is available and resource-appropriate. %

\begin{table}[t]
    \small
    \centering
        \begin{subtable}[t]{0.85\columnwidth}
        \centering
        \begin{tabular}{lrrr}\toprule
        &\textbf{Precision} &\textbf{Recall} &\textbf{F1 score}\\\midrule
        \multicolumn{4}{l}{\textbf{Flan-T5-3B (our)}} \\
        \arrayrulecolor{black!20}\midrule
        \textsc{Entail} &92.5 &93.7 &93.1 \\
        \textsc{No-mention} &93.0 &87.0 &89.9 \\
        \textsc{Contradict} &68.7 &82.7 &75.0 \\
        \arrayrulecolor{black!20}\midrule
        \textsc{Micro Avg.} &90.2 &89.7 &89.8 \\
        \textsc{Macro Avg.} & 84.7 & 87.7 & 86.0 \\
        \textsc{Macro (E,C)} & 80.6 & 88.2 & \textbf{84.1} \\
        \arrayrulecolor{black}\midrule
        \multicolumn{4}{l}{\textbf{BERT-Large} }\\ %
        \arrayrulecolor{black!20}\midrule
        \textsc{Entail} &- &- &83.4  \\
        \textsc{Contradict} &- &- &35.7 \\
        \arrayrulecolor{black!20}\midrule
        \textsc{Macro (E,C)} & - & - & 59.6 \\
        \arrayrulecolor{black}\bottomrule
        \end{tabular}
        \caption{ContractNLI}
        \label{tab:contractnli_results_all}
    \end{subtable}%
    \vspace{3mm} %
    \begin{subtable}[t]{0.85\columnwidth}
        \centering
        \begin{tabular}{lrrr}\toprule
        &\textbf{Precision} &\textbf{Recall} &\textbf{F1 score}\\\midrule
        \multicolumn{4}{l}{\textbf{Flan-T5-3B (our)}} \\
        \arrayrulecolor{black!20}\midrule
        \textsc{Increase}        &52.7	    &64.4	     &58.8 \\
        \textsc{No Diff}         &54.7	    &29.2	     &38.1 \\
        \textsc{Decrease}        &41.1	    &59.1	    &48.5 \\
        \arrayrulecolor{black!20}\midrule
        \textsc{Micro Avg.}      &50.7	    &49.1	     &47.6 \\
        \textsc{Macro Avg.}      &49.5	    &51.6	     & 48.5 \\
        \arrayrulecolor{black}\bottomrule
        \end{tabular}
        
        \caption{EvidenceInference v2 \emph{with retrieval}}
        \label{tab:ev_infer_results_all}
    \end{subtable}
    \caption{Finetuned Flan-T5-3B and the state-of-the-art reported results.   \citet{koreeda-manning-2021-contractnli-dataset} do not report finetuned BERT-Large F1 for the ``No-mention'' class.  EvidenceInference is proposed without retrieval, so there is no prior model performance to report. Results for SciFact and ILDC are in Table \ref{tab:results_ildc_scifact_appendix} (Appendix). Macro F1 averages F1 scores across classes, while Micro F1 calculates F1 from aggregated results.
    }
    \label{tab:all_results}
\end{table}

\begin{figure*}[t]
    \centering
    \includegraphics[width=0.79\textwidth]{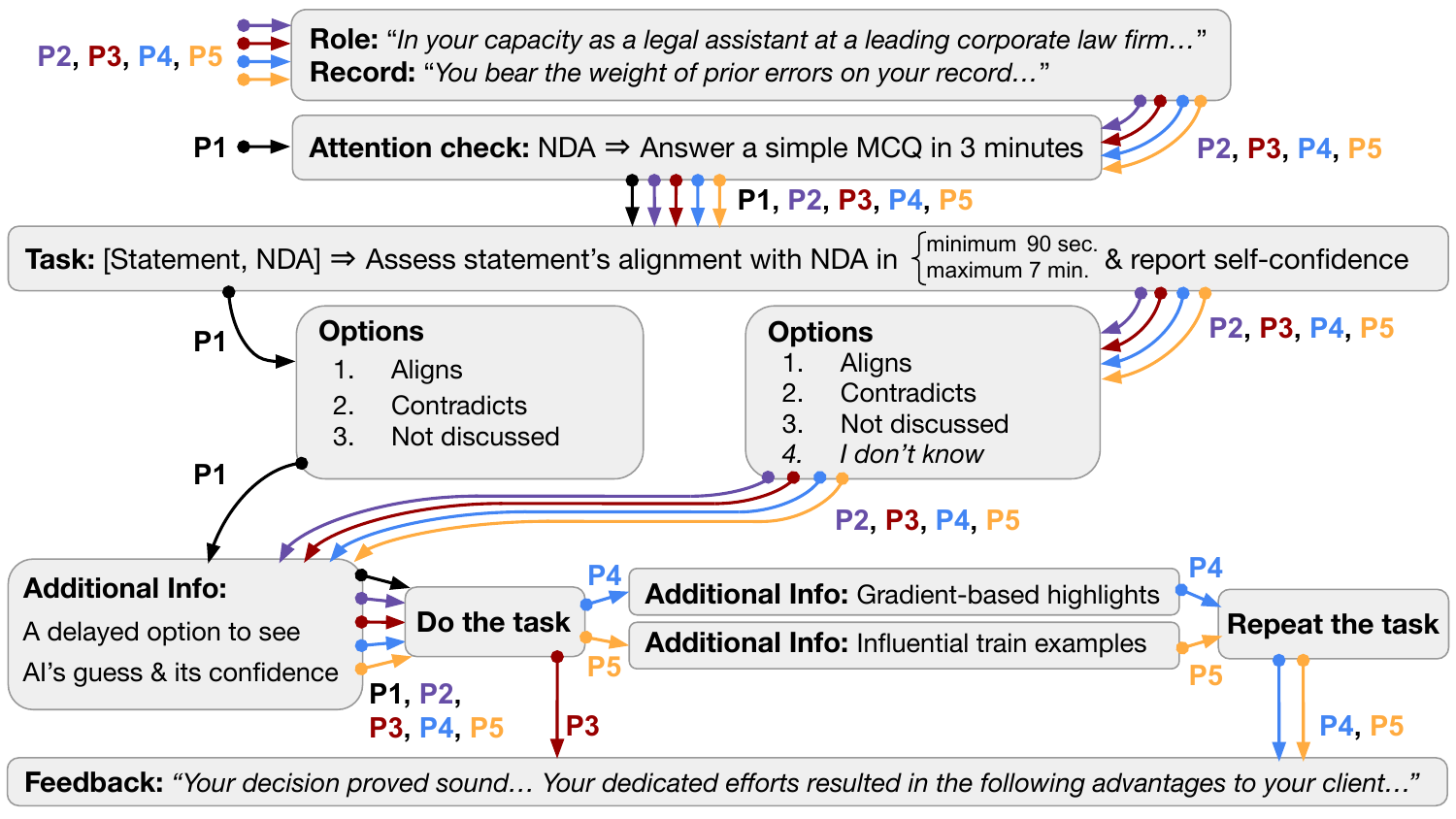}
    \caption{An overview of the five task presentations, \texttt{P1}--\texttt{P5}, used for Study I (see \sect{sec:task_conditions}).} 
    \label{fig:studyflow}
\end{figure*}

Finetuning large language models (LLMs) is an effective method for specializing a model to a task. 
We do so for each \staremoji\xspace task that fulfills all the criteria in \sect{sec:data_criteria} and quality checks. %
We use Flan-T5-3B \citep{DBLP:conf/iclr/WeiBZGYLDDL22} due to its size and versatility stemming from instruction finetuning with data of 1.8K tasks.\footnote{We did not explore prompting because we did not expect notable improvements from it compared to finetuning Flan-T5-3B at the time of writing this paper. We did not train Flan-T5 to self-rationalize with free-text explanations, as human-authored free-text explanations are not available for supervision.} %
Details and examples of task inputs to the model (Tables \ref{tab:scifactopen_example}--\ref{tab:evidence_infer_example}) are in Appendix \ref{baseline_appendix}. %

Tables \ref{tab:all_results} and \ref{tab:results_ildc_scifact_appendix} (Appendix) underscore the importance of reassessing baselines. 
We obtain a 24.5 point improvement in the average contradiction and entailment F1 scores on Contract-NLI (Table \ref{tab:all_results}), and a 22.8 macro-F1 point increase on SciFact-Open (Table \ref{tab:results_ildc_scifact_appendix}).  
Without reassessing performance on these datasets, we would not realize that it now makes sense to team up people with models for these tasks. %
Moreover, we find that the baselines for ILDC and SciFact-Open\,---\,the two datasets with the reported human performance\,---\,have not reached peak performance. %
Thus, human-AI teams might provide benefits over using AI alone. The performance of EvidenceInference-v2 with retrieval remains low, and studying human-AI teams on this task is not justified without a stronger model in the loop.  %

\section{Study I: People Make All Decisions}
\label{sec:study1_main}

In \sect{sec:background}, we overview two strategies for human-AI decision-making: (1) people make all final decisions with AI assistance, and (2) a deferral model refers only a fraction to people. %
We aim to provide an exemplar for user study design and initial insights into the usefulness of explanations in improving human-AI decision making in both of these setups for one of the tasks identified in our meta-analysis: verifying claims based on the ContractNLI Non-Disclosure Agreements (NDAs). %
In this section, we use (1), and 
in \sect{sec:study2}, we look at (2). 

\begin{table*}[t]
\centering
\resizebox{\textwidth}{!}{%
\begin{tabular}{p{0.25cm}p{4.73cm}>{\raggedleft\arraybackslash}p{2.5cm}>{\raggedleft\arraybackslash}p{1.9cm}>{\raggedleft\arraybackslash}p{1.9cm}>{\raggedleft\arraybackslash}p{2.2cm}>{\raggedleft\arraybackslash}p{2.1cm}>{\raggedleft\arraybackslash}p{1.1cm}}
\toprule
 &  & F1($\mathcal{M}$)-F1($\mathcal{H}$) $\downarrow$ & $\frac{\texttt{\# }\mathcal{H} \text{ correct}}{\texttt{\# } \mathcal{M} \text{ correct}}$ $\uparrow$ & $\frac{\texttt{\# } \mathcal{H} \text{ correct}}{\texttt{\# } \mathcal{M} \text{ wrong}}$ $\uparrow$ & Avg.\ \texttt{\#} Reveal & Avg.\ $\mathcal{H}$ Conf.\ & \% IDK \\
 \midrule
\texttt{P1} & AI + confidence & 17.0 & 76.0 & 28.1 & 3.8 & 3.8 & - \\
\texttt{P2} & \quad \rotatebox[origin=c]{180}{$\Lsh$} + IDK, role, record & 18.5 & 77.3 & 14.3 & 3.3 & 3.8 & 3.3 \\
\texttt{P3} & \quad \quad \rotatebox[origin=c]{180}{$\Lsh$} + feedback & 10.7 & 84.2 & 26.9 & 3.6 & 3.6 & 1.7 \\
\texttt{P4} & \quad \quad \quad \rotatebox[origin=c]{180}{$\Lsh$} +  highlights & 9.6 & 87.0 & 14.8 & - & 3.9 & 0.6\\
\texttt{P5} & \quad \quad \quad \rotatebox[origin=c]{180}{$\Lsh$} +  influential examples & 10.0 & 86.3 & 15.4 & - & 3.8 & 0.0\\
\bottomrule
\end{tabular}%
}
\caption{Human performance for ContractNLI claim verification across conditions in Study I (\sect{sec:study1_main}). $\mathcal{M}$ denotes the finetuned Flan-T5-3B, $\mathcal{H}$ human participants, \texttt{\#} Reveal for how many of 6 instances annotators reveal the model's prediction, $\mathcal{H}$ Conf.\ their self-confidence, and \% IDK the rate at which they chose the I don't know option.}
\label{tab:contractnli_results}
\end{table*}

\subsection{Study Design}
\label{sec:study1_design}

We overview various design choices we consider that we recommend integrating in future studies.  

To isolate the effects of explanations, in the first step, a participant may reveal the prediction and make their first guess. %
Only then we provide the explanation and ask for their final decision.\footnote{In Appendix \ref{sec:discussion_isolating_everything}, we discuss why we do not ask people to make three guesses to isolate the effects of the predictions. } %
To encourage thoughtful engagement with the model predictions in the first step, we use three \textbf{cognitive forcing functions} \cite{10.1145/3449287}. %
(1) Let participants choose whether to reveal the model's predictions. %
(2) Delay when the option to reveal the prediction becomes available.\footnote{We randomly select the delay time between 50--70 or 30--45 seconds, depending on the study. Randomly to avoid waiting out a fixed amount of time.} 
(3) Disable the option to move to the next step for a minimum amount of time considered necessary %
to reasonably attempt the task; in our case, for 90 seconds.

Making hypothetical decisions in a questionnaire differs from making real decisions, where people might reassess their trust and opt for more cautious actions, or not act at all. %
Consider claim verification. %
People may temporarily perceive false information to be true, but do not disseminate all such misconceptions. %
When considering making decisions based on them, one might reflect more deeply and consider their confidence more carefully. %

To bring participants \textbf{closer to real-world scenarios}, we have them adopt a specific \textbf{hypothetical role}, as shown in the example in Fig.\ \ref{fig:role}. %
Henceforth, all figures except Fig.\ \ref{fig:studyflow} appear in the Appendix. 
Additionally, they should have the \textbf{don't know option}, which allows them to refrain from making decisions when they normally would not, thereby avoiding an overestimation of overreliance. %
To prevent %
using this option unrealistically often, we show them ``their'' \textbf{prior record} that occasionally mentions they previously refrained from making judgments and that they are asked by their superior to decrease that rate; see Fig.\ \ref{fig:record}. %
Another method to discourage excessive use of the don't know option is to provide \textbf{feedback} after a decision is made; see Fig.\ \ref{fig:feedback}. %
Moreover, people typically receive some feedback when deciding for others,  which in turn influences their future decision making.%

We create roles, records, and feedback by prompting \texttt{gpt-4-0613} \cite{openai2024gpt4} and revising its generations. %
We also use it to create a simple multiple-choice question (MCQ) for each NDA serving as an \textbf{attention check} in the main annotation tasks. %
We provide details of these processes in Appendices  \ref{sec:roles_records_and_feedback} and \ref{sec:attention_check}.

Finally, in a small experiment, we check that we recruit annotators who can perform this task alone. 
We ask 5 participants to annotate 2 legal claims: one with a short NDA and one with a long. %
Note that in the main study (\sect{sec:study1_main}) we have $\approx$40 participants. 
The first claim should be accurately assessed given the short NDA, and all annotators meet our expectations. %
See details in Appendix \ref{sec:contractnli_experts}. %
For the claims with longer NDAs, 4/5 participants give the correct response. %
This suggests that annotators with a degree in administration and law who work in legal functions perform well on this task. %
Thus, they should be \textbf{time-constrained} when deciding for all instances; otherwise, the benefit of providing AI's predictions should not be expected.

\subsection{Task Presentations}
\label{sec:task_conditions}

Annotators' task is to assess whether a statement aligns with a given NDA. %
We test the following five task presentations, outlined in Fig.\ \ref{fig:studyflow}:
\begin{compactitem}
    \item \textbf{\texttt{(P1;Baseline)}} A delayed option to see Flan-T5's prediction and its \textbf{calibrated confidence}.\footnote{Temperature scaling \citep{pmlr-v70-guo17a} on the dev set.}
    \item \textbf{\texttt{(P2)}} \texttt{P1} information with the don't know option and priming with roles and records.
    \item \textbf{\texttt{(P3)}} \texttt{P2} information with providing feedback.
    \item \textbf{\texttt{(P4)}} \texttt{P2} information, then input highlights in the next step, and finally feedback.\footnote{\texttt{Input$\times$Gradient} highlights \cite{DBLP:journals/corr/ShrikumarGSK16} obtained with \href{https://github.com/inseq-team/inseq}{inseq} \cite{sarti-etal-2023-inseq}.} 
    \item \textbf{\texttt{(P5)}} \texttt{P2} information, followed by influential train examples, and at the end, feedback.\footnote{EK-FAC influences \cite{DBLP:conf/nips/GeorgeLBBV18,DBLP:journals/corr/abs-2308-03296} obtained with \href{https://github.com/pomonam/kronfluence/tree/main}{kronfluence}.}
\end{compactitem}
Figures \ref{fig:instructions123}--\ref{fig:instructions5} show the instructions given to the annotators.  %
In each task setting, participants are given 6 instances. %
For each, they first answer an attention-check MCQ about an NDA within 3 minutes. %
If time runs out, they are moved to the next instance.\footnote{Thus, the total num.\ of examples varies across conditions.} %
Following the MCQ, they should evaluate a statement based on the same NDA in 7 minutes (Fig.\ \ref{fig:statement_assess}) and then report their self-confidence (Fig.\ \ref{fig:confidence}). %
In the final two settings, after participants make the first guess, they are shown the model explanation and asked to guess again.
In \texttt{P4}, participants may see the NDA with top 5\%, 10\%, or 20\% of the important words highlighted (Fig.\ \ref{fig:highlight}), and in \texttt{P5}, the top 3 most influential labeled train examples with \texttt{Input$\times$Gradient} highlights incorporated\,---\,participants should not have to \emph{fully} review three different NDAs to \emph{quickly} verify if the AI's guess is correct. %
Appendix \ref{sec:highlight} provides more info.

For each study, we aim for 80 examples for each of the three labels. %
We gather 35-40 participant responses per study, with each participant annotating up to 6 samples.
We collect 1108 annotations across all conditions. %
Participant recruitment is done through \href{https://www.prolific.com/}{Prolific}. 
More on participants in \ref{sec:participation}.

\subsection{Results}
\label{sec:study1_results}

We provide our findings in Table \ref{tab:contractnli_results}. %
Across all conditions, time-constrained humans ($\mathcal{H}$) collaborating with the model perform at least 9.5 F1 score points worse than the model alone ($\mathcal{M}$). %
This underscores the need for major improvements in the effectiveness of this type of human-AI teams, regardless of explanations. %
We notice that the rate at which experts are correct when the model is also correct is far from 100\%. %
This hesitation among experts to rely on the model is also evident from their decision to reveal the model's prediction in just over half of the six provided examples.  %
\texttt{P3} (feedback) significantly improves this rate.\footnote{A two-proportion z-test yields $p$-value of 0.03 for \texttt{P3}.} %
Despite seemingly being more risk-averse, participants choose ``I don't know'' with a low rate of at most 3\%. %

The rate at which people are correct when the model is wrong never exceeds 28\%. %
When both the model and people are wrong, they concur in at least  80\% of such cases (not shown in Table \ref{tab:contractnli_results}). %
Future human-AI teaming should focus on strategies that could notably improve this. %
On average, people's confidence ranges 3--4 (moderately to highly confident), even when they make wrong decisions.

In Table \ref{tab:explanation_analysis} (Appendix), we present a breakdown of the impact of highlights. %
As evident by the number of different situations that arise from the people's first guess (1st col.), asking people to reasses their decisions and self-confidence with highlights helps better understand their effects. %
This analysis confirms that highlights do not consistently meet expectations to mitigate under- and overreliance; in some cases, they may even contribute to it. %

Our main takeaway is that the key challenge in this human-AI team setup is achieving that experts spend just enough time on an instance to develop an informed opinion and then effectively use the AI's guess, confidence, and explanation to \emph{quickly} transition from initial opinion to a confident, accurate decision. %
Only in this way can they speed up their work without sacrificing performance. %
However, this raises the question: what could help achieve this delicate balance? %
We suspect that adding more information to the AI's guess is not the solution. 

\section{Study II: People Decide Only for Deferred Instances} 
\label{sec:study2}
This 
 human-AI decision making setup 
defers a fraction of all instances to experts. %
We investigate explanation usefulness for finetuning/prompting LLMs to defer, and to human decision makers.

\subsection{Usefulness to Deferral Models} 

We finetune \texttt{Llama-2\--\-13B\--\-Chat} and \texttt{gpt\--\-3.5\--turbo-1106}, and prompt \texttt{gpt-4o-2024-05-13}. %
We use 1.4K model's dev-set predictions for finetuning, 371 predictions for testing deferral models, and 30 for in-context learning.\footnote{10 positive and 20 negative as fewer examples should be deferred to experts.} %
Wrong predictions are positive (defer) examples and correct ones are negative (don't defer) examples for deferral models. 
We provide all models with a ContractNLI claim+NDA to defer or not, and the model's label for the claim.\footnote{We exclude confidence because we find no correlation between it and prediction accuracy (see Fig.\ \ref{fig:conf_vs_predAcc}).} %
Some models also get a few demonstrations and/or \texttt{Input$\times$Gradient} highlights. 

The main challenge we observe with all models is a low recall for the deferral class, as seen in Table \ref{tab:deferral_models_performance}, with more results in Table \ref{tab:all_deferral_models_performance} (Appendix). %
Prompting \texttt{gpt-4o-2024-05-13} improves recall compared to the finetuned \texttt{Llama-2\--\-13B\--\-Chat}, if a few examples are provided. %
Finetuning \texttt{gpt\--\-3.5\--turbo-1106} notably improves recall, but remains low, reaching only 30\%. %
This shows that our best deferral model is not functional. %
Thus, building effective deferral models remains a challenge, and human-AI decision-making in this setup needs notable improvements, much like the previous setup. %

Including highlights (details on how we do that in Appendix~\ref{sec:deferral}) reduces recall, contrary to the belief they might provide additional signal about when the model is correct/wrong. %

\begin{table}[t]
\centering
\scriptsize
\resizebox{\columnwidth}{!}{%
\begin{tabular}{lllr}
\toprule
\textbf{Model} &\textbf{Train Setup} &\textbf{Input} & \textbf{Recall} \\
\midrule
\multirow{2}{*}{\texttt{Llama\--2\--Chat\--13B}} &\multirow{2}{*}{Finetuned} &P+T+H &5.0 \\
& &P+T &10.0 \\
\midrule
\multirow{3}{*}{\texttt{gpt\--4o\--2024\--05\--13}} & {Zero-shot} &P+T &10.0 \\
\arrayrulecolor{black!20}\cmidrule{2-4}
& \multirow{2}{*}{30-shot} &P+T &17.5  \\
& &P+T+H &15.0 \\
\arrayrulecolor{black}\midrule
\texttt{gpt\--3.5\--turbo\--1106} &Finetuned &P+T &\textbf{30.0} \\
\bottomrule
\end{tabular}
}
\caption{Recall of deferral models for the defer-to-expert (AI wrong) class (\sect{sec:study2}). P denotes the AI's prediction, T denotes ContractNLI instance text, and H highlights.}
\label{tab:deferral_models_performance}
\end{table}

\subsection{Usefulness to Experts}
We still aim to provide initial insights into whether explanations for likely-wrong predictions help decision makers, as we hypothesize in \sect{sec:background}. %
Without workable deferral models, we do not study this in a realistic setup with deferral mistakes; all participants get instances that the model mishandles. %
Given the lack of usefulness of explanations so far, we opt for a small-scale study and include a free-form question about how explanations are helpful, if at all. %
We use the setups in \texttt{P4} (w/ highlights) and \texttt{P5} (w/ highlighted influential examples) in \sect{sec:study1_main}, but modify the instructions to warn participants that AI most likely mislabeled examples they will review (see Fig.\ \ref{fig:instructions_defer}). 
We end with 23 participants for the former and 20 for the latter, each 
annotating 2 instances. %
The final number of annotated instances is 73 because some annotators reach the time limit. 

Participants, on average, rate the impact of explanations on their decision making with 2.56 (slightly to moderately). %
For 55/73 statements participants explain whether and how explanations are helpful. 
We categorize these insights in  Table~\ref{tab:free-form_response_codes} and provide examples. %
While explanations of a likely-wrong model can encourage different reasoning approaches as we hypothesize, highlights are more often helpful in finding relevant information in an NDA despite wrong predictions.\footnote{Another strong baseline could be offering ``model-free'' highlights based on the statement-NDA lexical overlap.} %
These results suggest that the model finds necessary information, but uses it in a way that leads it to a wrong decision. %
Thus, rather than suggesting to annotators that these explanations may reflect likely flawed reasoning, we should inform them that the model could identify the correct evidence yet still arrive at the wrong conclusion.

We notice a low F1 score of only 34.2\% in this study (Table~\ref{tab:deferral_results}, Appendix), which is a stark contrast to the previous small-scale study of people operating alone  (\sect{sec:study1_design}). %
Five participants state they disagree with feedback mentioning the correct label (see Appendix~\ref{sec:deferral}). %
Therefore, one author of this paper has checked 29 examples for which participants respond about their use of highlights.
The gold label is changed if the author's reasoning matches the participant's reasoning and they disagree with the gold label. %
This occurs in 11/29 cases, suggesting that examples that are challenging to the model might also have noisy labels. %
Table~\ref{tab:revised_gold_label} shows one such sample.
Using the revised gold labels, participants' F1 score increases from 34.9\% to 79.4\% on this reviewed subset (29 samples), and AI's F1 score from 0\% to 27.3\%. %
Future work should be cautious with deferred instances (where AI is wrong) because these likely present the tail-end distribution of the dataset. 

Finally, we evaluate the human-AI team with the default gold labels. 
237/2091 ContractNLI test instances (11.3\%) are mislabeled by the model and would be deferred to experts by a perfect deferral model. %
We assume that the experts' performance on all 237 deferred samples is similar to the 34.2\% F1 score they achieve on the 73 samples they annotated. %
88.7\% (AI correct) * 100\% (AI F1) + 11.3\% (AI wrong) * 34.2\% (experts F1) gives a joint performance of 92.6\% under the perfect deferral model. %
This is 13.9 points higher than the best team in \sect{sec:study1_main}, and would additionally increase with cleaner labels. %
Based on the average time needed for reviewing a statement-NDA pair in \sect{sec:study1_main} (212 seconds), we determine that the human decision makers' time needed for deferred instances (11.3\% of the data) would be 14 hours, while for all instances it would be 123 hours. %

These results suggest that achieving workable deferral is a promising way to advance human-AI teams. %
A new goal for explanations could be to help a deferral model to identify when the predictor is likely accurate. %
Moreover, given the challenges of %
the other human-AI team setup discussed in \sect{sec:study1_results}), we find the deferral setup more feasible. %
Here, the primary challenge is technical: developing accurate automatic deferral. %
Once this is done, domain experts can use their time to accurately assess a small set of instances, which we know they can do.

\section{Conclusions}

We provide guidelines for specifying a sound experimental setup for application-grounded evaluation of explanations in NLP. %
The dataset selection criteria we set should be used to determine a dataset's suitability for such evaluations. %
Future studies should consider 4 datasets we identify by validating over 50 datasets against our criteria, or validate/create new datasets guided by them.  %
We show that the performance of a resource-appropriate state-of-the-art model should be continuously reassessed: it should neither peak nor be low, otherwise teaming the model with people is unsound. %
We recommend following our user study design that isolates the effect of explanations, uses strong baseline conditions, deploys multiple cognitive forcing functions, integrates real-world situations, implements attention checks tailored to the application, among other things. %

We believe that the main reason AI assistance is unhelpful to specialists, regardless of explanations, is the difficulty of making the right decision without investing enough effort into a given problem. %
Yet, more effort makes teaming with AI unnecessary. %
We argue that integrating deferral could be a more viable approach to speeding up work without sacrificing overall performance. %

\section{Limitations}

The dataset selection criteria we set (\sect{sec:data_criteria}) could have been used to validate some promising datasets beyond the resource with existing explainability-research datasets we used. %
The application of the criteria could also be enhanced. %
For example, we approximate effort with length, but shorter examples people can solve could also require effort, e.g., certain math problems. %
On the other hand, finding an answer quickly in a longer task instance might be possible with a keyword search. %
Future work should continue to improve methods for quantifying human effort and ability. %
We subjectively determine hazard severity, and sometimes likelihood, which leaves room for disagreement. %
However, acknowledging this, we provide higher risk when less confident, such that a dataset is not unfairly filtered. %
While newer models than Flan-T5 could have been finetuned in \sect{sec:model_appropriateness}, and consequently not all of the 4 datasets might remain as suitable, the main takeaway of this section remains: reassess the performance of a model you plan to team people with. %
Finally, our user studies face challenges all user studies do. %
Specifically, they would be stronger with more participants, examples annotated, explanation types, and datasets evaluated, among other things, which is not possible due to financial restrictions. %
Despite our best efforts to provide reliable outcomes, as in most studies with human subjects, we cannot guarantee there are no confounders. 

\section*{Acknowledgments}
We thank anonymous reviewers for their useful feedback, Dana EeJae Ahn for her help in calculating the average length of instances in datasets we examine, Jason Wiese and Kyle Lo for helpful discussions on the study procedure, Kyle Lo for his feedback on the framing of this work, Nazanin Hashemi Chaleshtori for her assistance with understanding examples in medical domain, 
UtahNLP group, everyone who responded to our requests to share the data, and Q.\ Vera Liao and Alison Smith-Renner whose presentation at the NAACL'22 tutorial on ``Human-centered Evaluations of Explanations'' helped us write the background section. 

\bibliography{anthology,custom}

\clearpage
\appendix

\begin{table*}[t]
\centering
\resizebox{\textwidth}{!}{%
\begin{tabular}{llp{4.3cm}p{4.5cm}p{3.6cm}p{3.1cm}c}
\toprule
& \textbf{Task} & \textbf{Usefulness Measurements} & \textbf{Explanations Evaluated} & \textbf{Models Explained} & \textbf{Baseline Condition} & \textbf{Helpful} \\
 \midrule
\hypertarget{work1}{W1} & \hyperlink{dataset1}{D1} & Increased team performance & Input attribution from SVM's weights; k-NN train examples  &  Linear SVM with BoW
features  & Providing the accuracy of the SVM & \color{darkgreen}{\textbf{Yes}}  \\
 \midrule
\hypertarget{work2}{W2} & \hyperlink{dataset2}{D2} & Regression analysis est.\ how each condition influences player accuracy & Manually extracted evidence$\ddagger$ & TF-IDF to find \& return the label of the most similar doc or pre\-vi\-ously seen question & Similarity score between a question \& a retrieved doc & \color{darkgreen}{\textbf{Yes}} \\
 \midrule
 \hypertarget{work3}{W3} & \makecell[lt]{\hyperlink{dataset3}{D3};\\\hyperlink{dataset4}{D4}} & Increased team performance & LIME input attribution of top-1 or 2 predictions or \emph{human} free-text explanations & RoBERTA-Base (FT) & Post-hoc calibrated model confidence  & \color{red}{\textbf{No}} \\
 \midrule
\hypertarget{work4}{W4}    & \hyperlink{dataset5}{D5} & Increase in \% accept correct; Decrease in \% accept wrong  & Manually extracted evidence independent of the model$^\dagger$ & DPR~\citep{karpukhin-etal-2020-dense} & Post-hoc calibrated model confidence & \color{darkgreen}{\textbf{Yes}} \\
 \midrule
\hypertarget{work5}{W5} &\makecell[lt]{\hyperlink{dataset6}{D6};\\\hyperlink{dataset7}{D7}} & \% initially  wrong, correct after the explanation & Free-text explanations & T5-large (FT-full); T5-3B (FT-128); \texttt{davinci-instruct-beta} (ICL-6) & None &{\color{red}\textbf{No}}\\
\midrule 
\hypertarget{work6}{W6}  & \hyperlink{dataset6}{D6} & \% initially wrong, flip after the model's guess; \% stick with initial correct guess   &  LIME input attribution &  SVM & None & \color{darkgreen}{\textbf{Yes}} \\
\midrule
\hypertarget{work7}{W7}  & \hyperlink{dataset8}{D8} & Increased team performance Decreased task time & Free-text explanations  & Flan-T5-XLarge & None & \color{red}{\textbf{No}} \\
\midrule
\hypertarget{work8}{W8} & \hyperlink{dataset9}{D9} & Increased team performance Decreased task time & Free-text explanation; Contrastive explanations; Retrieved passages & ChatGPT + %
GTR-XXL \cite{ni-etal-2022-large} retriever & None & \color{red}{\textbf{No}} \\ 
\bottomrule
\end{tabular}%
}
\caption{Overview of prior application-grounded explanation usefulness evaluations involving a text-based task. 
\protect\hypertarget{work1}{W1}  \citep{10.1145/3287560.3287590}; 
\protect\hypertarget{work2}{W2} \citep{DBLP:conf/iui/FengB19}; 
\protect\hypertarget{work3}{W3} \citep{10.1145/3411764.3445717}; 
\protect\hypertarget{work4}{W4} \citep{gonzalez-etal-2021-explanations}; 
\protect\hypertarget{work5}{W5} \citep{joshi-etal-2023-machine}; 
\protect\hypertarget{work6}{W6} \citep{DBLP:conf/iui/SchemmerKBBS23};  \protect\hypertarget{work7}{W7} \citep{DBLP:conf/nips/MozannarLWSDS23};  \protect\hypertarget{work8}{W8} \citep{DBLP:journals/corr/abs-2310-12558}. 
FT stands for ``finetuned'' and ICL for ``in-context learning''. %
}
\label{tab:overview-prior-work}
\end{table*}

\section*{Appendix Overview}

In this supplementary material, we provide: 

\begin{compactitem}
\item \textbf{Appendix \ref{sec:appendix_background}.} Discussion on the lack of explanation evaluations in NLP that are application-grounded.
\item \textbf{Appendix \ref{sec:dataset_quality}.} Additional analyses of the quality and suitability of three datasets highlighted by our meta-analysis.    
\item \textbf{Appendix \ref{baseline_appendix}}. Details about the data processing and training of Flan-T5-3B on four datasets.  
\item \textbf{Appendix \ref{sec:experiments_design}.} 
Information on various components of our human studies and additional results complementing Study I (\sect{sec:study1_main}).
\item \textbf{Appendix \ref{sec:deferral}.} Details on preparing the input data for deferral (\sect{sec:study2}) and participants' feedback on the study's verdict for instances they annotate.
\item \textbf{Appendix \ref{sec:exnlp_task_cards}.} Details of our reasoning behind the categorization of each dataset.
\end{compactitem}

\begin{table*}[!ht]
    \small
    \centering
    \begin{adjustbox}{valign=t}
    \begin{subtable}[t]{.38\linewidth}
        \small
        \centering
        \resizebox{\linewidth}{!}{
        \begin{tabular}{lrrr}\toprule
        &\textbf{Precision} &\textbf{Recall} &\textbf{F1 score}\\\midrule
        \textsc{Support}        &42.3   &100.0    &59.4 \\
        \textsc{No-Info.}       &80.0     &1.6    &3.1 \\
        \textsc{Contradict}     &0.0      &0.0      &0.0 \\
        \arrayrulecolor{black!20}\midrule
        \textsc{Micro Avg.}     &40.8   &33.9   &20.9 \\
        \textsc{Macro Avg.}     &52.4   &42.7   &26.3 \\
        \arrayrulecolor{black}\bottomrule
        \end{tabular}
        }
        \caption{Using automatically retrieved articles}
        \label{tab:snopes_retrieved_nonfact}
    \end{subtable}%
    \end{adjustbox}
    \hspace{3mm}
    \begin{adjustbox}{valign=t}
    \begin{subtable}[t]{.255\linewidth}
        \small
        \centering
        \resizebox{\linewidth}{!}{
        \begin{tabular}{rrr}\toprule
        \textbf{Precision} &\textbf{Recall} &\textbf{F1 score}\\\midrule
        44.0      &84.3   &57.8 \\
        49.6    &21.2   &29.7 \\
        0.0       &0.0      &0.0 \\
        \arrayrulecolor{black!20}\midrule
        31.2    &35.2   &29.2 \\
        40.0      &44.6   &37.2 \\
        \arrayrulecolor{black}\bottomrule
        \end{tabular}
        }
        \caption{Using articles linked in an associated Snopes fact-checking report}
        \label{tab:snopes_gold_nonfact}
    \end{subtable}%
    \end{adjustbox}
    \hspace{3mm}
    \begin{adjustbox}{valign=t}
    \begin{subtable}[t]{.255\linewidth}
        \small
        \centering
        \resizebox{\linewidth}{!}{
        \begin{tabular}{rrr}\toprule
        \textbf{Precision} &\textbf{Recall} &\textbf{F1 score}\\\midrule
        81.6    &75.3   &78.3 \\
        76.0      &82.6   &79.1 \\
        67.1    &62.82  &64.9 \\
        \arrayrulecolor{black!20}\midrule
        74.9    &73.6   &\textbf{74.1} \\
        79.7    &77.8   &78.6 \\
        \arrayrulecolor{black}\bottomrule
        \end{tabular}
        }
        \caption{Using a Snopes fact-checking report}
        \label{tab:snopes_gold}
    \end{subtable}%
    \end{adjustbox}
    \caption{Finetuned Flan-T5-3B results for claim verification in the UKPSnopes dataset across three input setups: with automatically retrieved articles, with articles linked in a fact-checking report for the claim, and with a fact-checking report for the claim.}
    \label{tab:snopes_experiments}
\end{table*}

\section{Discussion: Scarcity of Application-Grounded Explanation Evaluations in NLP}
\label{sec:appendix_background}

Why are application-grounded evaluations of explanations currently limited in NLP? 
Such evaluations of explanations have predominantly been done for applications with interpretable features such as people's age or income  \citep{liao2022humancentered}. %
Explaining tasks that involve text has unique challenges: features are a sequence of high-dimensional non-interpretable vectors; an arbitrary number of features; continuous representations of discrete inputs; explaining models with billions of parameters; pretrained models; and LLMs outperform inherently interpretable models (e.g., linear models, short decision trees). %
Prior NLP explainability work has mostly focused on overcoming these challenges. %
Moreover, many realistic language technology applications have become evident and possible with LLMs in the past two years. 

\section{Quality Analysis of Filtered Datasets}
\label{sec:dataset_quality}

In this section, we explain why we exclude three datasets, although they meet the necessary criteria for studying the impact of explanations on human-AI teams.

\subsection{Analysis of \textsc{LIAR-RAW} (\protect\hyperlink{dataset17}{D17}) and \textsc{RAWFC} (\protect\hyperlink{dataset18}{D18})}
\label{sec:liar_raw_analysis}

The goal of this task is to assess the veracity of statements about a diverse range of topics, using a handful of reports as references. %
A few issues related to data quality became apparent after conducting a manual examination of a randomly selected sample from the dataset. %
Notably, it appears that perhaps during data processing stages, all instances of ``to be'' verbs are replaced with ``be'', sentences and phrases are truncated, unnecessary repetitions of sentences, and other grammatical problems are identified. %
We evaluate the quality of the data by analyzing a sample of 100 data points randomly selected from the dataset. 

A single author carefully reviews each data point to determine its acceptability based on the claim and the accompanying reports extracted from the relevant articles. %
This assessment encompassed confirming the coherence and alignment of the utterances with the claim. %
Out of the 100 data points reviewed, 38 are deemed acceptable.\footnote{We additionally calculate the perplexity for these data points using the \texttt{gpt-2-XL} (1.5B) model \citep{radford2019language}. %
We could use perplexity for data filtering if it is correlated with data acceptability. %
We get a low correlation between them.
}
We find similar issues with \textsc{RAWFC} (\protect\hyperlink{dataset18}{D18}). 
We thus exclude these datasets.

\subsection{Analysis of \textsc{UKPSnopes} (\protect\hyperlink{dataset42}{D42})}
\label{sec:ukp_snopes_analysis}
In this task, the objective is to evaluate the veracity of claims across various domains. Each claim is accompanied by a fact-checking report sourced from the fact-checking website, Snopes\footnote{\url{https://www.snopes.com}}. This configuration, however, does not mirror real-world situations since a fact-checking report may not always be available for every new claim. 
Instead, one needs to retrieve relevant documents. 
\citet{hanselowski-etal-2019-richly} note that relevant articles can be obtained from links within the Snopes fact-checking reports. %
To this end, we compile all linked articles into a corpus of size 13K. %
We retrieve the most relevant documents from this corpus for each claim and then finetune the Flan-T5-3B model for claim verification using this information.\footnote{Each claim in the dataset is matched with multiple snippets from the associated Snopes article, with each (claim, snippet) pair receiving a fact-checking label. 
We observe that approximately 15.2\% of claims in the training set have different inconsistent labels when matched with different snippets of the same Snopes article. 
We exclude these claims due to the uncertainty in determining the gold label in the absence of the accompanying Snopes report and pairing them with other gathered articles.}
For finetuning details, see how we approach this for other retrieval tasks in Appendix \ref{baseline_appendix}. 

Our retrieval model achieves a recall of 51\% for claims in the training set. %
The micro-F1 score for claim verification using Flan-T5-3B finetuned with retrieved documents is only 20.9 (Table~\ref{tab:snopes_retrieved_nonfact}). %
We conduct additional experiments to investigate what is responsible for this: the compiled corpus of documents for retrieval is insufficient and therefore the dataset is not suitable, or retrieval is hard for this task. %
We omit the retrieval step and finetune the Flan-T5-3B model using (1) the gold articles linked within the Snopes fact-checking report about the claim, and (2) the gold Snopes fact-checking reports. %
We get the micro-F1 score of only 29.2 with the articles linked in the fact-checking report (Table~\ref{tab:snopes_gold_nonfact}), but 74.1 F1 with the report (Table~\ref{tab:snopes_gold}). 
This indicates that the corpus of linked articles lacks the necessary information to address the task effectively, contrary to \citet{hanselowski-etal-2019-richly} hypothesis.

\section{Details of Model Finetuning}
\label{baseline_appendix}
\begin{table}[t]
    \small
    \centering
    \begin{subtable}[t]{0.85\columnwidth}
        \centering
        \begin{tabular}{lrrr}\toprule
        &\textbf{Precision} &\textbf{Recall} &\textbf{F1 score}\\\midrule
        \multicolumn{4}{l}{\textbf{Flan-T5-3B (our)}} \\
        \arrayrulecolor{black!20}\midrule
        \textsc{Accept}     &81.7 &72.6 &76.9  \\
        \textsc{Reject}     &75.1 &83.6 &79.1 \\
        \arrayrulecolor{black!20}\midrule
        \textsc{Micro Avg.} &78.4 &78.0 &78.0 \\
        \textsc{Macro Avg.} &78.4 &78.1 & \textbf{78.0}  \\
        \arrayrulecolor{black}\midrule
        \multicolumn{4}{l}{\textbf{XLNet + BiGRU}} \\ %
        \arrayrulecolor{black!20}\midrule
        \textsc{Macro Avg.} &76.8 &76.3 &76.5 \\
        \multicolumn{3}{l}{\textsc{Human Est. Accuracy}} & 93.9\\
        \arrayrulecolor{black}\bottomrule
        \end{tabular}
        \caption{ILDC} 
        \label{tab:ildc_results_all}
    \end{subtable}%
    \vspace{3mm} %
    \begin{subtable}[t]{0.85\columnwidth}
        \centering
        \begin{tabular}{lrrr}\toprule
        &\textbf{Precision} &\textbf{Recall} &\textbf{F1 score}\\\midrule
        \multicolumn{4}{l}{\textbf{Flan-T5-3B (our)}} \\
        \arrayrulecolor{black!20}\midrule
        \textsc{Support}        &79.0	 &79.0	 &79.0 \\
        \textsc{No-Info}     &71.8	 &70.5	 &71.2 \\ 
        \textsc{Contradict}     &74.2	 &76.6	 &75.4 \\
        \arrayrulecolor{black!20}\midrule
        \textsc{Micro Avg.}     &75.3	 &75.3	 &75.3 \\
        \textsc{Macro Avg.}     &75.0    &75.4	 &\textbf{75.2} \\
        \arrayrulecolor{black}\midrule
        \multicolumn{4}{l}{\textsc{\textbf{MultiVerS}}}\\ %
        \arrayrulecolor{black!20}\midrule
        \textsc{Macro Avg.} &73.6 &40.7 &52.4\\
        \textsc{Human Est.} & 94.8 & 84.1 & 89.1 \\
        \arrayrulecolor{black}\bottomrule
        \end{tabular}
        \caption{SciFact-Open}
        \label{tab:scifact_results_all}
    \end{subtable}%
    \caption{Finetuned Flan-T5-3B and the state-of-the-art reported results. XLNet+BiGRU \citep{malik-etal-2021-ildc}; MultiVerS \citep{wadden-etal-2022-multivers}. \citet{wadden-etal-2022-multivers} estimate the human performance in the ``abstract-provided'' setting. ContractNLI and EvidenceInference v2 (w/ retrieval) results are in Table \ref{tab:all_results}. }
    \label{tab:results_ildc_scifact_appendix}
\end{table}

In Tables \ref{tab:scifactopen_example}--\ref{tab:evidence_infer_example}, we provide illustrative instances demonstrating how we craft the input for each baseline model we develop following the recommended templates.\footnote{\url{https://github.com/google-research/FLAN/blob/main/flan/v2/templates.py}} 
Below, we describe the process of finetuning Flan-T5-3B for each of the four datasets highlighted in our meta-analysis.

\paragraph{EvidenceInference v2 (\protect\hyperlink{dataset38}{D38})}
This task aims to compare the effect of treatment A relative to treatment B on a specified outcome within a scientific article.
In a real-world scenario, the ideal scientific articles to look into might not always be readily available. 
Hence, we formulate the task to involve document retrieval, and thereby, we aggregate all articles within the dataset to establish a corpus of articles. 
Our approach consists of these two steps:
\begin{compactitem}
    \item Use the BM25Plus algorithm \citep{DBLP:conf/adcs/TrotmanPB14} to get the top 100 relevant documents for each query, after which rerank those 100 
    with the method introduced by \citet{nogueira-etal-2020-document}, and finally select the top 10. 
    \item Finetune Flan-T5-3B using the query and the top 10 documents obtained as input.
\end{compactitem}
\noindent 
Our retrieval module has a low recall rate of  3\%, i.e., it retrieves the true relevant document for only 3\% of the queries.
Note that we use the same retrieval procedure for other datasets, and we get a recall score as high as 51\%. %
Table \ref{tab:ev_infer_results_all} presents the results of finetuning the Flan-T5-3B model with retrieved documents, and Table~\ref{tab:baseline_ev_infer_gold} shows how well our model performs when finetuned with the true relevant document instead of retrieved ones. %
The significant difference in F1 scores between using the true relevant documents and the retrieved documents 
underscores the retrieval challenge, indicating the need for %
stronger retrieval models. %

\begin{table}[t]
\small
\centering
\begin{tabular}{lrrrr}\toprule
 &\textbf{Precision} &\textbf{Recall} &\textbf{F1 score}\\\midrule
\textsc{Increase} &87.8 &91.3 &89.5 \\
\textsc{No-Diff.} &90.8 &87.8 &89.3 \\
\textsc{Decrease} &87.5 &87.9 &87.7 \\
\midrule
\textsc{Micro avg.} &89.0 &89.0 &\textbf{89.0} \\
\textsc{Micro avg.} &88.7 &89.0 &88.8 \\
\bottomrule
\end{tabular}
\caption{\textsc{ERASER EvidenceInference} task performance with finetuned Flan-T5-3B when gold documents are provided to the model.}\label{tab:baseline_ev_infer_gold}
\end{table}

\paragraph{SciFact-Open (\protect\hyperlink{dataset44}{D44})} 
This is another fact-checking task, but the claims are limited to the scientific domain.
To finetune the Flan-T5-3B model, we follow the same two-step approach described in the previous task.
We extract the top 10 most pertinent documents related to each claim from a corpus of 500K research abstracts.
See Table \ref{tab:scifact_results_all} for results.

\paragraph{ContractNLI (\protect\hyperlink{dataset49}{D49})}
Given a contract and a set of legal statements, the objective is to determine whether each statement implies, contradicts, or remains neutral in relation to the contract. %
This is a three-class classification task, with \quotes{Yes} signifying the statement entailment to the contract, \quotes{No} denoting contradiction with the contract, and \quotes{Cannot say} indicating the statement is undiscussed within the contract. %
To prepare data for finetuning, we concatenate the contract and statement (example in Table~\ref{tab:contract_nli_example}), ensuring the statement remains in the input by truncating the left side.
Table \ref{tab:contractnli_results_all} presents the model performance.

\paragraph{Indian Legal Documents Corpus (\protect\hyperlink{dataset50}{D50})}
This task involves predicting whether claims presented by an appellant/petitioner against a respondent should be accepted or rejected using a case proceeding document sourced from the Supreme Court of India \citep{malik-etal-2021-ildc}. %
Following the proposed approach accompanying the dataset, we use as many final tokens of   ILDC$_{\text{single}}$ instances as we can for training our model. %
The later tokens are expected to encapsulate the key information and reasoning underpinning the judgment. %
\citet{malik-etal-2021-ildc} could fit only 512 tokens, but Flan-T5 does not have restrictions on the input size. %
The number of input tokens it can process is determined by memory capacity; hence, we could fit 4200 tokens. 
Results are presented in Table \ref{tab:ev_infer_results_all}.

\section{Additional Details and Results of ContractNLI (\protect\hyperlink{dataset49}{D49}) User Study I (\sect{sec:study1_main})}
\label{sec:experiments_design}

In this section, we provide additional details that complete \sect{sec:study1_main} and \sect{sec:study2}. 
We design our studies using the Qualtrics online survey maker.\footnote{\url{https://www.qualtrics.com/}} 
In Table \ref{tab:explanation_analysis}, we provide a breakdown of the impact of highlights that is discussed in \ref{sec:study1_results}.

\begin{table}[t]
\centering
\resizebox{\columnwidth}{!}{%
\begin{tabular}{p{1.5cm}p{1.05cm}p{1.6cm}p{4.1cm}}
\toprule
$\mathcal{M}$-$\mathcal{H}_1$ & $\mathcal{H}_2$ & \texttt{\#} Conf.\ $\Delta$ & Effect of $h$ \\
\midrule
\multirow{3}{*}{$c$-$w$ [46]} & $c$ [11] & \makecell[c]{-} %
& \textcolor{darkgreen}{corrects underreliance}  \\
\arrayrulecolor{black!20}\cmidrule{2-4}
 & \multirow{3}{*}{$w$ [36]} & 4 $\uparrow$ %
 & \textcolor{red}{reinforces underreliance} \\
 & & 5 $\downarrow$ %
 & \textcolor{darkgreen}{discourages underreliance} \\
 & & 27* & no effect \\
\arrayrulecolor{black} \midrule
\multirow{3}{*}{$w$-$c$ [8]} & \multirow{2}{*}{$c$ [5]} & 1 $\downarrow$ %
& \textcolor{red}{discourages self-reliance}  \\
& & 4* & no effect \\
\arrayrulecolor{black!20}\cmidrule{2-4}
 & \multirow{2}{*}{$w$ [3]} & \multirow{2}{*}{\centering -}%
 & \multirow{2}{*}{\textcolor{red}{causes overreliance}} \\
 & & %
 & %
 \\
\arrayrulecolor{black} \midrule
$c$-$c$ [323] &  $w$ [15] & \makecell[c]{-}%
& \textcolor{red}{causes underreliance}\\
 \bottomrule
\end{tabular}%
}
\caption{Effects of highlights ($h$). $c$ / $w$ stand for correct/wrong decisions. $\mathcal{M}$-$\mathcal{H}_1$ shows the correctness of a model prediction and the 1st human guess; $\mathcal{H}_2$ the correctness of the 2nd guess upon seeing $h$. $\uparrow$ / $\downarrow$ / * means the human confidence increased / decreased / stayed the same after seeing $h$. The numbers in brackets show counts for each scenario. %
}
\label{tab:explanation_analysis}
\end{table}

\subsection{On Isolating Effects of Both AI's Predictions and Explanations} 
\label{sec:discussion_isolating_everything}

\citet{DBLP:conf/iui/SchemmerKBBS23}'s two-step approach to measuring reliance (\sect{sec:background}) accurately isolates the impact of model predictions on final human decisions. %
However, applying this approach to tasks identified by our meta-analysis requires extra considerations. 

The users of models trained for these tasks are experts, not laypeople,  as task instances are highly specialized (see examples in Tables \ref{tab:scifactopen_example}--\ref{tab:evidence_infer_example} such as the SciFact statement ``\emph{A high microerythrocyte count raises vulnerability to severe anemia in homozygous alpha (+)- thalassemia trait subjects.}''). %
These tasks require notable effort, but experts are skilled (e.g., ILDC experts' average accuracy is 94\%). %
Thus, if experts are asked to make the initial guess without time constraints, human-AI teams likely will not outdo experts alone, making human-AI teaming unwarranted. %
However, if time constraints are imposed then AI can be helpful by assisting experts to make accurate decisions more quickly and confidently, i.e., teaming makes sense. %
Therefore, unlike almost all prior studies in NLP that involve only laypeople, application-grounded evaluation with our highlighted tasks should focus on \emph{time-constrained} decision-making by experts. Experts are more expensive, so the number of instances and participants may be different from studies with laypeople. %

\citet{DBLP:conf/iui/SchemmerKBBS23}'s robust protocol should be further extended (besides displaying model confidence) by dividing the second step as follows: reveal the prediction, have the participant reassess, provide the explanation, and ask for the final decision.  %
This approach isolates the effects of explanations; e.g., if participants switch to wrong AI predictions despite making correct guesses initially, this might be because they are blindly following the AI's advice while ignoring the explanation. %
If these participants persist with the wrong AI prediction but their self-confidence lowers upon receiving expla\-na\-ti\-ons, it suggests that explanations may be discouraging overreliance. 
The breakdown of all possibilities is in Table \ref{tab:schammer_extra}. %
However, by adding this step, annotators need to make three guesses for the same examples, which is burdensome and makes it less likely that they will change their decision by the third guess. %
For evaluating explanation usefulness, we believe it is more important to reliably isolate the effect of showing the explanation. %
We thus recommend first asking annotators to make a guess, giving them an option to reveal the model's prediction, and then asking them to guess again upon seeing the explanation.

\subsection{Generating Roles, Records, and Feedback}
\label{sec:roles_records_and_feedback}

We aim to encourage participants to envision themselves in a realistic scenario, shift away from a typical crowdsourcing task, and approach it as they would in real life. %
To this end, we investigate the effect of presenting the participants with a hypothetical role and the consequences of ``their'' past errors in that role. %
Here, we explain how each piece of the aforementioned information is generated.

\paragraph{Roles (Condition \texttt{P2}).}

\begin{figure}[h]
    \centering
    \includegraphics[width=0.47\textwidth]
    {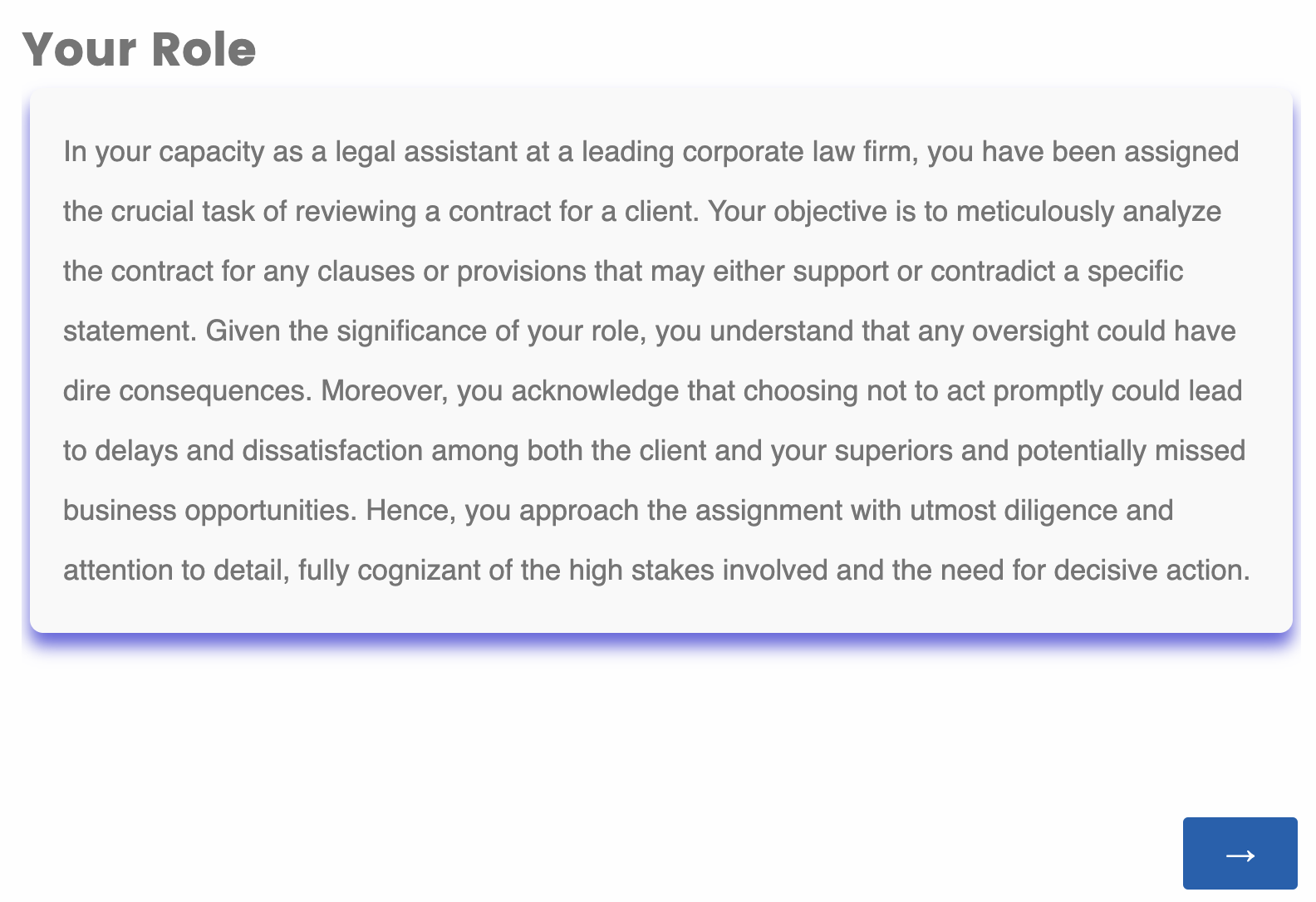}
    \caption{An example of the hypothetical roles we provide participants.}
    \label{fig:role}
\end{figure}

The role is exemplified by scenarios depicting a paralegal assistant's tasks within a corporate law firm, particularly involving work on NDAs. 
We add cautionary notes regarding the potential negative consequences of both errors and inaction.
We request \texttt{gpt-3.5-turbo} to refine and expand a manually crafted draft of the described scenario and warnings using more formal and technical language. 
Subsequently, we instruct \texttt{gpt-3.5-turbo} to reword the text, resulting in four variations.
Fig. \ref{fig:role} shows an instance.

\paragraph{Record (Condition \texttt{P2}).}
We construct the record information by integrating three components:

\begin{compactitem}
    \item We simply mention that the participant has prior errors. We use \texttt{gpt-3.5-turbo} to articulate this segment and generate four alternative phrasings to avoid redundancy.
    \item Providing one of the hypothetical situations described above to \texttt{gpt-4-0613}, we ask it to generate a list of potential consequences of making mistakes in this context. We end up with fourteen different consequences. Here is an example:

{\small
\texttt{Subjecting the client to legal liabilities and the possibility of facing lawsuits, increasing their financial burdens and legal complications.}
}
We present one or two randomly chosen consequences from the aforementioned list of 14, allocating two consequences for 50\% of the samples and one consequence for the remainder.%

\item We underscore the importance of avoiding indecision\,---\,equivalent to selecting the don't know option\,---\,as a common practice since it is deemed unacceptable. 
We randomly select a percentage between 20 and 35 to represent the frequency with which the participant has abstained from making decisions in the past, resulting in reprimands from their supervisor.
This last part is added to 40\% of the samples.
\end{compactitem}

\begin{figure}[h]
    \centering
    \includegraphics[width=0.47\textwidth]{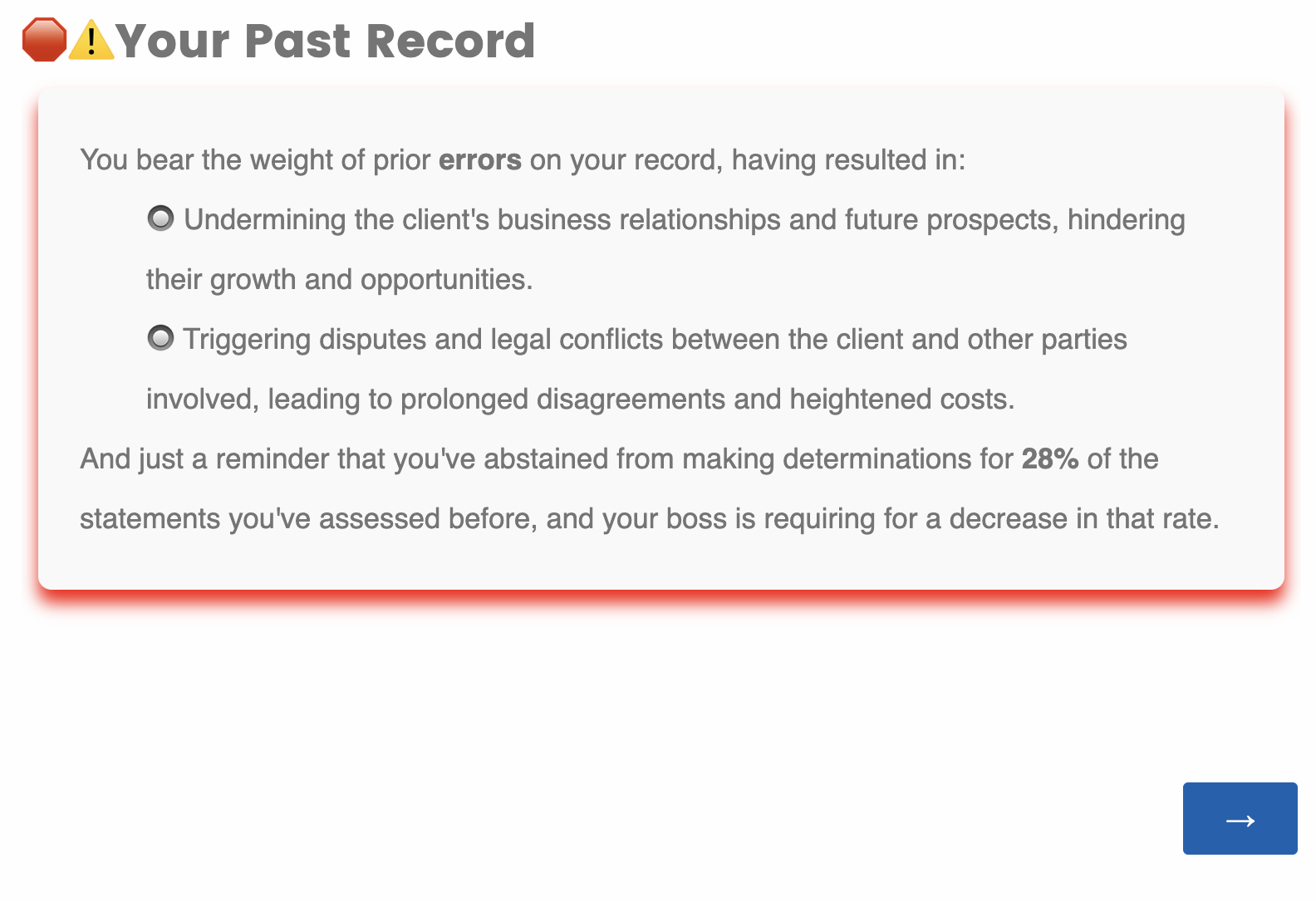}
    \caption{An example of the consequences of past errors we show to participants.}
    \label{fig:record}
\end{figure}

In Fig. \ref{fig:record}, we show an example containing two consequences and an inaction warning.

\paragraph{Feedback (Condition \texttt{P3}).}
We generate feedback based solely on the statement. %
Given three labels, there are 9 potential combinations of the statement's true label and what a participant might decide the label is. %
For each of the 17 unique statements in the ContractNLI dataset, we create feedback corresponding to each of the nine label combinations, resulting in a total of 153 unique feedback instances.
To generate feedback, we prompt \texttt{gpt-4-0613} in multiple iterations. 
First, we provide each of the 17 statements paired with each of the 3 true labels.
We ask \texttt{gpt-4-0613} to imagine it is a client who owns a company entering into a non-disclosure agreement with another company, and ask it to generate actions the client should take to adhere to the provided pair of statement and label. %
Table \ref{tab:gpt4_action_generation} illustrates one example per gold label. %
In the next iteration, we provide \texttt{gpt-4-0613} with the triplet of the statement, each of the three possible true labels, and each of the three labels a participant might choose, along with the actions provided in the previous step. We inform \texttt{gpt-4-0613} whether the actions were fulfilled or violated based on the match between the true label and the participant label, and we ask it to suggest a legal penalty or potential benefit based on the participant's mistakes or sound judgment. 
Additionally, we use \texttt{gpt-4-0613} to rephrase feedback according to the frequency of (statement, true label) pairs in the data.
Table \ref{tab:gpt4_feedback_generation} provides two examples of instructing \texttt{gpt-4-0613} to generate positive (for entail--entail combination) and negative (for undiscussed--entail combination) feedback. 
For other combinations, we slightly adjust the prompt to reflect the response accuracy compared to the true label. 

We use a generic declaration unrelated to task data to provide feedback on choosing the don't know option.
Again, we present a draft to \texttt{gpt-4-0613}, tasking it with elaborating and producing five distinct versions. 
For instance:

{\small
\texttt{
Your response did not adequately attend to your client's concerns. 
The lack of clarity is causing delays within the company and adversely affecting its business operations.
}
}

\begin{figure}[h]
    \centering
    \includegraphics[width=0.47\textwidth]{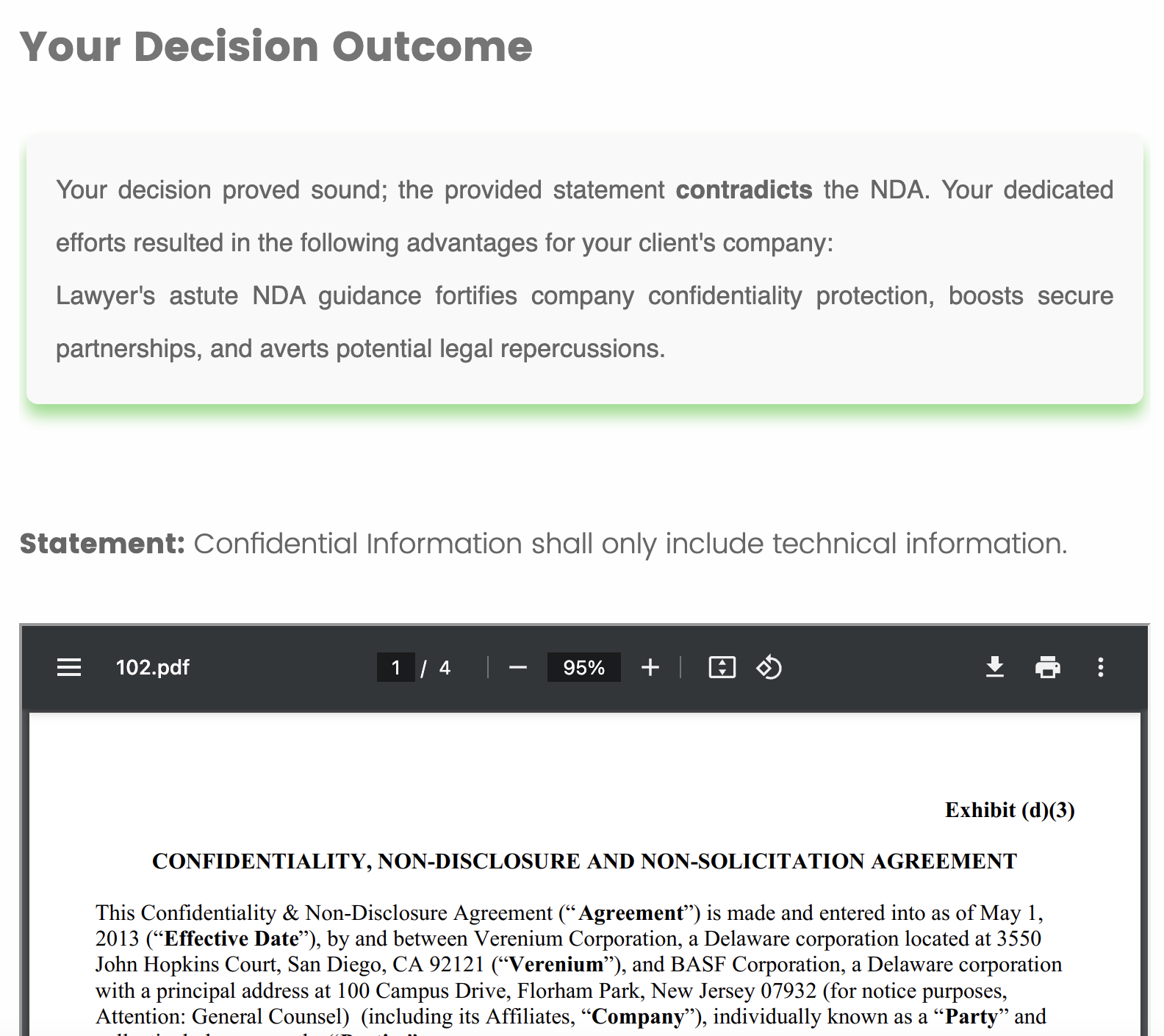}
    \caption{Positive feedback for labeling a specific statement correctly. We provide the statement and NDA again in this step for the participant's review.}
    \label{fig:feedback}
\end{figure}

An example of the feedback we present to the participants is shown in Fig. \ref{fig:feedback}.

\subsection{Constructing Qualification Exam and Attention Checks}
\label{sec:attention_check}

To ensure participant engagement and prevent spam, we create a qualification exam.
We create 30 multiple-choice questions (MCQs) based on 30 distinct NDAs from the training dataset; this prevents qualified participants from encountering NDAs used in the main tasks since they are picked from the test set. %
Below is the prompt we use to instruct \texttt{gpt-4-0613} to generate an MCQ based on a text excerpt:

{\small
\texttt{You are a helpful assistant specializing in the legal domain.
You want to check whether someone read a text. Construct a simple 4-choice question about this text so that anyone who reads the text is able to answer it. The correct answer must be explicitly mentioned in the text. Make sure only one answer is correct. Define the correct answer at the end. Try to balance the options in length (number of words). Format your question as follows:}
}

{\small
\texttt{<Question text>}

\texttt{1) option 1}

\texttt{2) option 2}

\texttt{3) option 3}

\texttt{4) option 4}

\texttt{Correct answer: [option\_number]}
}

\begin{figure}[h]
    \centering
    \includegraphics[width=0.47\textwidth]{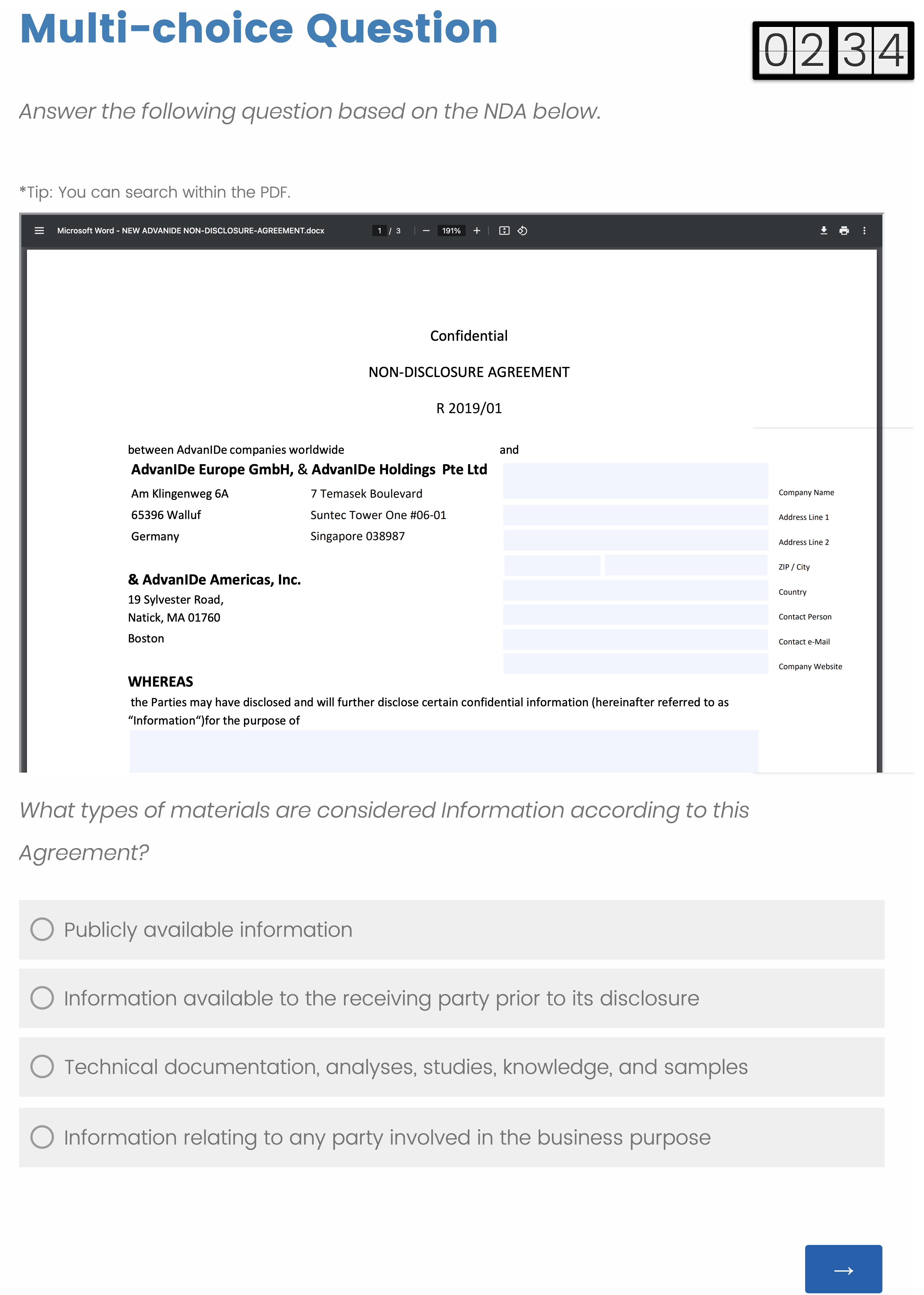}
    \caption{Attention check sample. Before solving the main task, participants are asked to solve a multi-choice question regarding the same NDA.}
    \label{fig:mcq}
\end{figure}

\noindent To verify its generation accuracy, we present the same text excerpt along with the question and options to \texttt{gpt-3.5-turbo} and check if the generated answer matches the response from \texttt{gpt-4-0613}. %
If they align, we assume the question is valid. %
However, in cases where the text contains negation or exclusions, the generated questions are generally inaccurate, requiring manual verification or replacement. %
From this pool of 30 questions, each participant is presented with 6 randomly chosen questions and the corresponding NDA. 
Passing the exam requires correctly answering at least 5 of these 6 questions. %
Around 74\% of participants passed the 10-minute exam. 
All participants receive \$2 compensation.
We compile a list of verified users and employ it in Prolific to regulate participation in the main studies.
Fig. \ref{fig:mcq} shows an example.

We use the same method to generate MCQs for each sample assessed by participants in the main human studies. Participants must answer the MCQ correctly to proceed to the next step. 
We monitor the number of attempts as an indicator of participants' attention. The average is 1.2 attempts.

\subsection{Estimating Expert Performance on ContractNLI} 
\label{sec:contractnli_experts}

\begin{figure}[h]
    \centering
    \includegraphics[width=0.4\textwidth]{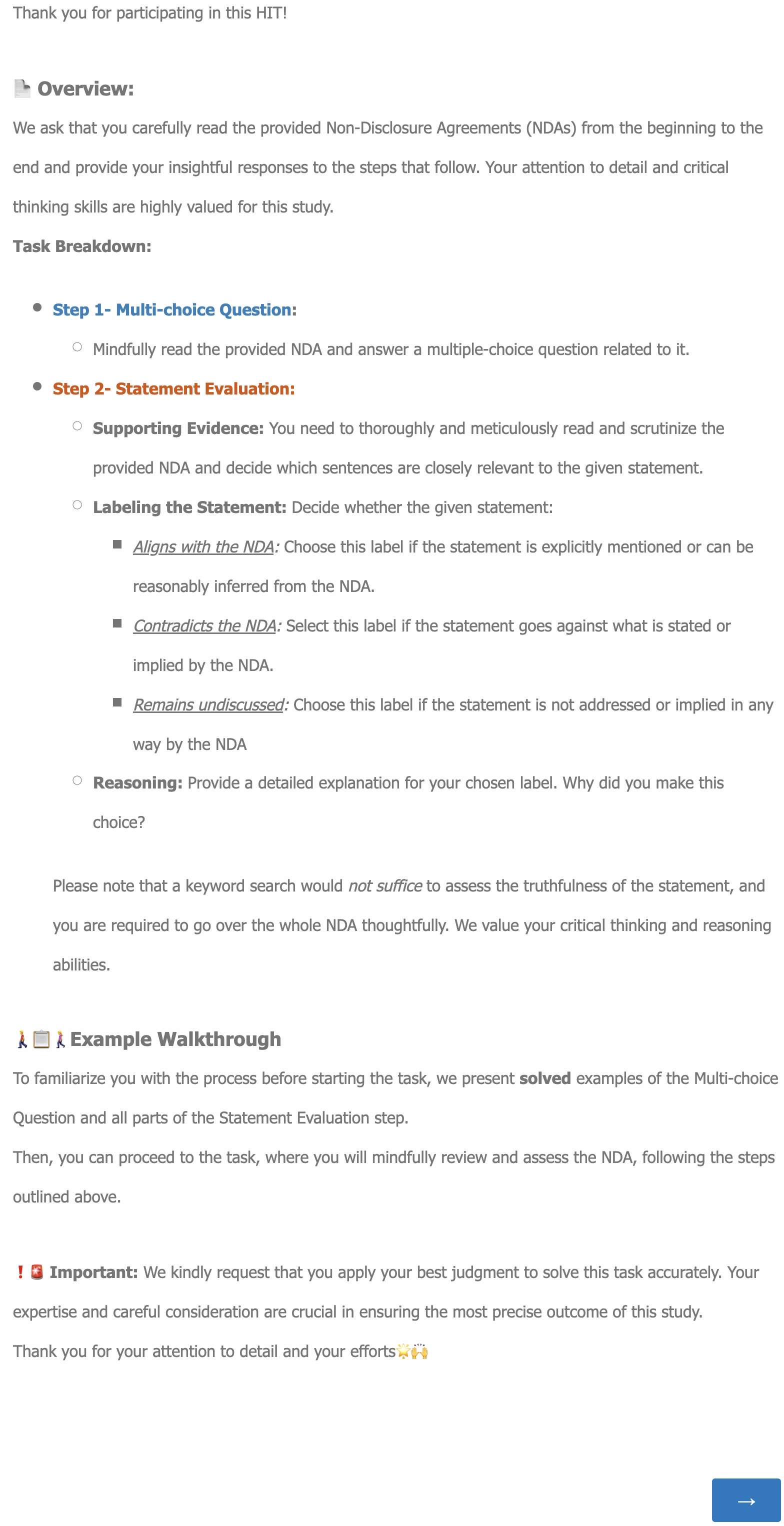}
    \caption{Instructions for expert performance estimation experiment.}
    \label{fig:instructions_perfromance}
\end{figure}

To approximate experts' performance, we conduct a small study consisting of an attention check MCQ followed by a two-step statement assessment: labeling and open-ended reasoning description. See Fig. \ref{fig:instructions_perfromance} for the instructions provided to the participants.
Each participant is given a unique, longer, potentially more challenging NDA with roughly 1650 words (close to the average length of NDAs in the ContractNLI dataset) and one smaller NDA of about 650 words, which is the same for all participants. 
The smaller NDA is paired with a statement explicitly mentioned in the NDA for easy labeling; this serves as an additional check for participants' attentiveness, complementing the MCQ, and all of them answered it correctly.
We pair each of the 5 longer NDAs with a randomly selected unique statement\footnote{All NDAs are paired with all 17 statements in the ContractNLI data.}.
Participants are recruited through the Prolific platform and meet the qualifications.

\subsection{Additional Details on Producing and Presenting Explanations}
\label{sec:highlight}
We present three types of explanations to participants across various studies: model confidence scores, input highlights, and influential training examples. In the following, we detail how we generate and present them to the participants.

\begin{figure}[h]
    \centering
    \includegraphics[width=0.47\textwidth]{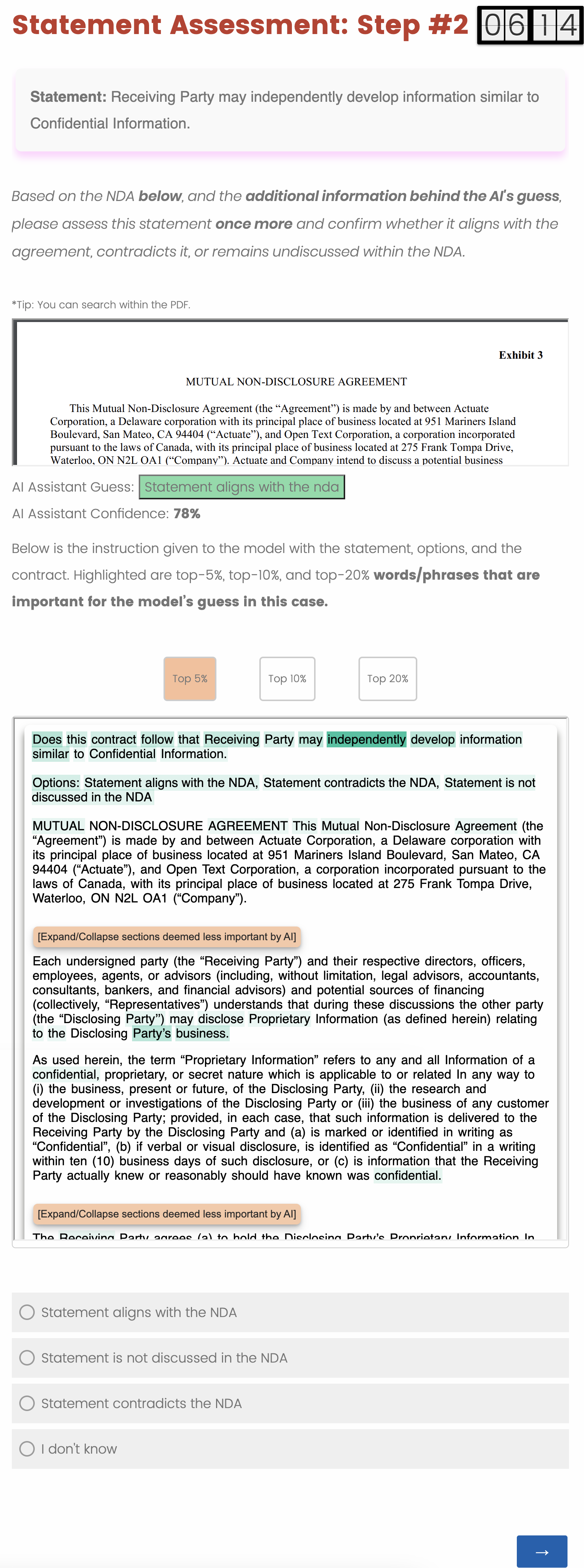}
    \caption{Second iteration of statement evaluation with access to \texttt{Input$\times$Gradient} highlights.}
    \label{fig:highlight}
\end{figure}

\begin{figure}[h]
    \centering
    \includegraphics[width=0.47\textwidth]{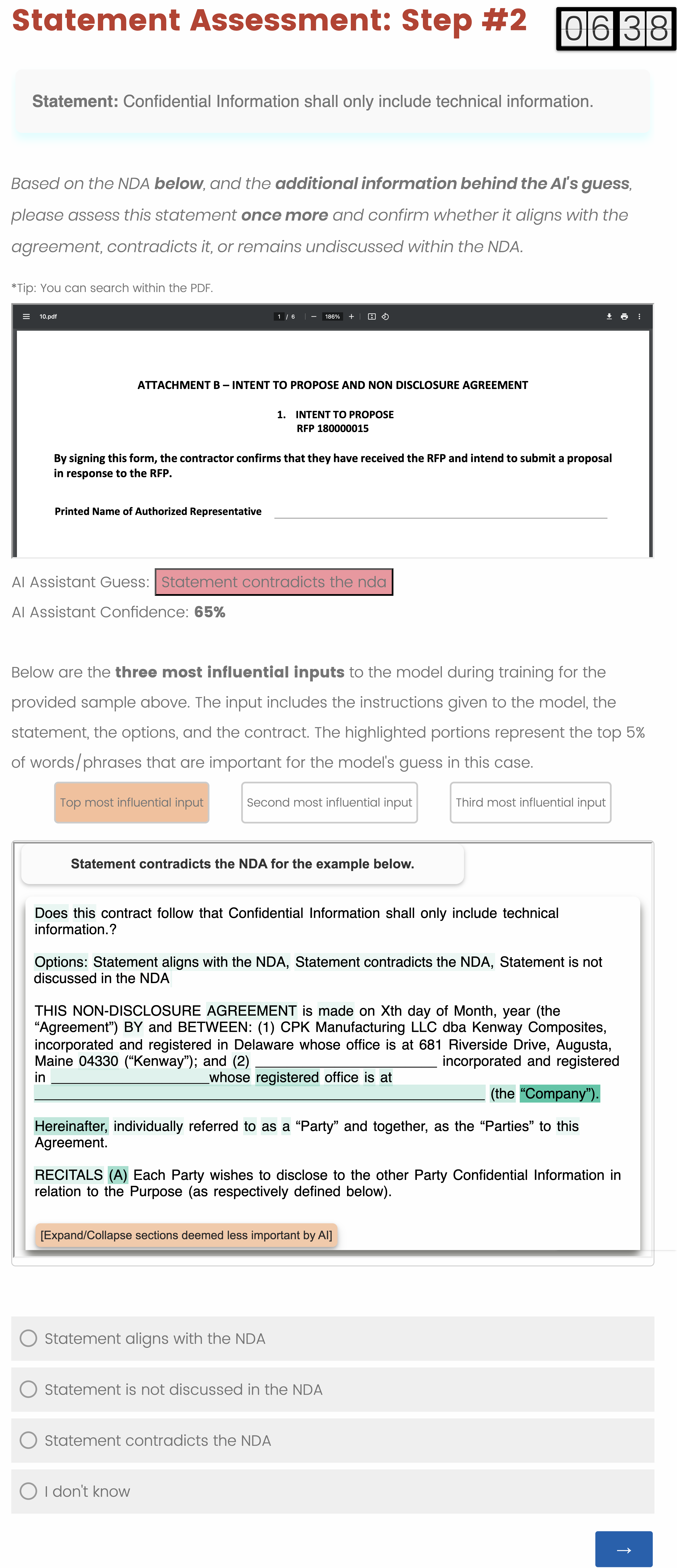}
    \caption{Second iteration of statement evaluation while accessing the EK-FAC influential training examples.}
    \label{fig:influential_examples}
\end{figure}

\paragraph{Confidence score.}
We use the model's prediction probabilities calibrated with temperature scaling \citep{pmlr-v70-guo17a} \footnote{\href{https://github.com/gpleiss/temperature_scaling}{https://github.com/gpleiss/temperature\_scaling}} over the development dataset of ContractNLI to calculate a confidence score for each sample, ranging from 0 to 100.
The confidence score is shown alongside the AI's guess when presented to participants. Examples can be seen in Fig. \ref{fig:highlight} and \ref{fig:influential_examples}.

\paragraph{Input highlights.}
Highlights show which parts of the input are important for the model to make its prediction. 
We use the \texttt{Input$\times$Gradient} \cite{DBLP:journals/corr/ShrikumarGSK16} scores obtained with \texttt{inseq} \cite{sarti-etal-2023-inseq}.
We display the highlight scores by varying the color intensity of the important words.
Higher scores correspond to more intense highlight colors presented to participants. 
To display the Top-N\% of highlights, we retain the highest N\% of highlight scores and set the remainder to zero. 
Each word's highlight intensity is determined by the average non-zero highlight scores of its tokens. 
To prevent overwhelming the participants while maintaining NDA integrity, we give them the option to expand/collapse sections with no highlights. An example is shown in Fig. \ref{fig:highlight}.

\paragraph{Influential training examples.}
We identify influential training examples with EK-FAC \cite{DBLP:conf/nips/GeorgeLBBV18,DBLP:journals/corr/abs-2308-03296}. %
As previously mentioned, the maximum number of input tokens Flan-T5-3B can process is determined by GPU memory capacity, which for us results in 4.2K tokens. %
Since NDAs in ContractNLI can be even 8K tokens long, truncation is necessary. %
Some statements, like those defining confidential information, are typically discussed at the beginning of contracts, while others, such as termination conditions, are addressed towards the end. %
To prevent loss of vital information in lengthier contracts, we produce two input versions\,---\,one where the statement is appended and another where it is prepended to the NDA (as shown in Table~\ref{tab:contract_nli_example})\,---\,and truncate input from the opposite side. %
We then calculate influence scores for both versions of inputs for each NDA-statement pair in an evaluation sample and NDA-statement in the training data, average them, and select the top three most influential training examples. 
For these three influential examples, we additionally incorporate \texttt{Input$\times$Gradient} highlights to streamline the verification process so that participants do not have to review three more NDAs to swiftly verify if the AI's guess is correct. %
See Fig. \ref{fig:influential_examples} for an example.

\subsection{Recruiting, Examining, and Paying Participants} 
\label{sec:participation}

All our studies begin with informed consent. %
We filter for individuals who have a degree in administration and law, work in legal functions, and are fluent in English. %
Unfortunately, even some domain experts on Prolific behaved suspiciously during our preliminary studies. %
Therefore, we create a qualification exam to deter spammers; details on its creation are provided in \sect{sec:attention_check}.  %
No annotator sees the same example more than once across conditions. %
However, ensuring no overlap across studies for the contradicting label is not possible due to insufficient samples. %
Nonetheless, we ensure that no sample is repeated across two consecutive studies.
We pay annotators \$12.45 per hour. %

\section{Details on Deferral Models Input and Deferral Results for User  Study II (\sect{sec:study2})}
\label{sec:deferral}
\paragraph{Data preprocessing.} 

We use the development set of the ContractNLI dataset, dividing it into 1,400 samples for training, 320 for evaluation, and 371 for testing, to develop the deferral models. We finetune \texttt{Llama-2\--\-13B\--\-Chat} using all the training samples. 
For finetuning \texttt{gpt\--\-3.5\--turbo-1106}, we balance the training set by preserving all 156 positive samples and randomly selecting an equal number of negative samples.

To incorporate highlights in the input, we wrap the top 5\% of highlighted words with an HTML-like tag: \texttt{<important></important>}.

\paragraph{End-of-study reflections.}
Below is the feedback we get from some participants in Study II, mainly regarding the accuracy of gold labels presented to them when they received the feedback on their assessment claiming they made a wrong decision. 

{\small
\texttt{\textbf{Participant-1}: \quotes{i disagree with your final feedback. additionally it is difficult to distinguish whether something is absent, or whether it contradicts, when there is a contradicting statement within the NDA}}
}

{\small
\texttt{\textbf{Participant-2}: \quotes{Interpretation of agreements is at times difficult. Lawyers by their training try to interprept agreements in many different ways. There was one example I disagreed with the ending decision.
}}
}

{\small
\texttt{\textbf{Participant-3}: \quotes{I felt like some of the answers were wrong - especially the one relating to verbal information. The NDA discussed orally conveyed information, but the answer still said that the NDA did not discuss it.
}}
}

{\small
\texttt{\textbf{Participant-4}: \quotes{I disagree with the outcomes (final feedback) based on the reasoning I gave.}}
}

{\small
\texttt{\textbf{Participant-5}: \quotes{I feel some of these answers were not correct.}}
}

{\small
\texttt{\textbf{Participant-6}: \quotes{It seemed more difficult than the previous studies!}}
}

\begin{table*}[t]
\centering
\resizebox{\textwidth}{!}{%
\begin{tabular}{llllll}
\toprule
\textbf{$\bm{y_h^{(1)}}$} & \textbf{\bm{$y_p$}} & \multicolumn{2}{l}{$\bm{y_h^{(2)}}$ \textbf{(after $\bm{y_h^{(1)}} \land \bm{y_p} \land \bm{c_p}$)}} & \multicolumn{2}{l}{$\bm{y_h^{(3)}}$ \textbf{(after $\bm{y_h^{(2)}} \land \bm{e_p}$)}} \\
\midrule
\multirow{8}{*}{\cmark} & \multirow{4}{*}{\cmark} & \multirow{2}{*}{\cmark} & \multirow{2}{*}{Confirmation} & \cmark & Effects of $e_p$ undetermined but also not interesting \\
 &  &  &  & \xmark & Unlikely (Spammers?) \\
 \cmidrule{3-6}
 &  & \multirow{2}{*}{\xmark} & \multirow{2}{*}{Unlikely (Spammers?)} & \multicolumn{2}{l}{(Don't give 3rd chance)} \\
 &  &  &  & \multicolumn{2}{l}{(Don't give 3rd chance)} \\
  \cmidrule{2-6}
 & \multirow{4}{*}{\xmark} & \multirow{2}{*}{\cmark} & \multirow{2}{*}{Correct Self-Reliance (CSR)} & \cmark & $e_p$ could be reinforcing CSR (good), doing nothing, or deterring from CSR \\
 &  &  &  & \xmark & $e_p$ causing OR \\
\cmidrule{3-6}
 &  & \multirow{2}{*}{\xmark} & \multirow{2}{*}{Overreliance (OR)} & \cmark & $e_p$ fixing OR \\
 &  &  &  & \xmark & $e_p$ could be reinforcing OR (bad), doing nothing, or deterring from OR (good) \\
  \midrule
\multirow{8}{*}{\xmark} & \multirow{4}{*}{\cmark} & \multirow{2}{*}{\cmark} & \multirow{2}{*}{Correct  Reliance (CR)} & \cmark & $e_p$ could be reinforcing CR (good), doing nothing, or deterring from CR (bad) \\
 &  &  &  & \xmark & $e_p$ causing UR \\
  \cmidrule{3-6}
 &  & \multirow{2}{*}{\xmark} & \multirow{2}{*}{Underreliance (UR)} & \cmark & $e_p$ fixing UR \\
 &  &  &  & \xmark & $e_p$ could be reinforcing UR (bad), doing nothing, or deterring from UR (good) \\
  \cmidrule{2-6}
 & \multirow{4}{*}{\xmark} & \multirow{2}{*}{\cmark} & \multirow{2}{*}{Unlikely (Spammers?)} & \multicolumn{2}{l}{(Don't give 3rd chance)} \\
 &  &  &  & \multicolumn{2}{l}{(Don't give 3rd chance)} \\
  \cmidrule{3-6}
 &  & \multirow{2}{*}{\xmark} & \multirow{2}{*}{Confirmation} & \cmark & Unlikely (Spammers?) \\
 &  &  &  & \xmark & Effects of $e_p$ undetermined but also interesting\\
\bottomrule\\
\end{tabular}
}
\caption{Extending \citet{DBLP:conf/iui/SchemmerKBBS23}'s study. Show a prediction, $y_p$, and confidence, $c_p$, and only then an explanation, $e_p$. $y_h^{(1)}$ is a human's initial guess, $y_h^{(2)}$ is the 2nd guess upon seeing $y_p$, and $y_h^{(3)}$ is the 3rd guess after seeing $e_p$. \cmark\xspace (\xmark\xspace) is the correct (wrong) guess.}
\label{tab:schammer_extra}
\end{table*}

\begin{table*}[h]\centering
\small
\resizebox{\textwidth}{!}{%
\begin{tabular}{p{15cm}p{0.01cm}}\toprule
\textbf{Determine if the claim is true based on the text below:}\textbackslash n \textbf{Claim:} A high microerythrocyte count raises vulnerability to severe anemia in homozygous alpha (+)- thalassemia trait subjects.\textbackslash n\textbackslash n\textbf{Options: True, False, Not enough information}\textbackslash n\textbackslash nBACKGROUND The heritable haemoglobinopathy alpha(+)-thalassaemia is caused by the reduced synthesis of alpha-globin chains that form part of normal adult haemoglobin (Hb).
\textbackslash nIndividuals homozygous for alpha(+)-thalassaemia have microcytosis and an increased erythrocyte count.\textbackslash nAlpha(+)-thalassaemia homozygosity confers considerable protection against severe malaria, including severe malarial anaemia (SMA) (Hb concentration < 50 g/l), but does not influence parasite count.
\textbackslash nWe tested the hypothesis that the erythrocyte indices associated with alpha(+)-thalassaemia homozygosity provide a haematological benefit during acute malaria.
\textbackslash nThis model predicted that children homozygous for alpha(+)-thalassaemia lose less Hb than children of normal genotype for a reduction in erythrocyte count of >1.1 x 10(12)/l as a result of the reduced mean cell Hb in homozygous alpha(+)-thalassaemia.\textbackslash nIn addition, children homozygous for alpha(+)-thalassaemia require a 10\% greater reduction in erythrocyte count than children of normal genotype (p = 0.02) for Hb concentration to fall to 50 g/l, the cutoff for SMA.
\textbackslash nWe estimated that the haematological profile in children homozygous for alpha(+)-thalassaemia reduces the risk of SMA during acute malaria compared to children of normal genotype (relative risk 0.52; 95\% confidence interval [CI] 0.24-1.12, p = 0.09).\textbackslash nCONCLUSIONS The increased erythrocyte count and microcytosis in children homozygous for alpha(+)-thalassaemia may contribute substantially to their protection against SMA.\textbackslash nOther host polymorphisms that induce an increased erythrocyte count and microcytosis may confer a similar advantage.\textbackslash nRBC counts, Hb, Hct, MCH, MCHC values were significantly higher in b- thalassemia minor comparing with IDA patients but MCV showed no significant difference in these two groups.
\textbackslash nThis point sometimes leads misdiagnosis particularly in coincident IDA and $\beta$-thalassemia minor.\textbackslash nTherefore in suspicious cases of $\beta$-thalassemia trait in IDA background, it is better to do hemoglobin electrophoresis after treatment of IDA.
\textbackslash nHowever, the Hb F level was significantly higher in patients with S/Thal having two XmnI sites carrying Arab-Indian and Senegal haplotypes as compared to Bantu, Benin and Cameroon haplotypes.
\textbackslash nThalassemia trait (TT)-related anemia is a common hematologic problem in Mediterranean region.
\textbackslash nThis type of anemia may be frequently confused with iron deficiency anemia (IDA).\textbackslash nAnemia becomes more severe in case of co-existence of both anemia types.
\textbackslash nThalassemia is a congenital hemolytic disorder caused by a partial or complete deficiency of $\alpha$- or $\beta$-globin chain synthesis.\textbackslash nHomozygous carriers of $\beta$-globin gene defects suffer from severe anemia and other serious complications from early childhood.\textbackslash nThe disease is treated by chronic blood transfusion.\textbackslash nSome forms of $\alpha$ thalassemia are also associated with a similar clinical picture.
\textbackslash nAs a consequence, additional previously undescribed, complications are now being recognized.
\textbackslash nBackground-Alpha-thalassemia is one of the most prevalent hemoglobin disorders in the world.\textbackslash nAs a result, a considerable number of patients with microcytic, hypochromic anemia and normal Hb A2 levels might be misdiagnosed as silent $\beta$-thalassemia.\textbackslash nThey were tested for the 2 most frequent $\alpha$ -thalassemia deletions (-$\alpha$ 3.7, -$\alpha$ 4.2).\textbackslash nResults of CBC, hemoglobin analysis, and average annual frequencies of severe pain episodes and numbers of transfused red cell units were documented.\textbackslash nSickle\u2013thalassemia association resulted in higher hemoglobin, hematocrit, and erythrocyte counts with reduced MCV and reticulocytes.		\textbackslash n				 \\
\bottomrule
\end{tabular}
}
\caption{A representative input sample for the \textsc{SciFact-Open} task. The template is: Determine if the claim is true based on the text below:\textbackslash n Claim: {\textbf{claim}}\textbackslash n\textbackslash \textbf{nOptions}: True, False, Not enough information\textbackslash n\textbackslash n{\textbf{text}}\textbackslash n{answer}: }
\label{tab:scifactopen_example}
\end{table*}

\begin{table*}[h]\centering
\small
\resizebox{\textwidth}{!}{%
\begin{tabular}
{p{15.1cm}p{0.01cm}}

\toprule

CONFIDENTIALITY AND NON-DISCLOSURE AGREEMENT\textbackslash nBUSINESS: \_\_\_\_\_\_\_\_\_\_\_\_\_\_\_\_\_\_\_\_\_\_\_\_\_\_\_\_\_\_\_\textbackslash nADDRESS: \_\_\_\_\_\_\_\_\_\_\_\_\_\_\_\_\_\_\_\_\_\_\_\_\_\_\_\_\_\_\_\_\_\_\_\textbackslash nDESCRIPTION:\textbackslash nASKING PRICE: .\textbackslash nThis is intended to be a legally binding document. 
This agreement shall be governed by and enforced in accordance with the laws of the State of California, USA as applicable to contracts to be performed therein. 
The undersigned (hereinafter, collectively, 'Buyer') acknowledges its/his/her desire to receive from Epsteen \& Associates ('Broker') and from the owner of the Business, described above ('Seller') certain information pertaining to the Business, the Seller and/or the possible sale of the Business (the 'Transaction'). 
For purpose of this Agreement, (a) the term 'Buyer' means all of undersigned, including both the potential buyer interested in the Transaction, and such buyer's broker, and both such buyer and buyer's broker are bound by the provisions of this agreement; and (b) any information provided to Buyer, or otherwise learned by Buyer, concerning the Business, Seller or Transaction shall collectively be referred to herein as 'Confidential Information'. 
In consideration of Broker providing Confidential Information to Buyer, Buyer agrees to the following:\textbackslash nKEEP INFORMATION CONFIDENTIAL\textbackslash nBuyer acknowledges that any Confidential Information disclosed to others may be damaging to the Business and the Seller. 
Buyer understands that Confidential Information includes, without limitation: the fact that Business is for sale; financial details; identity of suppliers and customers; and any information not generally known by public. Buyer agrees not to disclose Confidential Information to anyone other than its/his/her advisors and affiliates who both (a) have a need to know the information in connection with the Transaction; and (b) have agreed by signing a copy of this agreement to be bound by the terms of this agreement. 
Buyer agrees that all copies of materials and data provided to Buyer (and any information derivative of such information) shall also be 'Confidential Information'; and all Confidential Information shall be returned to Broker in the event that Buyer decides not to pursue the Transaction. 
Buyer shall be legally responsible for the actions and omissions of Buyer's advisors and affiliates.\textbackslash nDIRECT ALL CONTACT THROUGH BROKER\textbackslash nBuyer shall not contact the Seller or any other individual or entity associated with Seller or the Business including, without limitation, landlords, employees, suppliers or customers except upon the prior written consent of Seller. 
All correspondence, inquiries, and offers to purchase, and other documents relating to the Business or Transaction (all of which is 'Confidential Information) will be delivered solely through Broker, and all negotiations relating to the Business or Transaction will be conducted exclusively through Broker.\textbackslash nUSE INFORMATION FOR EVALUATION PURPOSE ONLY\textbackslash nWithout limiting the other restrictions in this agreement, Buyer agrees to use Confidential Information solely to internally evaluate the Business for the possible Transaction and not for any other purposes whatsoever.\textbackslash nDO NOT CIRCUMVENT SELLER AND/OR BROKER Buyer will not circumvent Seller and/or Broker by contacting any person or persons involved with the Business including, without limitation, landlords, employees, suppliers or customers. 
CONFIDENTIAL INFORMATION IS PROVIDED BY SELLER\textbackslash nAll information about the Business is provided by the Seller and is not verified by Broker. 
Buyer understands that purchasing any business represents investment risks and that Buyer should obtain professional assistance from independent accounting, legal, and financial advisors to verify all information prior to consummating an agreement to purchase the Business. 
Buyer will not rely on the information provided by Broker or Seller, including the Confidential Information, but shall conduct its own independent due diligence. 
Seller (and not Broker) is the source of all information and statements about the Business. 
Broker makes no warranty, guarantee, expressed or implied, as to the accuracy of such information.\textbackslash nBuyer agrees to defend, indemnify, protect and hold harmless Broker, and release Broker, in connection with all information provided to Buyer, including all Confidential Information, and in connection with any breach by Buyer of any of its obligations under this agreement.\textbackslash nPROVIDE EVIDENCE OF FINANCIAL ABILITY\textbackslash nShould Buyer present an offer to purchase the Business, Buyer will provide a financial statement and a personal and business history, and Buyer authorizes Broker and Seller to obtain through standard reporting agencies, financial and credit information about Buyer and/or the companies Buyer represents.\textbackslash nENFORCEMENT\textbackslash nBuyer acknowledges and agrees that any breach of any of its/his/her obligations hereunder will cause Seller and the Business irreparable harm for which Seller and the Business have no adequate remedy at law, and that Seller and the Business shall be entitled to injunctive and other equitable relief to prevent a breach or continued breach of this agreement, in addition to any other remedies Seller and Business may have at law or in equity, and that this agreement shall be specifically enforceable in accordance with its terms. 
Both Broker and Seller are beneficiaries of this agreement and are both entitled to enforce this agreement.\textbackslash nIn any action or proceeding, whether or not resulting in litigation, between Buyer (or either of them) and Seller, or between Buyer (or either of them) and Broker, including any litigation to enforce any of the terms of this agreement, the prevailing party shall be entitled to recover, in addition to any damages or compensation received, its costs and expenses incurred in connection with such action or proceeding, including any reasonable attorneys' fees, expenses and court costs.\textbackslash nWe, the undersigned, understand and agree that this agreement is legally binding upon us. 
We understand that Seller and/or Broker have the right to seek any and all lawful remedies to enforce the terms of this agreement. 
We acknowledge that we have read and understand the disclosures contained herein.\textbackslash nBUYER: BUYER'S BROKER/AGENT:\textbackslash n\_\_\_\_\_\_\_\_\_\_\_\_\_\_\_\_\_\_\_\_\_\_\_\_\_\_\_\_\_\_\_\_\_\_\_\_\_\_\_\_\_ \_\_\_\_\_\_\_\_\_\_\_\_\_\_\_\_\_\_\_\_\_\_\_\_\_\_\_\_\_\_\_\_\_\_\_\_\_\_\_\_\_\textbackslash nSIGNATURE DATE SIGNATURE DATE\textbackslash n\_\_\_\_\_\_\_\_\_\_\_\_\_\_\_\_\_\_\_\_\_\_\_\_\_\_\_\_\_\_\_\_\_\_\_\_\_\_\_\_\_\_ \_\_\_\_\_\_\_\_\_\_\_\_\_\_\_\_\_\_\_\_\_\_\_\_\_\_\_\_\_\_\_\_\_\_\_\_\_\_\_\_\_\_\textbackslash nPRINT NAME PRINT NAME\textbackslash n
\textbackslash n\textbackslash n\textbackslash n\textbf{Does this contract follow that} Receiving Party may create a copy of some Confidential Information in some circumstances?\textbackslash n\textbf{Options: Yes, No, Cannot say} \\
\bottomrule
\end{tabular}
}
\caption{A representative input sample for the Contract-NLI task. The temples is: \quotes{\{\textbf{premise}\}\textbackslash n\textbackslash nDoes this contract follow that \quotes{\{\textbf{hypothesis}\}}?\textbackslash nOptions: Yes, No, Cannot say \textbackslash n\{answer\}} }\label{tab:contract_nli_example}
\end{table*}

\begin{table*}[h]\centering
\small
\resizebox{\textwidth}{!}{%
\begin{tabular}{p{15cm}p{0.01cm}}\toprule
civil appellate jurisdiction civil appeal number 1415 of\textbackslash n1981.\textbackslash nappeal by special leave from the judgment and order\textbackslash ndated the 7th january 1981 of the allahabad high companyrt in\textbackslash ncivil misc. 
application number 113 of 1981 in second appeal number\textbackslash n1484 of 1973.\textbackslash n\textbackslash np. rana m. qamaruddin and mrs. m. qamaruddin for the\textbackslash nappellants. k. sanghi for respondent number 1.\textbackslash nthe judgment of the companyrt was delivered by\textbackslash ndesai j. special leave granted. we have heard mr. o. p. rana learned companynsel for the\textbackslash nappellant and mr. a.k. sanghi learned companynsel for the\textbackslash nrespondent. the high companyrt disposed of the appeal preferred\textbackslash nby the present appellant in the absence of the learned\textbackslash ncounsel for the appellant. 
when the appellant became aware\textbackslash nof the fact that his appeal had been disposed of in the\textbackslash nabsence of his advocate he moved an application in the high\textbackslash ncourt to recall the order dismissing his appeal and permit\textbackslash nhim to participate in the hearing of the appeal. this\textbackslash napplication was rejected by the high companyrt on the ground\textbackslash nthat though the application was prepared and drafted and an\textbackslash naffidavit was sworn on 29th october 1980 the same was number\textbackslash npresented to the companyrt till numberember 12 1980 and that there\textbackslash nis numbersatisfactory explanation for this slackness on the\textbackslash npart of the learned advocate who was requested to file the\textbackslash napplication. 
the disturbing feature of the case is that under our\textbackslash npresent adversary legal system where the parties generally\textbackslash nappear through their advocates the obligation of the\textbackslash nparties is to select his advocate brief him pay the fees\textbackslash ndemanded by him and then trust the learned advocate to do\textbackslash nthe rest of the things. the party may be a villager or may\textbackslash nbelong to a rural area and may have numberknumberledge of the\textbackslash ncourts procedure. after engaging a lawyer the party may\textbackslash nremain supremely companyfident that the lawyer will look after\textbackslash nhis interest. 
at the time of the hearing of the appeal the\textbackslash npersonal appearance of the party is number only number required\textbackslash nbut hardly useful. therefore the party having done\textbackslash neverything in his power to effectively participate in the\textbackslash nproceedings can rest assured that he has neither to go to\textbackslash nthe high companyrt to inquire as to what is happening in the\textbackslash nhigh companyrt with regard to his appeal number is he to act as a\textbackslash nwatchdog of the advocate that the latter appears in the\textbackslash nmatter when it is listed. it is numberpart of his job. mr. a.k. sanghi stated that a practice has grown up in the high companyrt\textbackslash nof allahabad amongst the lawyers that they remain absent\textbackslash nwhen they do number like a particular bench. maybe he is better\textbackslash ninformed on this matter. ignumberance in this behalf is our\textbackslash nbliss. 
even if we do number put our seal of imprimatur on the\textbackslash nalleged practice by dismissing this matter which may\textbackslash ndiscourage such a tendency would it number bring justice\textbackslash ndelivery system into disrepute. what is the fault of the\textbackslash nparty who having done everything in his\textbackslash npower and expected of him would suffer because of the\textbackslash ndefault of his advocate. if we reject this appeal as mr.\textbackslash n\textbackslash nk. sanghi invited us to do the only one who would suffer\textbackslash nwould number be the lawyer who did number appear but the party\textbackslash nwhose interest he represented. the problem that agitates us\textbackslash nis whether it is proper that the party should suffer for the\textbackslash ninaction deliberate omission or misdemeanumberr of his agent. 
the answer obviously is in the negative. maybe that the\textbackslash nlearned advocate absented himself deliberately or\textbackslash nintentionally. we have numbermaterial for ascertaining that\textbackslash naspect of the matter. we say numberhing more on that aspect of\textbackslash nthe matter. however we cannumber be a party to an innumberent\textbackslash nparty suffering injustice merely because his chosen advocate\textbackslash ndefaulted. therefore we allow this appeal set aside the\textbackslash norder of the high companyrt both dismissing the appeal and\textbackslash nrefusing to recall that order. we direct that the appeal be\textbackslash nrestored to its original number in the high companyrt and be\textbackslash ndisposed of according to law. if there is a stay of\textbackslash ndispossession it will companytinue till the disposal of the\textbackslash nmatter by the high companyrt. there remains the question as to\textbackslash nwho shall pay the companyts of the respondent here.\textbackslash n\textbackslash n\textbf{Multi-choice problem: Determine whether this petition should be accepted or not.}\textbackslash n\textbf{Options: Accept, Reject}\textbackslash n\textbf{Answer:}  \\
\bottomrule
\end{tabular}
}
\caption{A representative input sample for the ILDC$_{\text{single}}$ task. The template is: {\textbf{petition}}\textbackslash n\textbackslash nMulti-choice problem: Decide whether this petition should be accepted or not.\textbackslash nOptions: Accept, Reject\textbackslash nAnswer: {answer} }
\label{tab:ildc_example}
\end{table*}

\begin{table*}[h]\centering
\small
\resizebox{\textwidth}{!}{%
\begin{tabular}{p{15cm}p{0.01cm}}\toprule
Energetic 3.20 ± 0.10 3.28 ± 0.10 < 0.05\textbackslash nParticipants reported being significantly more relaxed, calmer, more energetic, less tired, less sluggish, and felt a higher overall sense of well-being during the intervention period compared to the control.\textbackslash n\textbf{Based on the above text, what's the best answer to this question}: Does administering the treatment `sit-stand desks ( SSDs )' significantly change the Energetic compared to the baseline treatment?\textbackslash n\textbackslash n \textbf{Options:}\textbackslash n\textbf{Significantly increase}\textbackslash n\textbf{, No significantly difference}\textbackslash n\textbf{, Significantly decrease}\textbackslash n\textbackslash n\textbf{Answer:} \\
\bottomrule
\end{tabular}
}
\caption{A representative input sample for the \textsc{ERASER EvidenceInference V2} task. In this sample, there is only one document, but additional documents should be added through retrieval. The template is: 
\quotes{\{\textbf{text}\}\textbackslash nBased on the above text, what is the best answer to this question: \{\textbf{question}\}\textbackslash n\textbackslash n\textbf{Options}:\textbackslash nsignificantly increase\textbackslash nno significantly difference\textbackslash nsignificantly decrease\textbackslash n\textbackslash n\textbf{Answer}: }}
\label{tab:evidence_infer_example}
\end{table*}

\begin{table*}[th]\centering
\small
\begin{tabular}{p{14.5cm}p{0.01cm}}\toprule

\small{
\textcolor{red}{\texttt{\textbf{Prompt}}:}
You are a client who owns a company, entering into a non-disclosure agreement with another company. The following statement \underline{\textit{entails}} the NDA:

{Agreement shall not grant Receiving Party any right to Confidential Information.
}

{What actions will you take to satisfy the above statement?}

\textcolor{blue}{\texttt{\textbf{GPT-4 Response}}:}
To satisfy the statement, I would ensure that the receiving party does not gain any rights to the confidential information. This means they will not be allowed to use the information for their own benefit or disclose it to third parties. The information shared would be strictly for the purpose of the project or partnership at hand. I would also insist on clear guidelines in the NDA about how the confidential information should be handled, stored, and disposed of after the agreement ends. Additionally, I would ensure there are severe penalties for breach of the agreement to deter any misuse of the confidential information. This could include financial penalties or legal action. Finally, I would monitor compliance with the NDA to ensure the receiving party is respecting the terms of the agreement.}

\\
\midrule

\small{
\textcolor{red}{\texttt{\textbf{Prompt}}:}
You are a client who owns a company, entering into a non-disclosure agreement with another company. The following statement \underline{\textit{is not discussed}} in the NDA:

{Receiving Party shall destroy or return some Confidential Information upon the termination of Agreement.
}

{Considering that the above statement is undiscussed and thus unclear, what are you allowed to do while still being within the bounds of the NDA?}

\textcolor{blue}{\texttt{\textbf{GPT-4 Response}}:}
Without a clear statement in the NDA regarding the handling of confidential information after termination of the agreement, you are not explicitly required to destroy or return the information. However, maintaining confidentiality remains paramount. You should continue to treat the information as confidential and not disclose it to any third parties. If you're unsure about how to handle the information post-agreement, it would be best to consult with your legal advisor. To maintain good business relations, you could also reach out to the other party for further clarification. It's always advisable to handle such matters carefully to avoid potential legal disputes.

}

\\
\midrule

\small{
\textcolor{red}{\texttt{\textbf{Prompt}}:}
You are a client who owns a company, entering into a non-disclosure agreement with another company. The following Statement \underline{\textit{contradicts}} the NDA:

{
All Confidential Information shall be expressly identified by the Disclosing Party.
}

{
Considering that the above statement does NOT hold, what are you allowed to do while still being within the bounds of the NDA?
}

\textcolor{blue}{\texttt{\textbf{GPT-4 Response}}:}
Even with the statement not holding, you are still required to protect all confidential information that you receive from the other party under the NDA. This means you cannot disclose this information to anyone who is not part of the agreement, use it for any purpose other than what is stipulated in the agreement, or use it to the detriment of the disclosing party. You are also expected to take reasonable steps to protect the information and limit access to those who really need to know it. However, you are not obligated to identify or label the information as confidential since the statement does not hold, and you can treat all information received in the course of the agreement as confidential. It's also important to remember that any obligations under the NDA continue even after the termination of the agreement until a specified time. Always consult with a legal professional to understand your obligations under any legal agreement.

}

\\
\bottomrule
\end{tabular}
\caption{Samples of input prompts along with the responses from GPT-4 for generating actions regarding pairs of statements and labels.}
\label{tab:gpt4_action_generation}
\end{table*}

\begin{table*}[th]\centering
\small
\begin{tabular}{p{14.5cm}p{0.01cm}}\toprule

\small{
\textcolor{red}{\texttt{\textbf{Prompt}}:}
You are a client who owns a company, entering into a non-disclosure agreement with another company.

{
The statement <Some obligations of Agreement may survive termination of Agreement.>  \underline{\textit{entails}} the NDA, and your lawyer correctly confirms this inference.
}

{
Because your lawyer was able to inform you of the conditions, you were able to fulfill the following intended requirements/actions:
}

{
"As a client entering into a non-disclosure agreement (NDA), I would be required to continue to uphold certain obligations even after the termination of the agreement, as per the statement. To satisfy these obligations, I would ensure that any confidential information received during the term of the agreement remains confidential and is not disclosed to any third parties. This could involve continued secure storage of the information or its proper destruction if it's not needed. I would also ensure that my employees, contractors or anyone who had access to the information during the partnership understand their ongoing obligations to maintain confidentiality. I would refrain from using the confidential information for any purpose other than what was specified in the agreement, even after its termination. Any breach of these obligations could result in legal consequences, so it's crucial to respect the terms of the NDA."
}

{
Can you state some benefits that the above actions give to your company due to your lawyer's correct advice directly specific to the statement above? Phrase your answer in one to two sentences, in a punchy way, as if you were writing a report TL;DR to a supervisor so that later on can be used to provide feedback to the lawyer.
}

\textcolor{blue}{\texttt{\textbf{GPT-4 Response}}:}
By adhering to our lawyer's advice on the NDA terms, we not only safeguard our company's reputation and maintain trust with external partners, but also mitigate potential legal liabilities, ensuring our ongoing business operations remain uncompromised.
}

\\
\midrule

\small{
\textcolor{red}{\texttt{\textbf{Prompt}}:}
You are a client who owns a company, entering into a non-disclosure agreement with another company.

{Although the statement <Receiving Party may create a copy of some Confidential Information in some circumstances.>  \underline{\textit{is not discussed}} in the NDA, your lawyer made a mistake and reported that the statement \underline{\textit{entails}} the NDA.
}

{
As a consequence, your company failed to comply with the following intended actions:
}

{"In the absence of explicit permission in the NDA, we should assume that we are not allowed to make copies of the Confidential Information. NDAs are designed to protect sensitive information, and typically restrict unauthorized copying, sharing, or usage of such information. To stay within the bounds of the NDA, we should only use the Confidential Information as specifically permitted in the agreement and avoid creating any copies. If we find it necessary to create a copy for your business operations, it is advisable to seek written consent from the disclosing party. This ensures that we remain compliant with the NDA and respect the other party's proprietary rights."
}

{
Can you mention a concrete legal penalty you might face due to your lawyer's incorrect advice directly specific to the statement above? Phrase your answer in one to two sentences, in a punchy way, as if you were writing a report TL;DR to a supervisor so that later on can be used to provide feedback to the lawyer.
}

\textcolor{blue}{\texttt{\textbf{GPT-4 Response}}:}
Due to the lawyer's incorrect advice, our company may face severe legal penalties including a breach of contract lawsuit, potentially resulting in substantial financial damages and injunctions prohibiting further use of the confidential information. Additionally, this breach could tarnish our reputation, jeopardizing future business partnerships and opportunities.
}

\\
\bottomrule
\end{tabular}
\caption{Two instances of input prompts and GPT-4 responses for generating feedback on the benefits of correctly labeling a specific statement (top example) and the consequences of mislabeling a particular statement (bottom example).}
\label{tab:gpt4_feedback_generation}
\end{table*}

\begin{table*}[!htp]\centering

\scriptsize
\resizebox{\textwidth}{!}{%
\begin{tabular}{p{1.8cm}p{12cm}}\toprule
\textbf{Statement} &Receiving Party may retain some Confidential Information even after the return or destruction of Confidential Information. \\
\midrule
\textbf{Gold label} &No mention \\
\midrule
\textbf{AI label} &Contradict \\
\midrule
\textbf{Expert label} &Contradict \\
\midrule
\textbf{Author label} &Contradict \\
\midrule
\textbf{Expert comment} &\quotes{The highlighted phrases show statements that may be in relation to the statement that has been asked. It shows the most accurate statements and paragraphs to search the answer from . the highlighred phrases gave guidance in terms of where to look and how to answer so in conclusion it did reveal that some confidential information must be returned and never to be retained.} \\
\midrule
\textbf{Relevant spans of NDA (by author)} &5. RETURN OF CONFIDENTIAL INFORMATION Either party may request in writing at any time that any confidential information or any documents containing such confidential information, disclosed pursuant to the terms of this agreement and any copies thereof be returned with a written statement to the effect that upon such return the receiving party has not knowingly retained in its possession or under its control, either directly or indirectly, any such confidential information or any documents containing such confidential information or any copies thereof and the receiving party shall comply with any such request within seven days of receipt of such request. \\
\midrule
\textbf{NDA file} &\url{https://contract-nli.s3.us-east-2.amazonaws.com/hit_files/57.pdf} \\
\bottomrule
\end{tabular}
}
\caption{One example where we change the gold label after reviewing the NDA and the expert's reasoning.}
\label{tab:revised_gold_label}
\end{table*}

\begin{table*}[t]
\centering
\small
\resizebox{\textwidth}{!}{%
\begin{tabular}{p{1.6cm}p{3cm}p{9.9cm}}
\toprule
\textbf{Theme} & \textbf{Description} & \textbf{Examples} \\
\midrule
Relevant information highlighted (49.1\%) & Participants use the highlights to identify specific clauses or phrases that are crucial to answering the question. & \texttt{\textbf{Example-1:} \quotes{the highlighted phrases were very important cause they helped me too lok through the document thoroughly and see what the NDA say about the higlighted word. This made it easier to do a process of elimination. It is easier and quickier to go through a document this way. it ensures accuracy and also saves time.} } \\
\cmidrule{3-3}
& & \texttt{\textbf{Example-2:} \quotes{It still contradicts the statement because no where in there does it say that confidential information may be shared. It states that they should not disclose confidential information at any time. It is discussed in this NDA. The examples highlight that this is the case. It mentions third-parties, however it specifically states it cannot be shared with them.}} \\
\midrule
\multirow{2}{1.8cm}{Comparison (16.4\%)} & \multirow{2}{3cm}{Examples and highlights verify or adjust the participants' initial assessment.} & \texttt{\textbf{Example-1:} \quotes{It is not express within the document that the definition of 'Confidential Information' includes verbally conveyed information, but it flows from the highlighted statements logically that one of the 'forms' of transmitting 'Confidential Information' could logically include verbally conveyed information. Thus, it is not on all fours within the document but I would say generally the statement aligns with the document. I was on the fence between identifying the document as aligning, or not including express reference to the statement, so therefore not referenced in the document. AI has highlighted the relevant area, and so I have in this case leant slightly more towards it aligning.}  } \\
\cmidrule{3-3}
& & \texttt{\textbf{Example-2:} \quotes{i did not see any discussion of technical information within the NDA during my own analysis.  when looking at the highlighted portions, i still did not see any discussion of technical information.  this added to the confidence level of my own analysis and decision that the statement is not included within the NDA.}} \\
\midrule
\multirow{2}{1.8cm}{Irrelevant or misleading information highlighted (10.1\%)} & \multirow{2}{3cm}{Examples sometimes highlight irrelevant or misleading sections, causing confusion for participants and potentially helping them detect AI's mistakes.} & \texttt{\textbf{Example-1:} \quotes{All the highlighted sentence were not consistent with the statement made by the model. This led me to make a different final decision that the one made by the model.  Actually from the highlighted sentence it is evident that the the statement is not discussed in the NDA and therefore the answer provided by the model is not correct.}} \\
\cmidrule{3-3}
& & \texttt{\textbf{Example-2:} \quotes{the three most influential examples refer to the form of the disclosure of the confidential information and not the form in which the confidential information is provided to the receiving party (written or verbal). that means that the receiving party must refrain from disclosing the confidential information either in written or verbal form, but does not discus the form in which the confidential information was provided to the receiving party}} \\
\midrule
\multirow{2}{1.8cm}{Not helpful (7.3\%)} & \multirow{2}{3cm}{Some participants found highlights not to be useful in addressing the statement, leading to confusion or wasted time.} & \texttt{\textbf{Example-1:} \quotes{the highlighted sections provided by the AI are not all that helpful.  it highlighted so many sections, and a lot of those sections are not relevant to the question or statement.  looking at the highlighted sections wasted a lot of time and in the end was not helpful at all in making my decision.}} \\
\cmidrule{3-3}
& & \texttt{\textbf{Example-2:} \quotes{The three most influential examples in the training model had rather little impact on my decision.  It seems the AI is not looking at the most relevant portion of the agreement to answer the quested posed by the researcher.  In fact, some of the information in these examples is actually completely irrelevant.}  } \\
 \bottomrule
\end{tabular}%
}
\caption{Analysis of open-ended responses to the question on the impact of explanations on the participants' decision-making process. This table shows the most common themes in the responses. For each theme, the first example is from responses in the study with highlights, and the second one is for influential examples.
}
\label{tab:free-form_response_codes}
\end{table*}

\begin{table*}[!htp]\centering
\scriptsize
\resizebox{\textwidth}{!}{%
\begin{tabular}{llllrrrr}\toprule
\textbf{Model} &\textbf{Train Setup} &\textbf{Input} &\textbf{Class} &\textbf{Precision} &\textbf{Recall} &\textbf{F1 score} \\\midrule
\multirow{8}{*}{\texttt{Llama-2-13B-chat-hf}} &\multirow{8}{*}{Finetuned} &\multirow{4}{*}{P+T} &\textsc{Defer} &5.6 &10.0 &7.1 \\
& & &\textsc{Do-not-defer} &75.6 &33.2 &46.2 \\
& & &\textsc{Micro Avg.} &66.9 &30.3 &41.3 \\
& & &\textsc{Macro Avg.} &40.6 &21.6 &26.7 \\
\cmidrule{3-7}
& &\multirow{4}{*}{P+T+H} &\textsc{Defer} &2.9 &5.0 &3.6 \\
& & &\textsc{Do-not-defer} &54.6 &21.4 &30.8 \\
& & &\textsc{Micro Avg.} &48.1 &19.4 &27.4 \\
& & &\textsc{Macro Avg.} &28.7 &13.2 &17.2 \\
\midrule
\multirow{12}{*}{\texttt{GPT-4o}} &\multirow{4}{*}{Zero-shot} &\multirow{4}{*}{P+T} &\textsc{Defer} &12.5 &10.0 &11.1 \\
& & &\textsc{Do-not-defer} &87.5 &90.0 &88.7 \\
& & &\textsc{Micro Avg.} &78.1 &80.0 &79.0 \\
& & &\textsc{Macro Avg.} &50.0 &50.0 &49.9 \\
\cmidrule{2-7}
&\multirow{8}{*}{30-shot} &\multirow{4}{*}{P+T} &\textsc{Defer} &\textbf{18.9} &17.5 &18.2 \\
& & &\textsc{Do-not-defer} &88.3 &89.3 &88.8 \\
& & &\textsc{Micro Avg.} &79.7 &80.3 &80.0 \\
& & &\textsc{Macro Avg.} &53.6 &53.4 &53.5 \\
\cmidrule{3-7}
& &\multirow{4}{*}{P+T+H} &\textsc{Defer} &17.7 &15.0 &16.2 \\
& & &\textsc{Do-not-defer} &88.1 &90.0 &89.1 \\
& & &\textsc{Micro Avg.} &79.3 &80.6 &79.9 \\
& & &\textsc{Macro Avg.} &52.9 &52.5 &52.6 \\
\midrule
\multirow{4}{*}{\texttt{GPT-3.5-turbo-1106}} &\multirow{4}{*}{Finetuned} &\multirow{4}{*}{P+T} &\textsc{Defer} &12.5 &\textbf{30.0} &17.7 \\
& & &\textsc{Do-not-defer} &87.5 &70.0 &77.8 \\
& & &\textsc{Micro Avg.} &78.1 &65.0 &70.3 \\
& & &\textsc{Macro Avg.} &50.0 &50.0 &47.7 \\
\bottomrule
\end{tabular}
}
\caption{Complete results of deferral models (\sect{sec:study2}). P denotes the finetuned Flan-T5-3B's prediction, T a ContractNLI statement-NDA pair text, and H \texttt{Input$\times$Gradient} highlights.}\label{tab:all_deferral_models_performance}
\end{table*}

\begin{table*}[!htp]\centering
\scriptsize
\resizebox{\textwidth}{!}{%
\begin{tabular}{lllrrrrr}\toprule
\textbf{Decision Maker} &\textbf{Data Sample} &\textbf{Class} & \textbf{Precision} &\textbf{Recall} &\textbf{F1 score} & \textbf{\# Examples} \\\midrule
\multirow{15}{*}{Experts} &\multirow{5}{*}{Deferred w/ highlights} &\textsc{Entail} &15.4 &50.0 &23.5 &4 \\
& &\textsc{No-mention} &77.8 &24.1 &36.8 &29 \\
& &\textsc{Contradict} &12.5 &40.0 &19.0 &5 \\
& & \textsc{Micro Avg.} &28.9 &28.9 &28.9 &38 \\
& & \textsc{Macro Avg.} &35.2 &38.0 &26.5 &38 \\
\cmidrule{2-7}
&\multirow{5}{*}{Deferred w/ highlighted influential examples} &\textsc{Entail} &0.0 &0.0 &0.0 &3 \\
& &\textsc{No-mention} &75.0 &44.4 &55.8 &27 \\
& &\textsc{Contradict} &13.3 &40.0 &20.0 &5 \\
& & \textsc{Micro Avg.} &40.0 &40.0 &40.0 &35 \\
& & \textsc{Macro Avg.} &29.4 &28.1 &25.3 &35 \\
\cmidrule{2-7}
&\multirow{5}{*}{All examples deferred to experts} &\textsc{Entail} &11.8 &28.6 &16.7 &7 \\
& &\textsc{No-mention} &76.0 &33.9 &46.9 &56 \\
& &\textsc{Contradict} &12.9 &40.0 &19.5 &10 \\
& & \textsc{Micro Avg.} &34.2 &34.2 &\textbf{34.2} &73 \\
& & \textsc{Macro Avg.} &33.6 &34.2 &27.7 &73 \\
\midrule
Flan-T5-3B & Undeferred samples & \textsc{All} & 100.0 & 100.0 & 100.0 & 1854 \\
\bottomrule
\end{tabular}
}
\caption{Performance of experts and the finetuned Flan-T5-3B on different sets of data in Study II (\sect{sec:study2}).}
\label{tab:deferral_results}
\end{table*}

\begin{figure*}[t]
    \centering
    \includegraphics[width=0.99\textwidth]{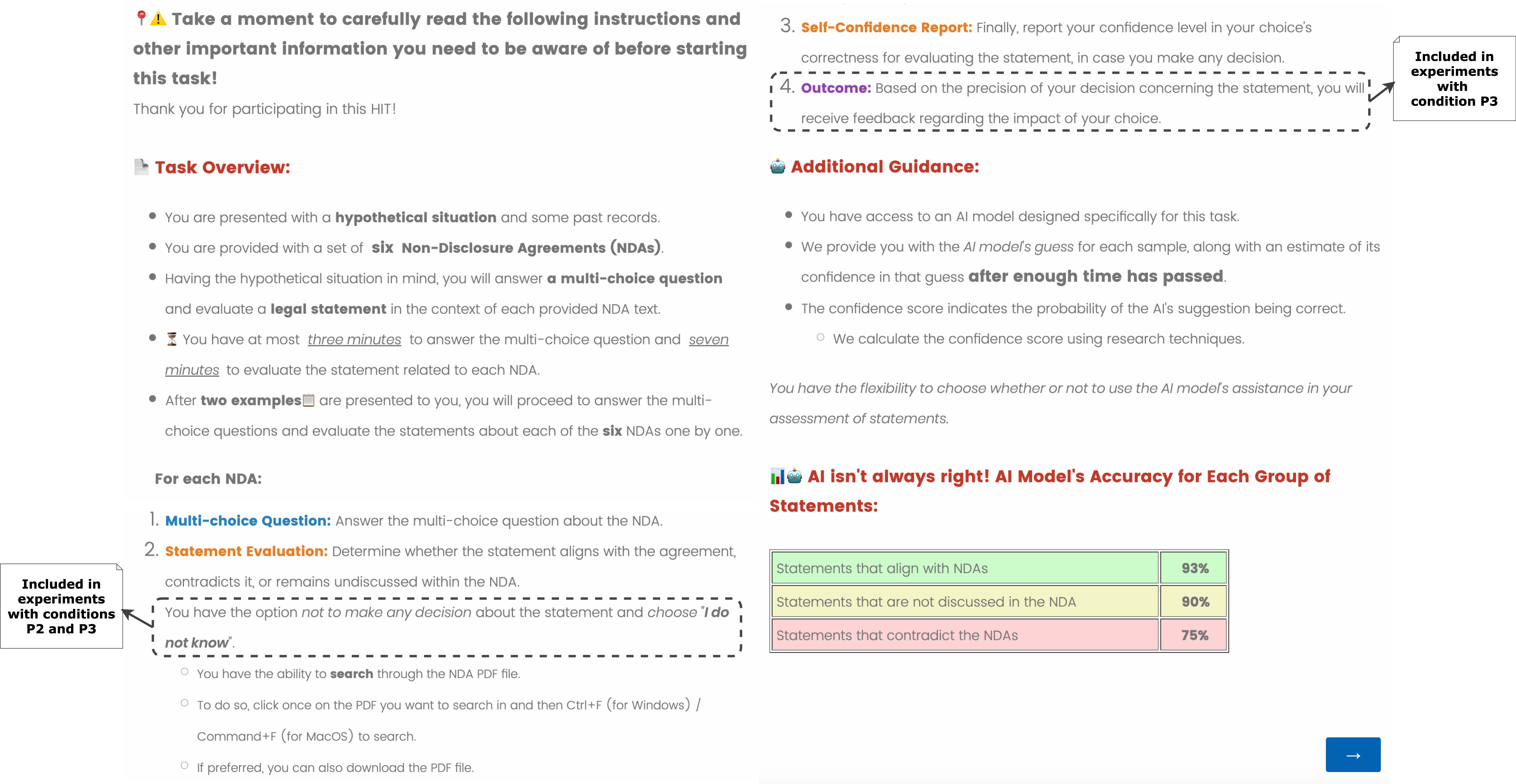}
    \caption{Instructions for Study I (\sect{sec:study1_main}), the experiments with conditions \texttt{P1} to \texttt{P3}.}
    \label{fig:instructions123}
\end{figure*}

\begin{figure*}[h]
    \centering
    \includegraphics[width=0.9\textwidth]{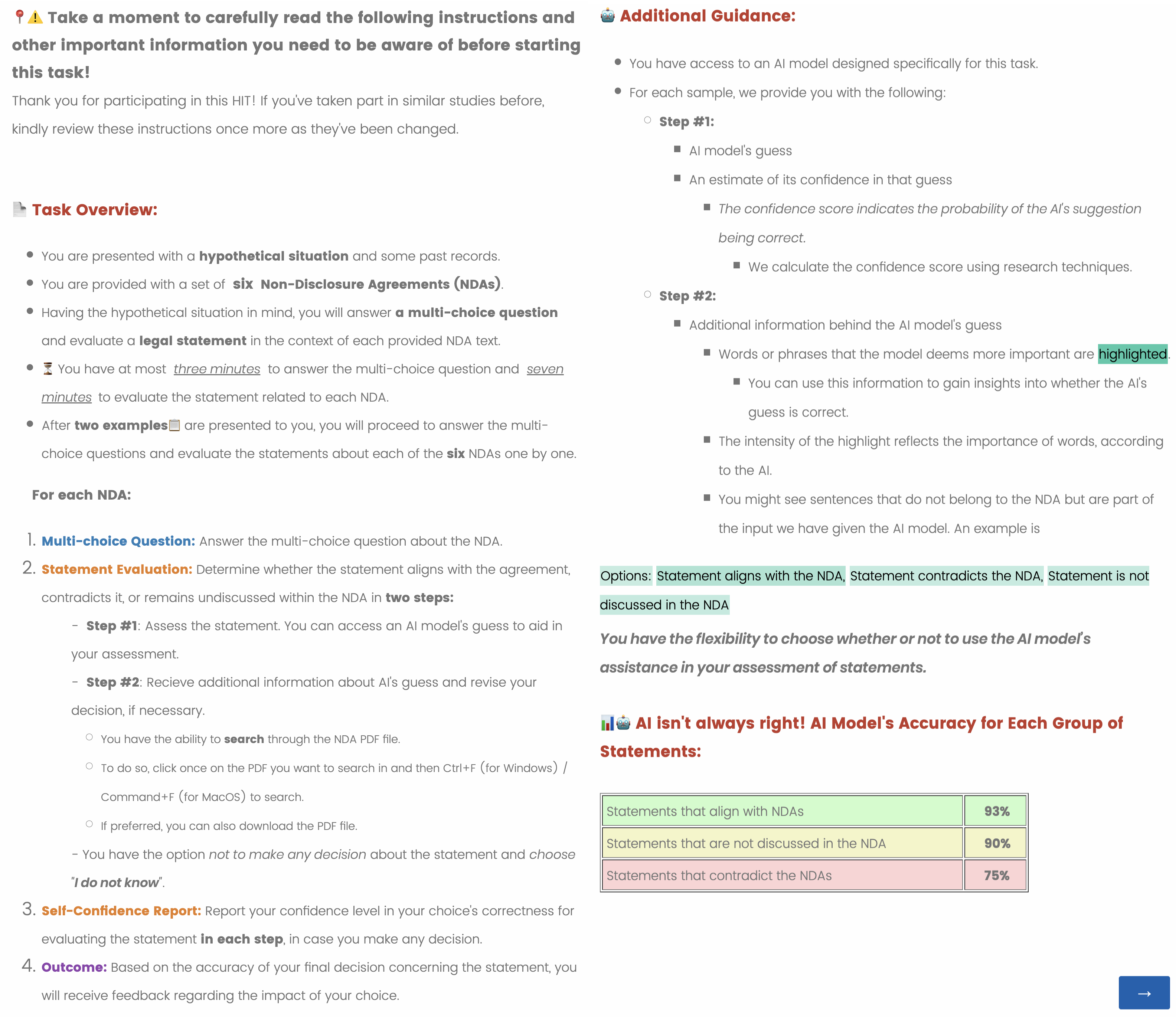}
    \caption{Instructions for Study I (\sect{sec:study1_main}), the experiment with conditions \texttt{P4}.}
    \label{fig:instructions4}
\end{figure*}

\begin{figure*}[h]
    \centering
    \includegraphics[width=0.9\textwidth]{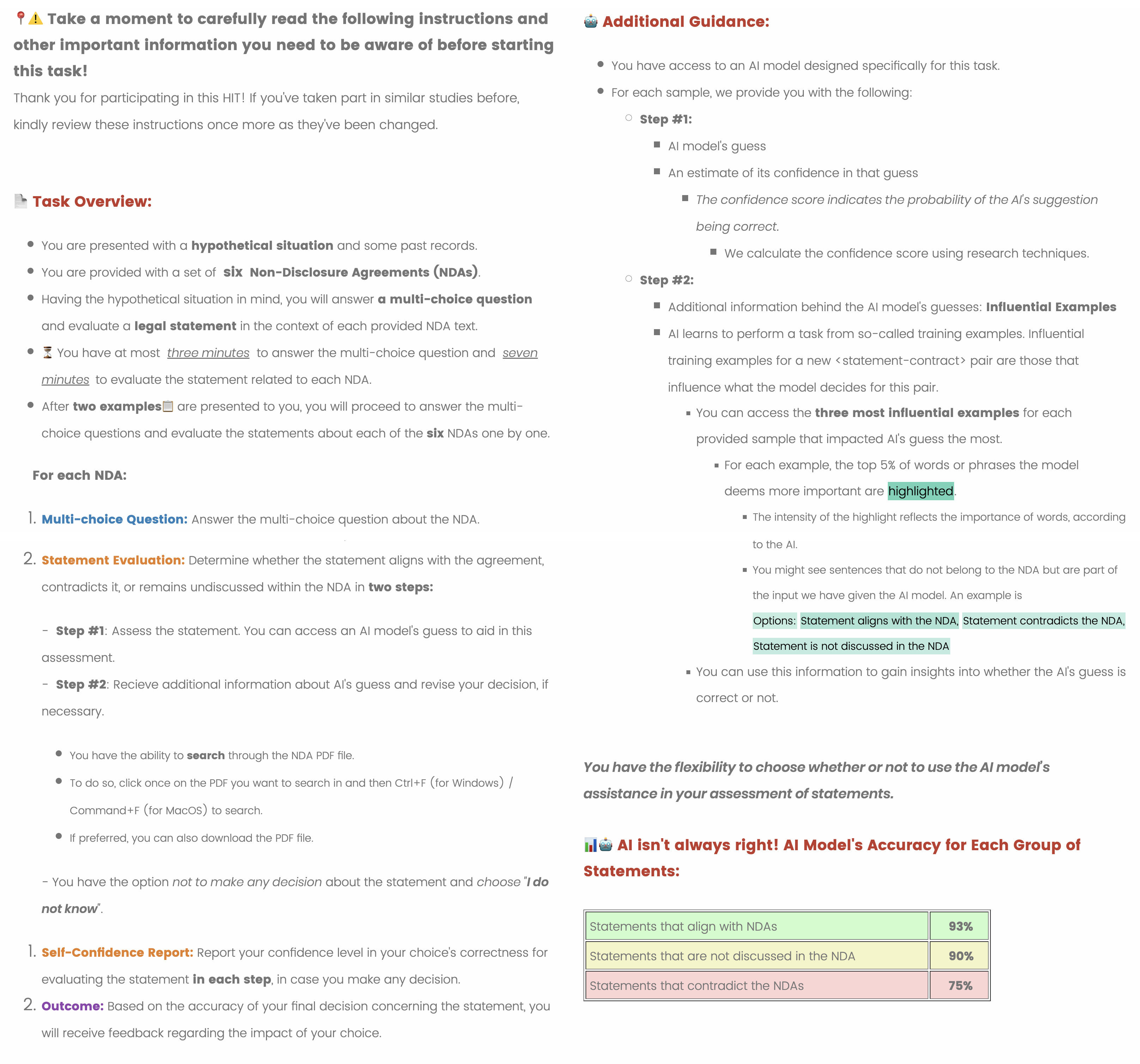}
    \caption{Instructions for Study I (\sect{sec:study1_main}), the experiment with conditions \texttt{P5}.}
    \label{fig:instructions5}
\end{figure*}

\begin{figure}[h]
    \centering
    \includegraphics[width=0.4\textwidth]{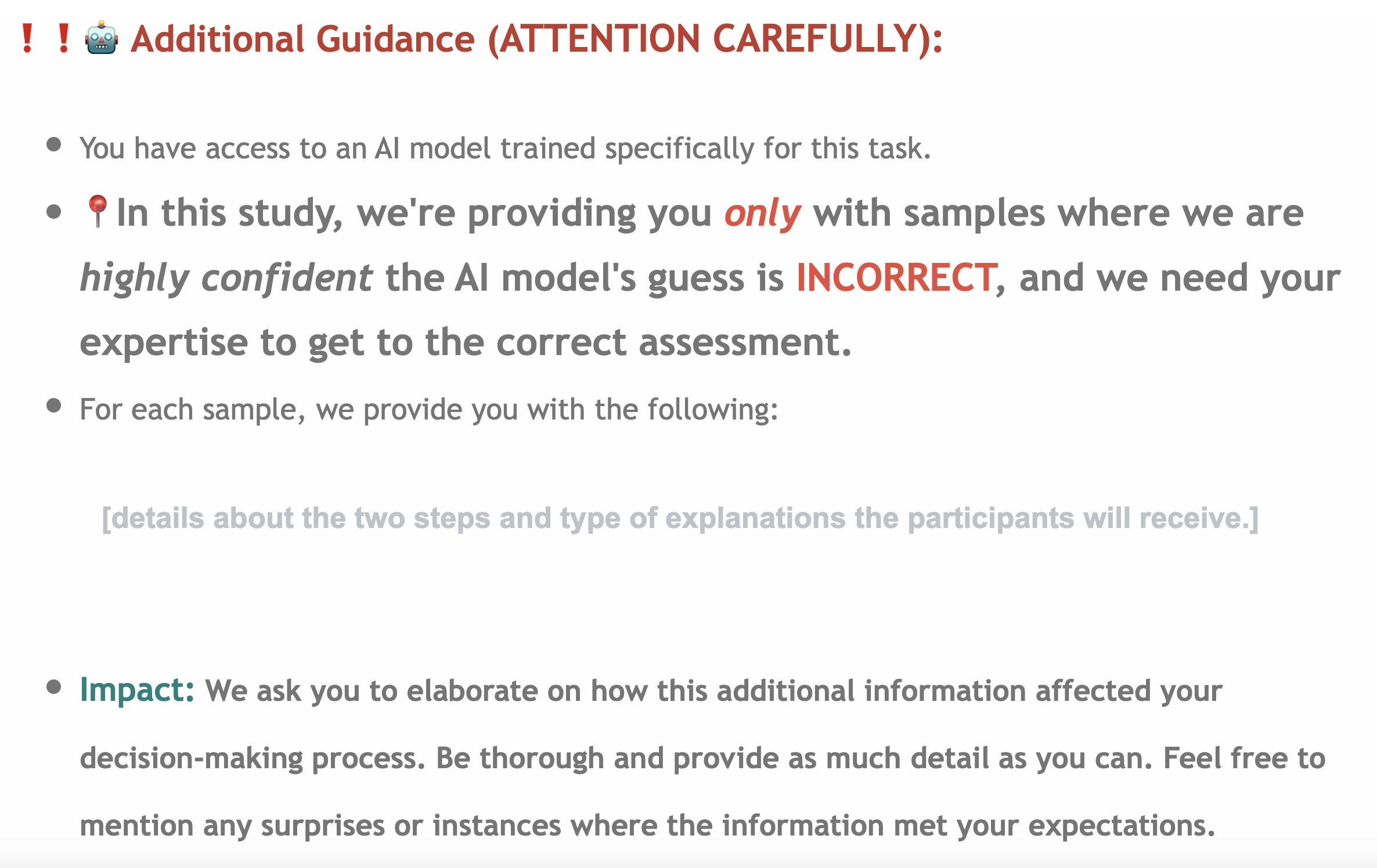}
    \caption{Parts of instructions unique to Study I (\sect{sec:study1_main}). The rest is the same as in Fig. \ref{fig:instructions4}--\ref{fig:instructions5} for the experiment with conditions \texttt{P4} and \texttt{P5}, respectively.}
    \label{fig:instructions_defer}
\end{figure}

\begin{figure}[t]
    \centering
    \includegraphics[width=0.47\textwidth]{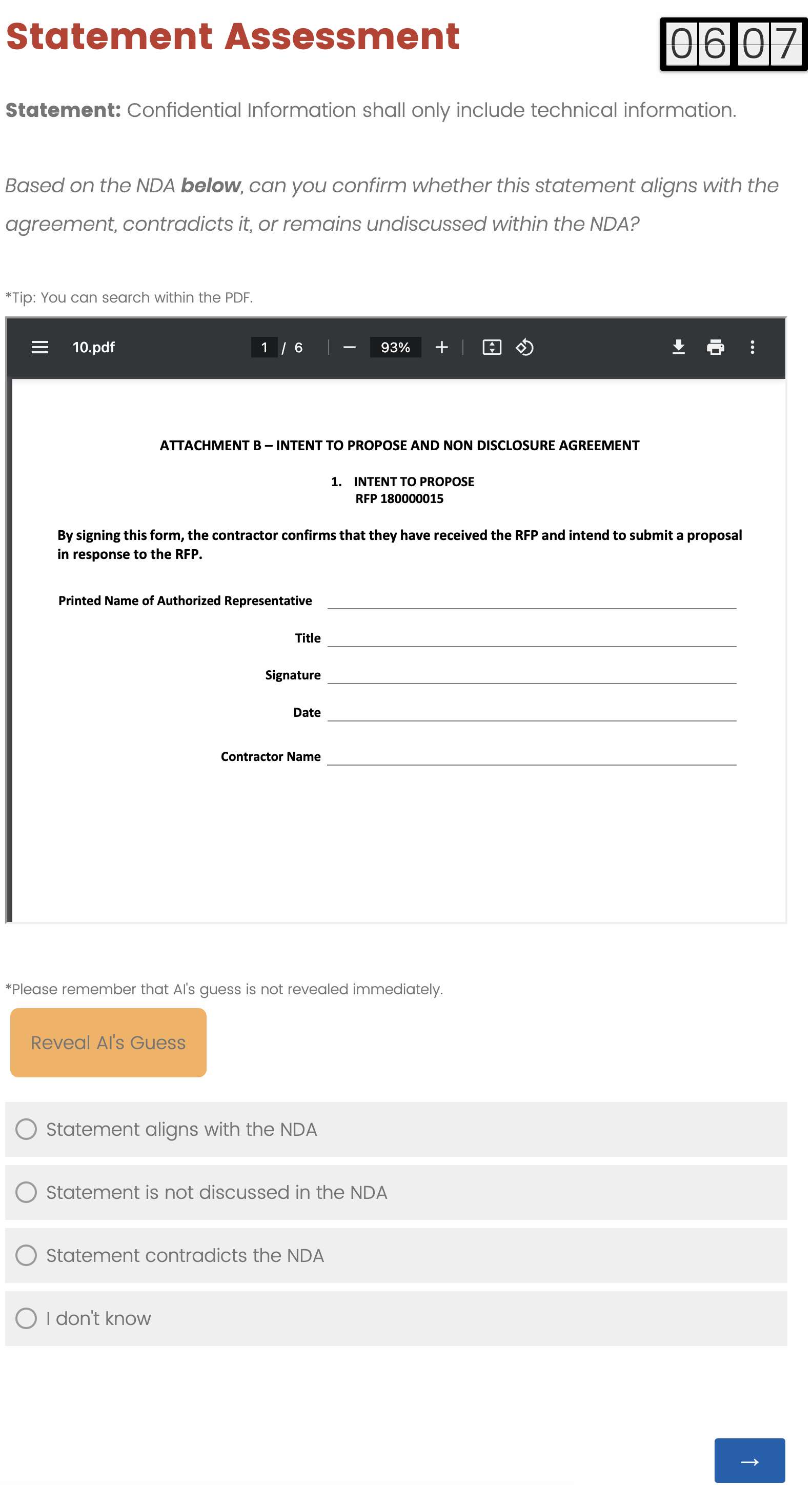}
    \caption{Statement Assessment. Participants can click the ``Reveal AI'' button to view AI's prediction. Upon doing so, a GIF animation is displayed for approximately 5 seconds, followed by presenting AI's prediction and confidence score. AI's prediction is color-highlighted (green, yellow, or red) according to the label.}
    \label{fig:statement_assess}
\end{figure}

\begin{figure}[h]
    \centering
    \includegraphics[width=0.47\textwidth]{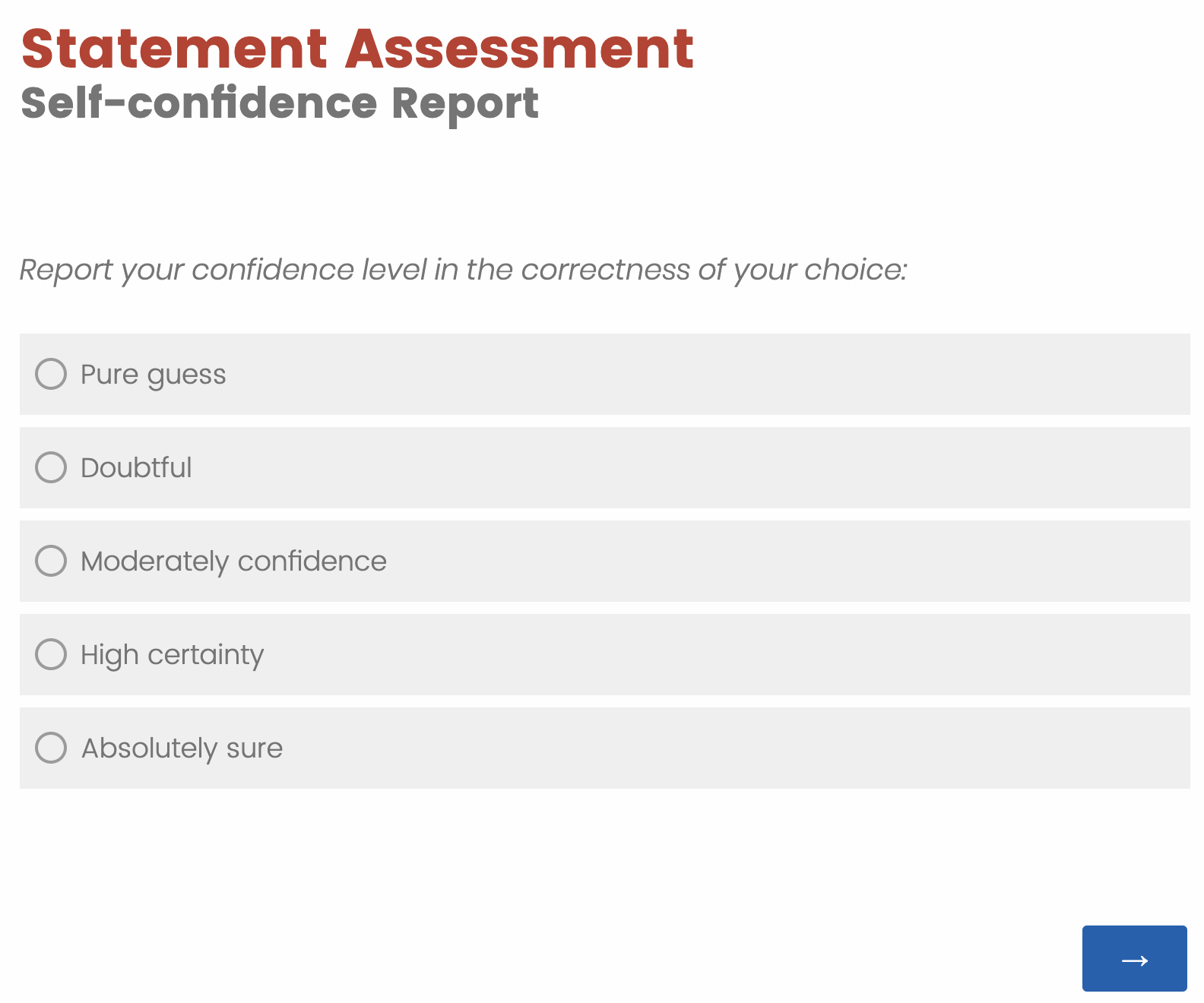}
    \caption{Likert-scale self-confidence report question.}
    \label{fig:confidence}
\end{figure}

\begin{figure}
    \centering
    \includegraphics[width=0.45\textwidth]{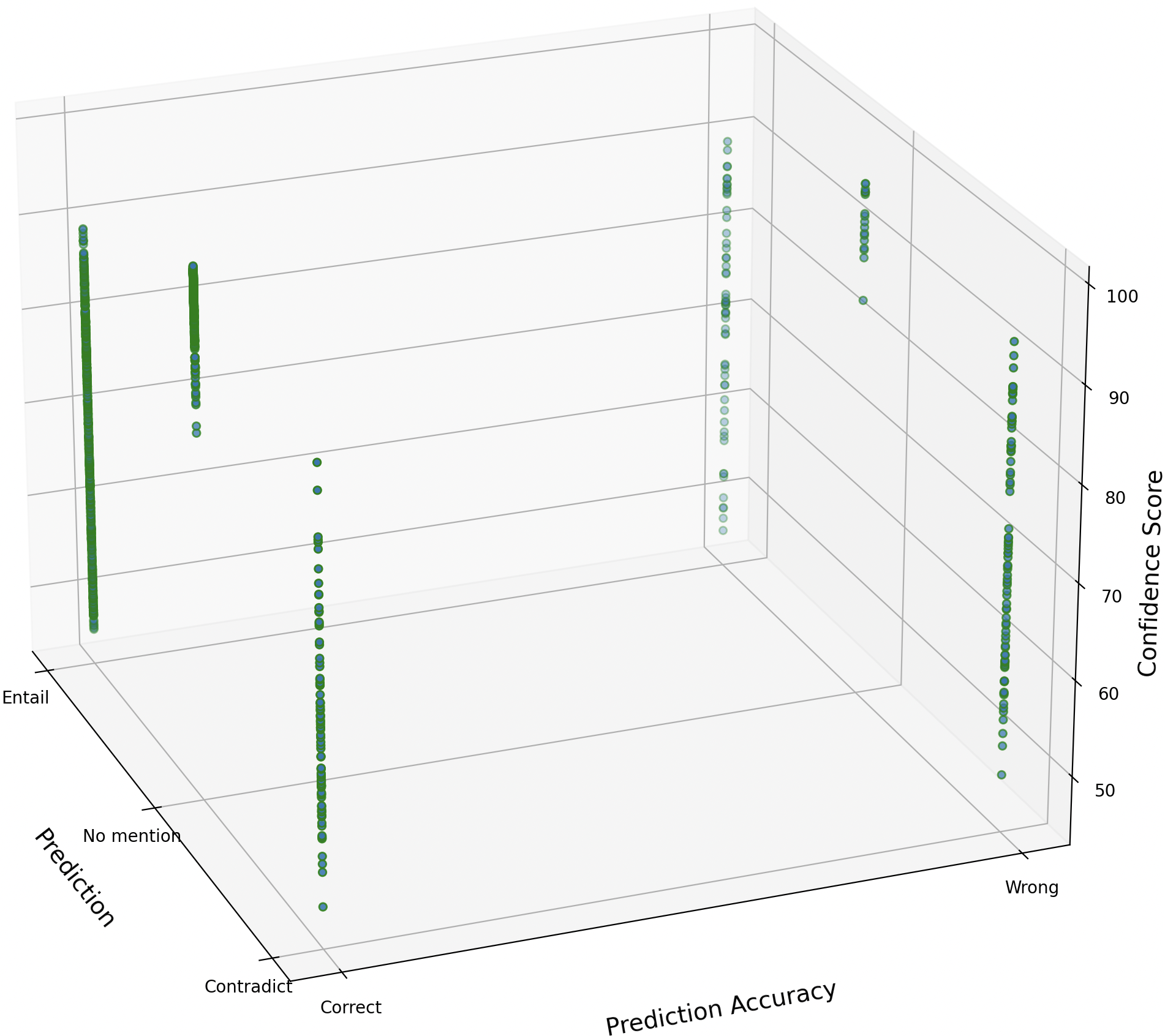}
    \caption{This plot illustrates the relationship between the model's prediction, confidence score, and prediction accuracy. Despite confidence score calibration, there is no clear correlation between the model's confidence scores and prediction accuracy.}
    \label{fig:conf_vs_predAcc}
\end{figure}

\clearpage

\section{Categorization of ExNLP Tasks}
\label{sec:exnlp_task_cards}

\begin{center} \parbox{\linewidth}{\hrulefill} \end{center} %

\subsection*{[\hypertarget{dataset1}{D1}] \citet{ott-etal-2011-finding} }

\noindent\textbf{Prediction Task:}  Finding deceptive opinion spam (``fictitious opinions that have been deliberately written to sound authentic, in order to deceive
the reader'') in the context of hotel reviews.

\noindent\textbf{Average Input Length:}  146 words [review]

\noindent\textbf{Human Ability:} 53-62\% (majority baseline 58\%)

\noindent\textbf{Application:} Deciding whether to engage with a hotel review and book the hotel 

\noindent\textbf{Hazard from Immediate Usage:}
\begin{compactitem}
    \item \emph{Who:} A person booking a hotel 
    \item \emph{Hazard:} Booking a disappointing hotel 
    \item \emph{Probability:} Low. People take multiple factors, not only a few reviews, when booking a hotel, especially if more expensive/important. However, if we assume that they looked at only reviews, we still expect the probability to be low since today's models accurately classify the sentiment of reviews in other domains.
    \item \emph{Severity:} Depends on personal circumstances and expense, but generally low. 
    \item \emph{Risk:} Low
\end{compactitem}

\noindent\textbf{Hazard from Downstream Impact:}
\begin{compactitem}
    \item \emph{Who:} Hotel management 
    \item \emph{Hazard:} Public complaints that the room was not as described; A customer with the right expectations does not get a room  
    \item \emph{Probability:} Low, since the probability from the immediate usage is low
    \item \emph{Severity:} Moderate, since repeatedly getting public complaints and missing the right customers can hurt the business to some degree 
    \item \emph{Risk:} Low   
\end{compactitem}

\begin{center} \parbox{\linewidth}{\hrulefill} \end{center} %

\subsection*{[\hypertarget{dataset2}{D2}] \textsc{QuizBowl} \citep{DBLP:conf/iui/FengB19}}

\noindent\textbf{Prediction Task:} Quizbowl (answering questions from all areas of knowledge with as few clues as possible).

\noindent\textbf{Average Input Length:}  $\sim$20 words [question] based on the similar data \citep{DBLP:journals/corr/abs-1904-04792}

\noindent\textbf{Human Ability:} An average player ``buzzes with 65\% of the question shown with 60\% accuracy'' \citep{DBLP:journals/corr/abs-1904-04792}

\noindent\textbf{Application:} Playing Quizbowl as a cooperation with a machine. This version does not exist yet but could happen. 

\noindent\textbf{Hazard from Immediate Usage:}
\begin{compactitem}
    \item \emph{Who:} Quizbowl player 
    \item \emph{Hazard:} Loosing a game 
    \item \emph{Probability:} Depends on the player 
    \item \emph{Severity:} Low
    \item \emph{Risk:} Low
\end{compactitem}

\noindent\textbf{Hazard from Downstream Impact:} If a player loses, they are affecting only themselves.

\begin{center} \parbox{\linewidth}{\hrulefill} \end{center} %

\subsection*{[\hypertarget{dataset3}{D3}] \textsc{ReClor} \citep{DBLP:conf/iclr/YuJDF20}}

\noindent\textbf{Prediction Task:}  Multiple-choice reading comprehension targeting logical reasoning.

\noindent\textbf{Average Input Length:} 65 [context] + 15 [question] + 75 words [choices] = 155 words

\noindent\textbf{Human Ability:} Although it can  be 100\%, \citet{10.1145/3411764.3445717} report 67\%

\noindent\textbf{Application:} No. Models trained on this data could be used to practice for law school admissions if new exams with multiple choices are available but not correct solutions. However, practice exams come with solutions. 

\noindent\textbf{Hazard from Immediate Usage:} None, as there is no realistic application. 

\noindent\textbf{Hazard from Downstream Impact:} None, as there is no realistic application. 

\begin{center} \parbox{\linewidth}{\hrulefill} \end{center} %

\subsection*{[\hypertarget{dataset4}{D4}] \textsc{BeerAdvocate} \citep{DBLP:conf/icdm/McAuleyLJ12}}

\noindent\textbf{Prediction Task:} Sentiment classification of beer reviews.

\noindent\textbf{Average Input Length:} 88 words [review]

\noindent\textbf{Human Ability:} 87\%; Although this is already high,  \citet{10.1145/3411764.3445717} show this is not the upper bound 

\noindent\textbf{Application:} Deciding whether to buy a beer

\noindent\textbf{Hazard from Immediate Usage:}
\begin{compactitem}
    \item \emph{Who:} Beer buyers
    \item \emph{Hazard:} Buying a beer they do not like  
    \item \emph{Probability:} Assuming that beers that are positively reviewed are liked by new customers, we expect the probability to be low since today's models accurately classify the sentiment of reviews in other domains.
    \item \emph{Severity:} Low since the cost of a bottle/can of beer is generally low 
    \item \emph{Risk:} Low 
\end{compactitem}

\noindent\textbf{Hazard from Downstream Impact:} Nothing noteworthy. 

\begin{center} \parbox{\linewidth}{\hrulefill} \end{center} %

\subsection*{[\hypertarget{dataset5}{D5}] \textsc{NaturalQuestions} \citep{kwiatkowski-etal-2019-natural}}

\noindent\textbf{Prediction Task:} Identifying a span in a Wikipedia article that answers an open-ended question (originally asked in Google Search).

\noindent\textbf{Average Input Length:} 5197 [document] + 9 [question] = 5206 words

\noindent\textbf{Human Ability:} Reported human F1 score is 87\% (long answers), 76\% (short answers)

\noindent\textbf{Application:} Information search 

\noindent\textbf{Hazard from Immediate Usage:}
\begin{compactitem}
    \item \emph{Who:} Anyone
    \item \emph{Hazard:} Propagating misinformation from accepting the wrong answer 
    \item \emph{Probability:} Moderate ($\exists$ model-human performance gap)\footnote{\url{https://ai.google.com/research/NaturalQuestions/leaderboard};  \url{https://paperswithcode.com/sota/question-answering-on-natural-questions}}
    \item \emph{Severity:} Low (questions are not about critical information such as health, law, etc.)
    \item \emph{Risk:} Low
\end{compactitem}

\noindent\textbf{Hazard from Downstream Impact:} Same as for the immediate usage.

\begin{center} \parbox{\linewidth}{\hrulefill} \end{center} %

\subsection*{[\hypertarget{dataset6}{D6}] \textsc{StrategyQA} \citep{geva-etal-2021-aristotle}}

\noindent\textbf{Prediction Task:} Open-ended QA (1) without any additional context or (2) in the context of retrieved Wikipedia paragraphs.

\noindent\textbf{Average Input Length:} 960\footnote{The models are set to retrieve 10 Wikipedia paragraphs from the corpus, and the average paragraph length is 96.} [context] + 46 [question] = 1003 words 

\noindent\textbf{Human Ability:} Reported human accuracy is 87\%  (``given access to Wikipedia articles and an option to reveal the decomposition for every question'')

\noindent\textbf{Application:} Information search

\noindent\textbf{Hazard from Immediate Usage:}
\begin{compactitem}
    \item \emph{Who:} Anyone
    \item \emph{Hazard:} Propagating misinformation from accepting the wrong answer 
    \item \emph{Probability:} Moderate (currently 12\% model-human accuracy gap)\footnote{\url{https://leaderboard.allenai.org/strategyqa/submissions/public};   \url{https://paperswithcode.com/sota/strategyqa-on-big-bench}} 
    \item \emph{Severity:} Low (questions are not about critical information such as health, law, etc.)
    \item \emph{Risk:} Low
\end{compactitem}

\noindent\textbf{Hazard from Downstream Impact:} Same as for the immediate usage.

\subsection*{[\hypertarget{dataset7}{D7}] \textsc{OpenBookQA} \citep{mihaylov-etal-2018-suit}; \textsc{e-OBQA} \citep{jhamtani-clark-2020-learning}}

\noindent\textbf{Prediction Task:}  Multi-choice middle-school level science exam QA.  

\noindent\textbf{Average Input Length:}  12 [question] + 11 [options] = 23 words

\noindent\textbf{Human Ability:} Reported human performance is ~92\%, but it could be anything from 0 to 100\% depending on a person's knowledge 

\noindent\textbf{Application: } No. Models trained on this data could be used for students in 3rd  through 9th grade to practice for science exams if exams are available, but not correct solutions. However, practice exams come with solutions. 

\noindent\textbf{Hazard from Immediate Usage:} None, as there
is no realistic application.

\noindent\textbf{Hazard from Downstream Impact:} None, as there
is no realistic application.

\begin{center} \parbox{\linewidth}{\hrulefill} \end{center} %

\subsection*{[\hypertarget{dataset8}{D8}] \textsc{MMLU} \citep{DBLP:conf/iclr/HendrycksBBZMSS21}}

\noindent\textbf{Prediction Task:} Multi-choice QA in multiple domains such as elementary mathematics, US history, and law.

\noindent\textbf{Average Input Length:}  50 [question] + 4 [choices] = 54 words

\noindent\textbf{Human Ability:} 34.5\% (Unspecialized people), but it varies based on the question domains and user expertise.

\noindent\textbf{Application:} No. The dataset does not represent a realistic task setup (similar to [\protect\hyperlink{dataset11}{D11}]).

\noindent\textbf{Hazard from Immediate Usage:} None, as there is no realistic application.

\noindent\textbf{Hazard from Downstream Impact:} None, as there is no realistic application.

\begin{center} \parbox{\linewidth}{\hrulefill} \end{center} %

\subsection*{[\hypertarget{dataset9}{D9}] \textsc{FoolMeTwice} \citep{eisenschlos-etal-2021-fool}}

\noindent\textbf{Prediction Task:} 
Determine whether a given claim, spanning various domains, is supported or refuted by the provided evidence from a Wikipedia page.

\noindent\textbf{Average Input Length:}  15 [claim] + 30 [evidence text] = 45 words

\noindent\textbf{Human Ability:} Reported average accuracy is 78.1\%.

\noindent \emph{The remaining information is the same as for \textsc{FEVER} ([\protect\hyperlink{dataset12}{D12}]).}

\begin{center} \parbox{\linewidth}{\hrulefill} \end{center} %

\subsection*{[\hypertarget{dataset10}{D10}] \textsc{Amazon Book Reviews} \citep{10.1145/2872427.2883037}}

\noindent\textbf{Prediction Task:} Sentiment classification of book reviews.

\noindent\textbf{Average Input Length:} 105 words [review]

\noindent\textbf{Human Ability:} Not reported, but we expect people to be good at this task

\noindent\textbf{Application:} Deciding whether to buy a book

\noindent\textbf{Hazard from Immediate Usage:}
\begin{compactitem}
    \item \emph{Who:} Book buyers
    \item \emph{Hazard:} Buying a book they do not like  
    \item \emph{Probability:} Assuming that books that are positively reviewed are liked by new customers, we expect the probability to be low since today's models accurately classify the sentiment of reviews in other domains.
    \item \emph{Severity:} Low since the cost of a book is generally low 
    \item \emph{Risk:} Low 
\end{compactitem}

\noindent\textbf{Hazard from Downstream Impact:} Nothing noteworthy. 

\begin{center} \parbox{\linewidth}{\hrulefill} \end{center} %

\subsection*{[\hypertarget{dataset11}{D11}]  \citet{jansen-etal-2016-whats}}

\noindent\textbf{Prediction Task:} Multiple-choice science exam QA.

\noindent\textbf{Average Input Length:} 20 [question] + 20 [choices] = 40 words

\noindent\textbf{Human Ability:} Depends, but can  be 100\%

\noindent\textbf{Application:}  No. Models trained on this data could be used for students in 3\textsuperscript{rd} to 5\textsuperscript{th} grade to practice for science exams if exams are available, but not correct solutions. However, practice exams come with solutions. 

\noindent\textbf{Hazard from Immediate Usage:} None, as there is no realistic application. 

\noindent\textbf{Hazard from Downstream Impact:} 
None as there is no realistic application.

\begin{center} \parbox{\linewidth}{\hrulefill} \end{center} %

\subsection*{[\hypertarget{dataset12}{D12}] \textsc{FEVER} \citep{thorne-etal-2018-fever}}

\noindent\textbf{Prediction Task:}  Verification of claims ``containing a single piece of
information, focusing on the entity that its original
Wikipedia page was about'', given Wikipedia articles.

\noindent\textbf{Average Input Length:} 8 [claim] + 227 [article] = 235 words

\noindent\textbf{Human Ability:} Not reported 

\noindent\textbf{Application:} Open-ended QA with Wikipedia articles. FEVER claims are simple facts about entities and the task in the real world resembles open-ended QA more than a task that a professional fact-checker does.\footnote{An example of a claim in FEVER is: ``Berlin is the capital of Germany.''}

\noindent\textbf{Hazard from Immediate Usage:}
\begin{compactitem}
    \item \emph{Who:} Anyone
    \item \emph{Hazard:} Propagating misinformation from accepting the wrong answer 
    \item \emph{Probability:} Moderate; SOTA achieves $\sim$80\% accuracy \citep{dehaven2023bevers}
    \item \emph{Severity:} Low (questions are not about critical information such as health, law, etc.)
    \item \emph{Risk:} Low
\end{compactitem}

\noindent\textbf{Hazard from Downstream Impact:} Same as for the immediate usage.

\begin{center} \parbox{\linewidth}{\hrulefill} \end{center} %

\subsection*{[\hypertarget{dataset13}{D13}] \textsc{e-SNLI} \citep{NEURIPS2018_4c7a167b}}

\noindent\textbf{Prediction Task:} Natural language inference 

\noindent\textbf{Average Input Length:} 13 [premise] + 7 [hypothesis] = 20 words

\noindent\textbf{Human Ability:} 89\% \citep{bowman-etal-2015-large} 

\noindent\textbf{Application:} No. SNLI is introduced to probe models' understanding of entailment and contradiction.\footnote{There are application-grounded versions of NLI such as EvidenceInference v2 (\hyperlink{dataset38}{D38}).}

\noindent\textbf{Hazard from Immediate Usage:} None, as there is no realistic application. 

\noindent\textbf{Hazard from Downstream Impact:} None, as there is no realistic application. 

\begin{center} \parbox{\linewidth}{\hrulefill} \end{center} %

\subsection*{[\hypertarget{dataset14}{D14}] \textsc{e-$\delta$-SNLI} \citep{DBLP:conf/aaai/BrahmanSRC21}}

\noindent\textbf{Prediction Task:}  Defeasible natural language inference \citep{rudinger-etal-2020-thinking}.

\noindent \emph{The remaining information is the same as for \textsc{e-SNLI}  above.}

\begin{center} \parbox{\linewidth}{\hrulefill} \end{center} %

\subsection*{[\hypertarget{dataset15}{D15}] \textsc{LIAR-PLUS} \citep{alhindi-etal-2018-evidence}}

\noindent\textbf{Prediction Task:}  Verification of claims about a broad range of topics based on (1) metadata, or (2) metadata and a summary of a report written by a fact checker that discusses the veracity of a claim.

\noindent\textbf{Average Input Length:} (1) 17 [claim] + 50 [metadata] = 67 words; (2) 17 [claim] + 50 [metadata] + 69 [summary] = 136 words

\noindent\textbf{Human Ability:} Not reported. (1) We expect that fact-checking a claim based on metadata, without any reports on the claim, is hard. (2) We expect that it is easy to fact-check a claim based on a short report written by a professional fact-checker that summarizes their research on the veracity of the claim. 

\noindent\textbf{Application:} No. (1) Fact-checking without reading any reports on the claim is not realistic. (2) A summary written by professionals to fact-check a claim already clearly indicates the author's decision of veracity. The LIAR-RAW version (see [\hyperlink{dataset17}{D17}]), where the input is the statement and a few reports, some of which are unreliable, is a reasonable application. 

\noindent\textbf{Hazard from Immediate Usage:} None, as there is no realistic application. 

\noindent\textbf{Hazard from Downstream Impact:} None, as there is no realistic application.

\begin{center} \parbox{\linewidth}{\hrulefill} \end{center} %

\subsection*{[\hypertarget{dataset16}{D16}] \textsc{PubHealth} \citep{kotonya-toni-2020-explainable-automated}}

\noindent\textbf{Prediction Task:}  Verification of claims about public health from a fact-checking/news article discussing the claim written by a professional.

\noindent\textbf{Average Input Length:} 14 [claim] + 707 [article] = 721 words

\noindent\textbf{Human Ability:} Not reported, but we expect that it is easy to fact-check a claim based on a report written by a professional fact-checker that summarizes their research on the veracity of the claim.  

\noindent\textbf{Application:} No. A summary written by professionals to fact-check a claim already clearly indicates the author's decision of veracity. The LIAR-RAW version (see [\hyperlink{dataset17}{D17}]), where the input is the statement and a few documents, some of which are unreliable, is a reasonable application. 

\noindent\textbf{Hazard from Immediate Usage:} None, as there is no realistic application. 

\noindent\textbf{Hazard from Downstream Impact:} None, as there is no realistic application. 

\begin{center} \parbox{\linewidth}{\hrulefill} \end{center} %

\subsection*{[\hypertarget{dataset17}{D17}] \textsc{LIAR-RAW} \citep{yang-etal-2022-coarse}}

\noindent\textbf{Prediction Task:}  Verification of claims about a broad range of topics, given a few reports  (media reports, user
comments, blogs, etc.), some of which are unreliable.

\noindent\textbf{Average Input Length:} 17 [claim] + 1568 [reports] = 1585 words

\noindent\textbf{Human Ability:} Not reported 

\noindent\textbf{Application:} The task setup is realistic because people will first find related articles  (some of which are unreliable) to go about verifying a claim.

\noindent\textbf{Hazard from Immediate Usage:}
\begin{compactitem}
    \item \emph{Who:} Fact checker; Anyone 
    \item \emph{Hazard:} Job performance problems; Propagating misinformation 
    \item \emph{Probability:} Moderate, models' performance is not high \citep{yang-etal-2022-coarse}
    \item \emph{Severity:} Can be high (e.g.\ if someone was defamed); Moderate (the statements are about more important information than in open-ended QA datasets, but not all are about vital information such as health)  
    \item \emph{Risk:} High; Moderate
\end{compactitem}

\noindent\textbf{Hazard from Downstream Impact:}
\begin{compactitem}
    \item \emph{Who:} An entity that false statements were made about and that a fact checker falsely confirmed; Anyone  
    \item \emph{Hazard:} Defamation; Propagating misinformation
    \item \emph{Probability:} Moderate (same as above)
    \item \emph{Severity:} High; Moderate (same as above)
    \item \emph{Risk:} High; Moderate 
\end{compactitem}

\begin{center} \parbox{\linewidth}{\hrulefill} \end{center} %

\subsection*{[\hypertarget{dataset18}{D18}] \textsc{RAWFC} \citep{yang-etal-2022-coarse}}

\noindent\textbf{Prediction Task:}  Verification of short statements on a broad range of topics based on a few reports  (media reports, user
comments, blogs, etc.), some of which are unreliable.

\noindent\textbf{Average Input Length:}  18 [claim] + 4075 [reports] = 4093 words

\noindent \emph{The remaining information is the same as for LIAR-RAW above.}

\begin{center} \parbox{\linewidth}{\hrulefill} \end{center} %

\subsection*{[\hypertarget{dataset19}{D19}] \textsc{ECQA} \citep{aggarwal-etal-2021-explanations}\footnote{\textsc{ECQA} explanation annotations replace  \textsc{CoS-E}'s \citep{rajani-etal-2019-explain} that are too nosiy.}}

\noindent\textbf{Prediction Task:} Multiple-choice QA targeting commonsense.

\noindent\textbf{Average Input Length:} 13 [question] + 13 [choices] = 26 words

\noindent\textbf{Human Ability:} 88.9\% \citep{talmor-etal-2019-commonsenseqa}

\noindent\textbf{Application:} No. \textsc{CQA} is introduced to test models' commonsense understanding.  People do not need answers to commonsense-probing questions.

\noindent\textbf{Hazard from Immediate Usage:} None, as there is no realistic application. 

\noindent\textbf{Hazard from Downstream Impact:} None, as there is no realistic application. 

\begin{center} \parbox{\linewidth}{\hrulefill} \end{center} %

\subsection*{[\hypertarget{dataset20}{D20}] \textsc{Sensemaking} \citep{wang-etal-2019-make}}

\noindent\textbf{Prediction Task:} Given two sentences, predict which one is nonsensical.  

\noindent\textbf{Average Input Length:}  17 [sentence1 + sentence2] = 17 words

\noindent\textbf{Human Ability:} 99.1\%

\noindent\textbf{Application:} No. \textsc{Sensemaking} is introduced to test models' commonsense understanding. People do not need predictions of which of two sentences is nonsensical. 

\noindent\textbf{Hazard from Immediate Usage:} None, as there is no realistic application. 

\noindent\textbf{Hazard from Downstream Impact:} None, as there is no realistic application. 

\begin{center} \parbox{\linewidth}{\hrulefill} \end{center} %

\subsection*{[\hypertarget{dataset}{D21}] \textsc{WinoWhy} \citep{zhang-etal-2020-winowhy}}

\noindent\textbf{Prediction Task:} Winograd
Schema Challenge (pronoun coreference resolution).

\noindent\textbf{Average Input Length:} 16 [sentence 1] + 24 [sentence 2] = 40 words

\noindent\textbf{Human Ability:}  92.1\% accuracy \citep{DBLP:conf/maics/Bender15}

\noindent\textbf{Application:} No. \textsc{WSC} is introduced to test models' commonsense understanding. People do not need such pronouns resolved (in isolation). 

\noindent\textbf{Hazard from Immediate Usage:} None, as there is no realistic application. 

\noindent\textbf{Hazard from Downstream Impact:} None, as there is no realistic application. 

\begin{center} \parbox{\linewidth}{\hrulefill} \end{center} %
 
\subsection*{[\hypertarget{dataset22}{D22}] \textsc{ChangeMyView} \citep{atkinson-etal-2019-gets}} 

\noindent\textbf{Prediction Task:}  Predicting is a forum counter-argument to someone's opinion persuasive.

\noindent\textbf{Average Input Length:}  351 [opinion] + 215 [counterargument] = 566 words

\noindent\textbf{Human Ability:} Not reported

\noindent\textbf{Application:} Assistant writing when the goal is to write a convincing, but not deceptive, counter-argument by showing the writer if their current response is predicted to change someone's mind.

\noindent\textbf{Hazard from Immediate Usage:}
\begin{compactitem}
    \item \emph{Who:} Anyone
    \item \emph{Hazard:} Failing to change someone's opinion 
    \item \emph{Probability:} Undetermined, as the models' performance for this application is not known
    \item \emph{Severity:} Low
    \item \emph{Risk:} Low
\end{compactitem}

\noindent\textbf{Hazard from Downstream Impact:}
\begin{compactitem}
    \item \emph{Who:} Person who changes their mind
    \item \emph{Hazard:} Not changing their opinion. This does not lead to propagating misinformation as original posts are presented as opinions, not facts. 
    \item \emph{Probability:} Undetermined, as recent models' performance for this application is not known
    \item \emph{Severity:} Low
    \item \emph{Risk:} Low
\end{compactitem}

\begin{center} \parbox{\linewidth}{\hrulefill} \end{center} %

\subsection*{[\hypertarget{dataset23}{D23}] \textsc{SBIC} \citep{sap-etal-2020-social}}

\noindent\textbf{Prediction Task:}  Classify a social media or forum post as offensive or not.

\noindent\textbf{Average Input Length:}  19 words [post]

\noindent\textbf{Human Ability:} Not reported

\noindent\textbf{Application:} Content moderation

\noindent\textbf{Hazard from Immediate Usage:}
\begin{compactitem}
    \item \emph{Who:} Content moderator 
    \item \emph{Hazard:} Job performance problems from repeatedly not flagging attacking comments or flagging non-attacking comments 
    \item \emph{Probability:} Undetermined, as recent models' and human performance are not known
    \item \emph{Severity:} Moderate 
    \item \emph{Risk:} Depends on the probability, but can be moderate 
\end{compactitem}

\noindent\textbf{Hazard from Downstream Impact:}
\begin{compactitem}
    \item \emph{Who:} Someone who is targeted (in-group or personally) by an attacking comment; A poster of an inoffensive post that is flagged
    \item \emph{Hazard:} Mental health harms
    \item \emph{Probability:} Undetermined, as recent models' and human performance are not known
    \item \emph{Severity:} Depends on personal circumstances, but can be moderate
    \item \emph{Risk:} Depends on the probability, but can be moderate
\end{compactitem}

\begin{center} \parbox{\linewidth}{\hrulefill} \end{center} %

\subsection*{[\hypertarget{dataset24a}{D24a}] \citet{DBLP:conf/iclr/WangQZ0YN0R20}; relation extraction}

\noindent\textbf{Prediction Task:}  Relation extraction between people and organizations \cite[\textsc{tacred};][]{zhang-etal-2017-position} or relations that are chosen because they have broad coverage \cite[\textsc{SemEval};][]{hendrickx-etal-2009-semeval}.

\noindent\textbf{Average Input Length:}  36 words [sentence] (\textsc{tacred}) / 19 words [sentence] (\textsc{SemEval})

\noindent\textbf{Human Ability:} Not reported, but we expect good human abilities for the task

\noindent\textbf{Application:} Extraction of TACRED relations will be requested by people in form of open-ended QA. SemEval relations are too generic and we do not see a specific application for them.

\noindent\textbf{Hazard from Immediate Usage:}
\begin{compactitem}
    \item \emph{Who:} Anyone 
    \item \emph{Hazard:} Propagating misinformation about relations between certain people and organizations.
    \item \emph{Probability:} Low, a RoBERTa-based model gets a 91+ F1-score \citep{zhou-chen-2022-improved}.
    \item \emph{Severity:} Low (relations are not about critical information such as health, law, etc.).
    \item Low
\end{compactitem}

\noindent\textbf{Hazard from Downstream Impact:} Same as for the immediate usage. 

\begin{center} \parbox{\linewidth}{\hrulefill} \end{center} %

\subsection*{[\hypertarget{dataset24b}{D24b}] \citet{DBLP:conf/iclr/WangQZ0YN0R20}; sentiment analysis}

\noindent\textbf{Prediction Task:}  Sentiment classification of laptop and restaurant reviews.

\noindent\textbf{Average Input Length:} 15 words [laptop reviews]; 13 words [restaurant reviews]

\noindent\textbf{Human Ability:} Not reported 

\noindent\textbf{Application:} Deciding whether to buy a laptop / visit a restaurant

\noindent\textbf{Hazard from Immediate Usage:}
\begin{compactitem}
    \item \emph{Who:} Laptop buyers, restaurant-goers
    \item \emph{Hazard:} Dissatisfying laptop/restaurant
    \item \emph{Probability:} Low. People take multiple factors, not only a few reviews when buying a laptop or booking a restaurant, especially if more expensive/important. However, if we assume that they looked at only reviews, we still expect the probability to be low since today's models accurately classify the sentiment of  reviews in other domains.
    \item \emph{Severity:} Depends on personal circumstances and expense, but generally low.
    \item \emph{Risk:} Low
\end{compactitem}

\noindent\textbf{Hazard from Downstream Impact:} Nothing noteworthy.

\begin{center} \parbox{\linewidth}{\hrulefill} \end{center}

\subsection*{[\hypertarget{dataset25}{D25}] \textsc{COPA-SSE} \citep{brassard-etal-2022-copa}}

\noindent\textbf{Prediction Task:} Given a premise and two choices, select the choice that more plausibly has a causal relation with the premise.

\noindent\textbf{Average Input Length:} 6 [premise] + 12 [choices] = 18 words

\noindent\textbf{Human Ability:}  ``We have established that human raters can
perform extremely well on this task, with near perfect agreement.'' \citep{DBLP:conf/aaaiss/RoemmeleBG11}

\noindent\textbf{Application:} No. COPA is introduced to test models’ commonsense causal reasoning that people possess.

\noindent\textbf{Hazard from Immediate Usage:} None, as there is no realistic application.

\noindent\textbf{Hazard from Downstream Impact:} None, as there is no realistic application.

\begin{center} \parbox{\linewidth}{\hrulefill} \end{center} %

\subsection*{[\hypertarget{dataset26}{D26}] \textsc{WorldTree v1} \citep{jansen-etal-2018-worldtree}}

\noindent\textbf{Prediction Task: } Multi-choice middle-school level science exam QA  

\noindent\textbf{Average Input Length: } 23 [question] + 20 [options] = 43 words 

\noindent\textbf{Human Ability: } Depends, but can be 100\%

\noindent\textbf{Application: } No. Models trained on this data could be used that students in 3\textsuperscript{rd} through 5\textsuperscript{th} to practice for science exams if exams are available, but not correct solutions. However, practice exams come with solutions. 

\noindent\textbf{Hazard from Immediate Usage:} None, as there
is no realistic application.

\noindent\textbf{Hazard from Downstream Impact:} None, as there
is no realistic application.

\begin{center} \parbox{\linewidth}{\hrulefill} \end{center} %

\subsection*{[\hypertarget{dataset27}{D27}]  \textsc{WorldTree V2} \citep{xie-etal-2020-worldtree}}

\noindent\textbf{Prediction Task:} Multi-choice middle-school level science exam QA.

\noindent\textbf{Average Input Length: } 19 [question] + 15 [options] = 34 words

\noindent\textbf{Human Ability: } Depends, but can be 100\%

\noindent\textbf{Application: } No. Models trained on this data could be used for students in 3\textsuperscript{rd} through 9\textsuperscript{th} grade to practice for science exams if exams are available, but not correct solutions. However, practice exams come with solutions. 

\noindent\textbf{Hazard from Immediate Usage:} None, as there
is no realistic application.

\noindent\textbf{Hazard from Downstream Impact:} None, as there
is no realistic application.

\begin{center} \parbox{\linewidth}{\hrulefill} \end{center} %

\subsection*{[\hypertarget{dataset28}{D28}] \textsc{HotPotQA} \citep{yang-etal-2018-hotpotqa}}

\noindent\textbf{Prediction Task:} Reading comprehension targeting multi-hop reasoning.

\noindent\textbf{Average Input Length:}  4633 [context] + 15 [question] = 4648 words

\noindent\textbf{Human Ability:} 98.8 F1

\noindent\textbf{Application:} Information search 

\noindent\textbf{Hazard from Immediate Usage:}
\begin{compactitem}
    \item \emph{Who:} Anyone
    \item \emph{Hazard:} Propagating misinformation from accepting the wrong answer 
    \item \emph{Probability:} Moderate ($\exists$ model-human performance gap)\footnote{\url{https://hotpotqa.github.io/}; \url{https://paperswithcode.com/sota/question-answering-on-hotpotqa}}
    \item \emph{Severity:} Low (questions are not about critical information such as health, law, etc.)
    \item \emph{Risk:} Low
\end{compactitem}

\noindent\textbf{Hazard from Downstream Impact:} Same as for the immediate usage.

\begin{center} \parbox{\linewidth}{\hrulefill} \end{center} %

\subsection*{[\hypertarget{dataset29}{D29}] \textsc{QED} \citep{lamm-etal-2021-qed}} %

 Extended \textsc{NaturalQuestions} with their explanation annotations. See [\hyperlink{dataset5}{D5}].

\begin{center} \parbox{\linewidth}{\hrulefill} \end{center} %

\subsection*{[\hypertarget{dataset30}{D30}] \textsc{QASC} \citep{DBLP:conf/aaai/KhotCGJS20} / \textsc{e-QASC} \citep{jhamtani-clark-2020-learning}}

\noindent\textbf{Prediction Task:}  Multi-choice middle-school level science exam QA.

\noindent\textbf{Average Input Length:} 8 [question] + 13 [options] = 21 words

\noindent\textbf{Human Ability:} Reported human performance is 93\%

\noindent\textbf{Application: } No. Models trained on this data could be used that middle-school students to practice for science exams if exams are available, but not correct solutions. However, practice exams come with solutions. 

\noindent\textbf{Hazard from Immediate Usage:} None, as there
is no realistic application.

\noindent\textbf{Hazard from Downstream Impact:} None, as there
is no realistic application.

\begin{center} \parbox{\linewidth}{\hrulefill} \end{center} %

\subsection*{[\hypertarget{dataset31}{D31}]  \citet{ye-etal-2020-teaching}}

 Extended \textsc{NaturalQuestions} and \textsc{SQuAD} \citep{rajpurkar-etal-2016-squad} with their explanation annotations. See [\hyperlink{dataset5}{D5}].

\begin{center} \parbox{\linewidth}{\hrulefill} \end{center} %

\subsection*{[\hypertarget{dataset32}{D32}] \textsc{R4C}  \citep{inoue-etal-2020-r4c}}

 Extended \textsc{HotPotQA} with their explanation annotations. See [\hyperlink{dataset28}{D28}].

\begin{center} \parbox{\linewidth}{\hrulefill} \end{center} %

\subsection*{[\hypertarget{dataset33}{D33}] \textsc{TriggerNER} \citep{lin-etal-2020-triggerner}}

\noindent\textbf{Prediction Task:}  Named entity recognition.

\noindent\textbf{Average Input Length:}  14 words [sentence]

\noindent\textbf{Human Ability:} Not reported, but we expect good human abilities for this task

\noindent\textbf{Application:} While NER is a useful component of larger systems (automatic tag generation, information retrieval, content recommendation, etc.), it is not realistic to expect that a person will check each labeled entity manually for another purpose.

\noindent\textbf{Hazard from Immediate Usage:} None, as there is no realistic application.

\noindent\textbf{Hazard from Downstream Impact:} None, as there is no realistic application.

\begin{center} \parbox{\linewidth}{\hrulefill} \end{center} %

\subsection*{[\hypertarget{dataset34}{D34}] \citet{zaidan-etal-2007-using} / \textsc{ERASER Movie Reviews} \citep{deyoung-etal-2020-eraser}}

\noindent\textbf{Prediction Task:} Sentiment classification of movie reviews.

\noindent\textbf{Average Input Length:} 648 words [reviews]

\noindent\textbf{Human Ability:} Reported human performance ranges from 92--97\%

\noindent\textbf{Application:} Deciding whether to go see or rent a movie

\noindent\textbf{Hazard from Immediate Usage:}
\begin{compactitem}
    \item \emph{Who:} Movie watchers
    \item \emph{Hazard:} Buying a cinema ticket or renting a movie they do not like  
    \item \emph{Probability:} Low since sentiment classifiers of movie reviews work well\footnote{\url{https://paperswithcode.com/sota/text-classification-on-imdb}} 
    \item \emph{Severity:} Low since the cost of renting or seeing a movie is generally low 
    \item \emph{Risk:} Low 
\end{compactitem}

\noindent\textbf{Hazard from Downstream Impact:} Nothing noteworthy. 

\begin{center} \parbox{\linewidth}{\hrulefill} \end{center} %

\subsection*{[\hypertarget{dataset35}{D35}] \textsc{Stanford Sentiment Treebank} \citep{socher-etal-2013-recursive}}

\noindent\textbf{Prediction Task:} Sentiment classification of movie reviews.

\noindent\textbf{Average Input Length:} 16 words [review]

\noindent\textbf{Human Ability:} Not reported

\noindent \emph{The rest of the information is the same as for the dataset above ([\hyperlink{dataset35}{D35}]).}

\begin{center} \parbox{\linewidth}{\hrulefill} \end{center} %

\subsection*{[\hypertarget{dataset36}{D36}] \textsc{ERASER BoolQ} \citep{deyoung-etal-2020-eraser}}

\noindent\textbf{Prediction Task:} Answering yes/no questions from a Wikipedia passage.

\noindent\textbf{Average Input Length:} 9 [question] + 93 [passage] = 102 words

\noindent\textbf{Human Ability:} Reported human accuracy is 90\% \citep{clark-etal-2019-boolq}

\noindent\textbf{Application:} Information search 

\noindent\textbf{Hazard from Immediate Usage:}
\begin{compactitem}
    \item \emph{Who:} Anyone
    \item \emph{Hazard:} Propagating misinformation from accepting the wrong answer 
    \item \emph{Probability:} Low since models achieve accuracy above the estimated human accuracy\footnote{\url{https://paperswithcode.com/sota/question-answering-on-boolq}}
    \item \emph{Severity:} Low (questions are not about critical information such as health, law, etc.)
    \item \emph{Risk:} Low
\end{compactitem}

\noindent\textbf{Hazard from Downstream Impact:} Same as for the immediate usage.

\begin{center} \parbox{\linewidth}{\hrulefill} \end{center} %

\subsection*{[\hypertarget{dataset37}{D37}] \citet{hancock-etal-2018-training}}

\noindent\textbf{Prediction Task:} Given a sentence with highlighted (1) names of two people, predict whether they are spouses, (2) a chemical and a disease, predict whether the disease is chemical-induced, and (3) a protein and a kinase, predict ``whether or not the kinase influences the protein in terms of a physical interaction or phosphorylation''.

\noindent\textbf{Average Input Length:}  (1)  23 [sentence with a spouse relationship], (2) 10 words [a sentence with a chemical-disease pair], (3) The protein data is not available. 

\noindent\textbf{Human Ability:} Not reported

\noindent\textbf{Application:} (1) No, we expect there is no interest in a tool that only predicts whether two people named in a given sentence are spouses. (2) 
Automatic completion of bioinformatics databases based on new biomedical publications.  
(3) \citeauthor{hancock-etal-2018-training} say that predicting a relation between a given protein and kinase can be useful for ``targeting biological pathways of Parkinson’s disease''. 

\noindent\textbf{Hazard from Immediate Usage:} We focus on (2) that has an application and its data is available. 
\begin{compactitem}
    \item \emph{Who:} Biocurator 
    \item \emph{Hazard:} Accepting a wrong prediction and consequently (1) adding to a database a wrong relation or (2) not adding a correct relation. These can result in the biocurator's job performance problems if done repeatedly and propagating misinformation.  
    \item \emph{Probability:} Undetermined, as the recent models' performance for this application is not known
    \item \emph{Severity:} Moderate 
    \item \emph{Risk:} Depends on the probability, but could be moderate 
\end{compactitem}

\noindent\textbf{Hazard from Downstream Impact:}
\begin{compactitem}
    \item \emph{Who:} Scientist/biologist; Database owner
    \item \emph{Hazard:} Getting the wrong information about a relation; Providing wrong or missing information to their customers based on their biocurators' final decisions
    \item \emph{Probability:}  Undetermined, as the recent models' performance for this application is not known
    \item \emph{Severity:} Moderate 
    \item \emph{Risk:} Depends on the probability, but could be moderate 
\end{compactitem}

\begin{center} \parbox{\linewidth}{\hrulefill} \end{center} %

\subsection*{[\hypertarget{dataset38}{D38}] \textsc{ERASER EvidenceInference} \citep{deyoung-etal-2020-eraser,lehman-etal-2019-inferring} / \textsc{EvidenceInference v2} \citep{deyoung-etal-2020-evidence}}

\noindent\textbf{Prediction Task:}  Given a scientific article, predict whether administering treatment A will significantly increase/decrease some outcome compared to treatment B, or make no difference. The original task setup asks for a model to predict the effect of a treatment given a relevant publication. Having the relevant publication provided is not realistic, but the available data allows a more realistic setup where it must be retrieved first.

\noindent\textbf{Average Input Length:} 11 [query (treatment A | treatment B | outcome)] + 3627 [passage] = 3638 words

\noindent\textbf{Human Ability:} Not reported

\noindent\textbf{Application:} Treatment recommendations

\noindent\textbf{Hazard from Immediate Usage:}
\begin{compactitem}
    \item \emph{Who:} Clinicians
    \item \emph{Hazard:}  Cause physical harm to patients; Job performance problems 
    \item \emph{Probability:} Moderate; \citet{DBLP:conf/aaai/ChrysostomouA22} report 83\% F1 score
    \item \emph{Severity:} Can be high (for both hazards)
    \item \emph{Risk:} High (for both hazards)
\end{compactitem}

\noindent\textbf{Hazard from Downstream Impact:}
\begin{compactitem}
    \item \emph{Who:} Patients 
    \item \emph{Hazard:} Getting hurt
    \item \emph{Probability:} Moderate; \citet{DBLP:conf/aaai/ChrysostomouA22} report 83\% F1 score
    \item \emph{Severity:} Can be high 
    \item \emph{Risk:} High 
\end{compactitem}

\begin{center} \parbox{\linewidth}{\hrulefill} \end{center} %

\subsection*{[\hypertarget{dataset39}{D39}] \textsc{ERASER MultiRC} \citep{deyoung-etal-2020-eraser, khashabi-etal-2018-looking}}

\noindent\textbf{Prediction Task:}  Multiple-choice QA from a few passages.

\noindent\textbf{Average Input Length:} 15 [question] + 43 [passage] = 58 words

\noindent\textbf{Human Ability:} 84.3 F1-score

\noindent\textbf{Application:} No. MultiRC is introduced to probe models' multiple-choice reading comprehension abilities when they need to take ``into account information from multiple sentences''. If we imagine a version without answer choices, we still deem that there is no realistic application because the source documents are not broad enough for open-ended QA (search engines) but also not specific enough (e.g., healthcare-related questions).

\noindent\textbf{Hazard from Immediate Usage:} None, as there is no realistic application.

\noindent\textbf{Hazard from Downstream Impact:} None, as there is no realistic application.

\begin{center} \parbox{\linewidth}{\hrulefill} \end{center} %

\subsection*{[\hypertarget{dataset40}{D40}] \textsc{WikiQA} \citep{yang-etal-2015-wikiqa}}

\noindent\textbf{Prediction Task:} Identifying a span in a Wikipedia article that answers an open-ended question (originally asked in Bing).

\noindent\textbf{Average Input Length:} 234 [Wikipedia summary] + 7 [question] = 241 words

\noindent\textbf{Human Ability:} Not reported

\noindent\textbf{Application:} Information search 

\noindent\textbf{Hazard from Immediate Usage:}
\begin{compactitem}
    \item \emph{Who:} Anyone
    \item \emph{Hazard:} Propagating misinformation from accepting the wrong answer 
    \item \emph{Probability:} Low since models achieve high performance\footnote{\url{https://paperswithcode.com/sota/question-answering-on-wikiqa}}
    \item \emph{Severity:} Low (questions are not about critical information such as health, law, etc.)
    \item \emph{Risk:} Low
\end{compactitem}

\noindent\textbf{Hazard from Downstream Impact:} Same as for the immediate usage.

\begin{center} \parbox{\linewidth}{\hrulefill} \end{center} %

\subsection*{[\hypertarget{dataset41}{D41}] \textsc{WikiAttack} \citep{carton-etal-2018-extractive}}

\noindent\textbf{Prediction Task:}  Predict is a Wikipedia revision comment on a personal attack.

\noindent\textbf{Average Input Length:} 65 words [comment] 

\noindent\textbf{Human Ability:} Not reported

\noindent\textbf{Application:} Content moderation 

\noindent\textbf{Hazard from Immediate Usage:}
\begin{compactitem}
    \item \emph{Who:} Content moderator 
    \item \emph{Hazard:} Job performance problems from repeatedly not flagging attacking comments or flagging non-attacking comments 
    \item \emph{Probability:} Undetermined, as recent models' and human performance are not known
    \item \emph{Severity:} Moderate 
    \item \emph{Risk:} Depends on the probability, but can be moderate 
\end{compactitem}

\noindent\textbf{Hazard from Downstream Impact:}
\begin{compactitem}
    \item \emph{Who:} Someone who is targeted (in-group or personally) by an attacking comment; A poster of an inoffensive post that is flagged
    \item \emph{Hazard:} Mental health harms
    \item \emph{Probability:} Undetermined, as recent models' and human performance are not known
    \item \emph{Severity:} Depends on personal circumstances, but can be moderate
    \item \emph{Risk:} Depends on the probability, but can be moderate
\end{compactitem}

\begin{center} \parbox{\linewidth}{\hrulefill} \end{center} %

\subsection*{[\hypertarget{dataset42}{D42}] UKPSnopes \citep{hanselowski-etal-2019-richly}}

\noindent\textbf{Prediction Task:}  Verification of claims about a broad range of topics, given an article from Snopes fact-checking website%
, which is not a realistic application setup. 
However, the available data could possibly allow a more realistic setup where relevant documents (that are not fact-checking reports) must be retrieved first. 
After running various experiments, it became clear that these documents were insufficient for solving the task (refer to \sect{sec:ukp_snopes_analysis} for more details), and there is a need for constructing a more comprehensive and suitable document corpus to retrieve relevant articles from.

\noindent\textbf{Average Input Length:} 15 [claim] + 947 [documents] = 962 words

\noindent\textbf{Human Ability:} 80.2\% F1-score 

\noindent\textbf{Application:} No. The dataset does not represent a realistic task setup (similar to PubHlealth ([\protect\hyperlink{dataset16}{D16}]). The veracity of the claims is assessed based on an article that specifically discusses the target claim, which does not exist in real-world situations.

\noindent\textbf{Hazard from Immediate Usage:} None, as there is no realistic application.

\noindent\textbf{Hazard from Downstream Impact:} None, as there is no realistic application.

\begin{center} \parbox{\linewidth}{\hrulefill} \end{center} %

\subsection*{[\hypertarget{dataset43}{D43}] \textsc{CoQA} \citep{reddy-etal-2019-coqa}}

\noindent\textbf{Prediction Task:} ``Given a passage and a conversation so far, the task is to answer the next question in the conversation.''

\noindent\textbf{Average Input Length:} 264 [passage] + 5 [question] + 3 [answer] = 272 words%

\noindent\textbf{Human Ability:} Reported human F1 score is 88.8 

\noindent\textbf{Application:} No. Resembles conversational information search, but the first question in CoQA conversations is not standalone (without the passage), e.g., ``Who had a birthday'', so unlike StrategyQA (\hyperlink{dataset6}{D6}) and NaturalQuestions (\hyperlink{dataset5}{D5}) we cannot re-purpose CoQA such that for the first question in the conversation, we retrieve the relevant article then the most relevant passage in it, (i.e., for conversational information search). 

\noindent\textbf{Hazard from Immediate Usage:} None, as there is no realistic application. 

\noindent\textbf{Hazard from Downstream Impact:} 
None, as there is no realistic application.

\begin{center} \parbox{\linewidth}{\hrulefill} \end{center} %

\subsection*{[\hypertarget{dataset44}{D44}] \textsc{SciFact-Open} \citep{wadden-etal-2022-scifact}; \textsc{SciFact} \citep{wadden-etal-2020-fact}}

\noindent\textbf{Prediction Task:}  Given a claim and a set of abstracts, the \emph{open} scientific claim verification task asks a model to first retrieve abstracts that are relevant for verifying a given claim, and then for each retrieved abstract, predict whether it provides the evidence that supports or refutes the claim.

\noindent\textbf{Average Input Length:} 11 [claim] + 12 [title] + 1860 [retrieved abstracts] = 1883 words\footnote{The models are set to retrieve 10 relevant abstracts from the corpus and the average paragraph length is 186.}

\noindent\textbf{Human Ability:} \citet{wadden-etal-2022-multivers} estimate human performance in the setting where relevant abstracts are provided to be 89.1\% F1 score

\noindent\textbf{Application:} Scientific claim verification

\noindent\textbf{Hazard from Immediate Usage:}
\begin{compactitem}
    \item \emph{Who:} Clinicians; Researchers/readers of the relevant journals; Anyone
    \item \emph{Hazard:} Cause physical harm to patients; Publishing new articles based on wrong answers; Defamation; Job performance problems; Propagating misinformation from accepting the wrong answer 
    \item \emph{Probability:} Moderate--High (models do not achieve very high F1 score in the more realistic setup with 500K abstracts) 
    \item \emph{Severity:} Can be high (for all hazards)
    \item \emph{Risk:} High (for all hazards)
\end{compactitem}

\noindent\textbf{Hazard from Downstream Impact:}
\begin{compactitem}
    \item \emph{Who:} Patients; Anyone  
    \item \emph{Hazard:} Getting hurt; Propagating misinformation from accepting the wrong answer from a person who was misinformed by the model
    \item \emph{Probability:} Moderate--High (see immediate impact)
    \item \emph{Severity:} Can be high (for both hazards) 
    \item \emph{Risk:} High (for both hazards)
\end{compactitem}

\begin{center} \parbox{\linewidth}{\hrulefill} \end{center} %

\subsection*{[\hypertarget{dataset45}{D45}] \citet{DBLP:journals/jair/KutluMEL20}}

\noindent\textbf{Prediction Task:}  Rating the relevance of Web pages for
different search topics.

\noindent\textbf{Average Input Length:}Data (documents/webpages and search topics/queries) are not available.

\noindent\textbf{Human Ability:} Reported human accuracy is 65\% 

\noindent\textbf{Application:} Information search

\noindent\textbf{Hazard from Immediate Usage:}
\begin{compactitem}
    \item \emph{Who:} Anyone
    \item \emph{Hazard:} Propagating misinformation from accepting the wrong answer 
    \item \emph{Probability:} N/A
    \item \emph{Severity:} Low (questions are not about critical information such as health, law, etc.)\footnote{\url{https://trec.nist.gov/data/web/09/wt09.topics.queries-only}}
    \item \emph{Risk:} Low  
\end{compactitem}

\noindent\textbf{Hazard from Downstream Impact:} Same as for the immediate usage.

\begin{center} \parbox{\linewidth}{\hrulefill} \end{center} %

\subsection*{[\hypertarget{dataset46}{D46}] \textsc{ECtHR} \citep{chalkidis-etal-2021-paragraph}}

\noindent\textbf{Prediction Task:} ``Given a set of paragraphs that refer to the facts of each case [...] in judgments of the European Court of Human Rights (ECtHR), [...] predict the allegedly violated articles of the European Convention of Human Rights (ECHR).'' 

\noindent\textbf{Average Input Length:} 1579 words [facts sequence]

\noindent\textbf{Human Ability:} Not reported %

\noindent\textbf{Application:} No.  The facts of a
case are explicitly provided by legal professionals while in real-world situations, they are not. This is similar to PubHealth ([\protect\hyperlink{dataset16}{D16}]) where a claim is verified based on a report about this claim written by a professional fact checker.  The ILDC version (see [\hyperlink{dataset50}{D50}]) with unstructured/unannotated case proceedings is more realistic. %

\noindent\textbf{Hazard from Immediate Usage:} None, as there is no realistic application.

\noindent\textbf{Hazard from Downstream Impact:} None, as there is no realistic application.

\begin{center} \parbox{\linewidth}{\hrulefill} \end{center} %

\subsection*{[\hypertarget{dataset47}{D47}] \textsc{Hummingbird} \citep{hayati-etal-2021-bert}}

\noindent\textbf{Prediction Task:} Classifying text if it has the following styles: politeness, sentiment, offensiveness, and five emotion types.

\noindent\textbf{Average Input Length:} 184 words [sentence]

\noindent\textbf{Human Ability:} Inter-annotator agreement ranges from $\approx$63 (politeness) to $\approx$83 (joy) %

\noindent\textbf{Application:} Assistant writing when the goal is to write text with one of the styles above

\noindent\textbf{Hazard from Immediate Usage:}
\begin{compactitem}
    \item \emph{Who:} Anyone 
    \item \emph{Hazard:} Writing text in undesired style, e.g., not sufficiently polite or sad
    \item \emph{Probability:} Low-Moderate \cite[based on the 2021 model performance;][]{hayati-etal-2021-bert}
    \item \emph{Severity:} Depends who is the text written for, but generally low
    \item \emph{Risk:} Low
\end{compactitem}

\noindent\textbf{Hazard from Downstream Impact:} Nothing noteworthy.

\begin{center} \parbox{\linewidth}{\hrulefill} \end{center} %

\subsection*{[\hypertarget{dataset48}{D48}] \textsc{HateXplain} \citep{DBLP:conf/aaai/MathewSYBG021}}

\noindent\textbf{Prediction Task:} Hate speech detection.

\noindent\textbf{Average Input Length:} 23 words [sentence]

\noindent\textbf{Human Ability:} Not reported %

\noindent\textbf{Application:} Content moderation

\noindent\textbf{Hazard from Immediate Usage:}
\begin{compactitem}
    \item \emph{Who:} Content moderator 
    \item \emph{Hazard:} Job performance problems from repeatedly not flagging attacking comments or flagging non-attacking comments 
    \item \emph{Probability:} Undetermined, as recent models' and human performance are not known
    \item \emph{Severity:} Moderate 
    \item \emph{Risk:} Depends on the probability, but can be moderate 
\end{compactitem}

\noindent\textbf{Hazard from Downstream Impact:}
\begin{compactitem}
    \item \emph{Who:} Someone who is targeted (in-group or personally) by an attacking comment; A poster of an inoffensive post that is flagged
    \item \emph{Hazard:} Mental health harms
    \item \emph{Probability:} Undetermined, as recent models' and human performance are not known
    \item \emph{Severity:} Depends on personal circumstances, but can be moderate
    \item \emph{Risk:} Depends on the probability, but can be moderate
\end{compactitem}

\begin{center} \parbox{\linewidth}{\hrulefill} \end{center} %

\subsection*{[\hypertarget{dataset49}{D49}] \textsc{ContractNLI} \citep{koreeda-manning-2021-contractnli-dataset}}

\noindent\textbf{Prediction Task:}  ``Given a contract and a set of hypotheses (each being a sentence),  classify whether each hypothesis is entailed by, contradicting to or not mentioned by (neutral to) the contract''.

\noindent\textbf{Average Input Length:} 1631 [contract] + 13 [hypothesis] = 1644 words

\noindent\textbf{Human Ability:} Not reported %

\noindent\textbf{Application:} Reviewing a contract

\noindent\textbf{Hazard from Immediate Usage:} 
\begin{compactitem}
    \item \emph{Who:} Business owner; Person working for a company that reviews contracts
    \item \emph{Hazard:}  Incorrectly reviewing the contract leading to business damages/liability; Job performance problems
    \item \emph{Probability:} High \cite[based on the model's performance for the contradiction label;][]{koreeda-manning-2021-contractnli-dataset}
    \item \emph{Severity:} High
    \item \emph{Risk:} High
\end{compactitem}

\noindent\textbf{Hazard from Downstream Impact:}
\begin{compactitem}
    \item \emph{Who:} A company hired someone to review their contracts  
    \item \emph{Hazard:} Getting an incorrectly reviewed contract leading to business damages/liability
    \item \emph{Probability:} High \cite[based on the model's performance for the contradiction label;][]{koreeda-manning-2021-contractnli-dataset}
    \item \emph{Severity:} High
    \item \emph{Risk:} High
\end{compactitem}

\begin{center} \parbox{\linewidth}{\hrulefill} \end{center} %

\subsection*{[\hypertarget{dataset49}{D49}] \textsc{ContractNLI} \citep{koreeda-manning-2021-contractnli-dataset}}

\noindent\textbf{Prediction Task:}  ``Given a contract and a set of hypotheses (each being a sentence),  classify whether each hypothesis is entailed by, contradicting to or not mentioned by (neutral to) the contract''.

\noindent\textbf{Average Input Length:} 1631 [contract] + 13 [hypothesis] = 1644 words

\noindent\textbf{Human Ability:} Not reported %

\noindent\textbf{Application:} Reviewing a contract

\noindent\textbf{Hazard from Immediate Usage:} 
\begin{compactitem}
    \item \emph{Who:} Business owner; Person working for a company that reviews contracts
    \item \emph{Hazard:}  Incorrectly reviewing the contract leading to business damages/liability; Job performance problems
    \item \emph{Probability:} High \cite[based on the model's performance for the contradiction label;][]{koreeda-manning-2021-contractnli-dataset}
    \item \emph{Severity:} High
    \item \emph{Risk:} High
\end{compactitem}

\noindent\textbf{Hazard from Downstream Impact:}
\begin{compactitem}
    \item \emph{Who:} A company hired someone to review their contracts  
    \item \emph{Hazard:} Getting an incorrectly reviewed contract leading to business damages/liability
    \item \emph{Probability:} High \cite[based on the model's performance for the contradiction label;][]{koreeda-manning-2021-contractnli-dataset}
    \item \emph{Severity:} High
    \item \emph{Risk:} High
\end{compactitem}

\begin{center} \parbox{\linewidth}{\hrulefill} \end{center} %

\subsection*{[\hypertarget{dataset50}{D50}] \textsc{ILDC} \citep{malik-etal-2021-ildc}}

\noindent\textbf{Prediction Task:} Based on a case proceeding document from
the Supreme Court of India, predict ``whether the claim(s) filed
by the appellant/petitioner against the respondent
should be accepted or rejected''.

\noindent\textbf{Average Input Length:}  3731 words [petition]  (ILDC${Multi}$), 3731 words [petition] (ILDC${Single}$)

\noindent\textbf{Human Ability:} Reported average expert accuracy is 94\%

\noindent\textbf{Application:} AI-assisted judicial decision making

\noindent\textbf{Hazard from Immediate Usage:}
\begin{compactitem}
    \item \emph{Who:} SCI legal professionals 
    \item \emph{Hazard:} Accepting a claim that should be rejected or rejecting a claim that should be accepted 
    \item \emph{Probability:} Moderate \cite[based on the 2021 model's performance;][]{malik-etal-2021-ildc}
    \item \emph{Severity:} High
    \item \emph{Risk:} High
\end{compactitem}

\noindent\textbf{Hazard from Downstream Impact:}
\begin{compactitem}
    \item \emph{Who:} Appellants/petitioners; Respondents
    \item \emph{Hazard:} Getting a wrong decision for their claim; Wrongful accusation/defamation 
    \item \emph{Probability:} Moderate \cite[based on the 2021 model's performance;][]{malik-etal-2021-ildc}
    \item \emph{Severity:} High (for both hazards)
    \item \emph{Risk:} High (for both hazards)
\end{compactitem}

\begin{center} \parbox{\linewidth}{\hrulefill} \end{center} %

\subsection*{[\hypertarget{dataset51}{D51}]  \citet{ling-etal-2017-program}}

\noindent\textbf{Prediction Task:} Solving multiple-choice algebraic word problems.

\noindent\textbf{Average Input Length:} 31 [question] + 10 [choices] = 41 words

\noindent\textbf{Human Ability:} Depends, but can  be 100\%

\noindent\textbf{Application:} No. Models trained on this data could be used by college students to practice for GMAT/GRE if exams are available, but not correct solutions. However, practice exams come with solutions. 

\noindent\textbf{Hazard from Immediate Usage:} None, as there is no realistic application. 

\noindent\textbf{Hazard from Downstream Impact:} 
None, as there is no realistic application. 

\begin{center} \parbox{\linewidth}{\hrulefill} \end{center} %

\subsection*{[\hypertarget{dataset52}{D52}] \citet{srivastava-etal-2017-joint}}

\noindent\textbf{Prediction Task:}  Classification of the purpose of an email (including an email to oneself) into 7 categories: ``personally keep note of a person contact'', ``requesting something to be done [from an employee]'', ``asking [a friend] to meet up at some event'', sharing ``something humorous from the Internet'' to a friend, ``request a meeting about something'', ``announcement of some new policy'', ``reminder to do something''.

\noindent\textbf{Average Input Length:}  Data not available

\noindent\textbf{Human Ability:} Not reported

\noindent\textbf{Application:} No. These are personal reminders, and we anticipate that people do not want them to be categorized automatically in these specific 7 categories. 

\noindent\textbf{Hazard from Immediate Usage:} None, as there is no realistic application. 

\noindent\textbf{Hazard from Downstream Impact:} 
None, as there is no realistic application.

\begin{center} \parbox{\linewidth}{\hrulefill} \end{center} %

\end{document}